\documentclass[journal]{IEEEtran}
\ifCLASSINFOpdf
\else
\fi
\usepackage{times}
\usepackage{epsfig}
\usepackage{hyperref}
\usepackage{url}
\usepackage{array}
\usepackage{microtype}
\usepackage{graphicx}
\usepackage{amsmath}
\usepackage{bm}
\usepackage{dsfont}
\usepackage{amssymb}
\usepackage{multirow}
\usepackage{booktabs} 
\usepackage{algorithm}
\usepackage{algpseudocode}
\usepackage{varwidth}
\usepackage{pifont}
\usepackage{makecell}
\usepackage{soul}
\usepackage{subcaption}
\usepackage{makecell}
\usepackage[font=small,labelfont=bf,tableposition=top]{caption}
\usepackage{cite}
\usepackage{colortbl}

\newcommand{\norm}[1]{\left\lVert#1\right\rVert}

\begin{document}
%
\title{Font Completion and Manipulation by \\ Cycling Between Multi-Modality Representations}

\author{Ye Yuan, Wuyang Chen, Zhaowen Wang, Matthew Fisher, Zhifei Zhang, Zhangyang Wang, and Hailin Jin

\thanks{Ye Yuan is with the Texas A\&M University. E-mail: ye.yuan@tamu.edu}
\thanks{Wuyang Chen and Zhangyang Wang are with the Department of Electrical and Computer Engineering, The University of Texas at Austin, TX, 78712. E-mail: \{wuyang.chen,atlaswang\}@utexas.edu}
\thanks{Zhaowen Wang, Matthew Fisher, Zhifei Zhang, and Hailin Jin are with the Adobe Research, San Jose, CA, 95100. E-mail: \{zhawang,matfishe,zzhang,hljin\}@utexas.edu}}

\markboth{IEEE TRANSACTIONS ON MULTIMEDIA}%
{Shell \MakeLowercase{\textit{et al.}}: Bare Demo of IEEEtran.cls for IEEE Journals}
%



\maketitle

\begin{abstract}
Generating font glyphs of consistent style from one or a few reference glyphs, i.e., font completion, is an important task in topographical design.
As the problem is more well-defined than general image style transfer tasks, thus it has received interest from both vision and machine learning communities.
Existing approaches address this problem as a direct image-to-image translation task. In this work, we innovate to explore the generation of font glyphs as 2D graphic objects with the graph as an intermediate representation, so that more intrinsic graphic properties of font styles can be captured. Specifically, we formulate a cross-modality cycled image-to-image model structure with a graph constructor between an image encoder and an image renderer. The novel graph constructor maps a glyph’s latent code to its graph representation that matches expert knowledge, which is trained to help the translation task. Our model generates improved results than both image-to-image baseline and previous state-of-the-art methods for glyph completion. Furthermore, the graph representation output by our model also provides an intuitive interface for users to do local editing and manipulation. Our proposed cross-modality cycled representation learning has the potential to be applied to other domains with prior knowledge from different data modalities. Our code is available at \url{https://github.com/VITA-Group/Font_Completion_Graph}.
\end{abstract}


%
\IEEEpeerreviewmaketitle

\begin{figure}[ht!]
  \centering
  \includegraphics[width=\columnwidth]{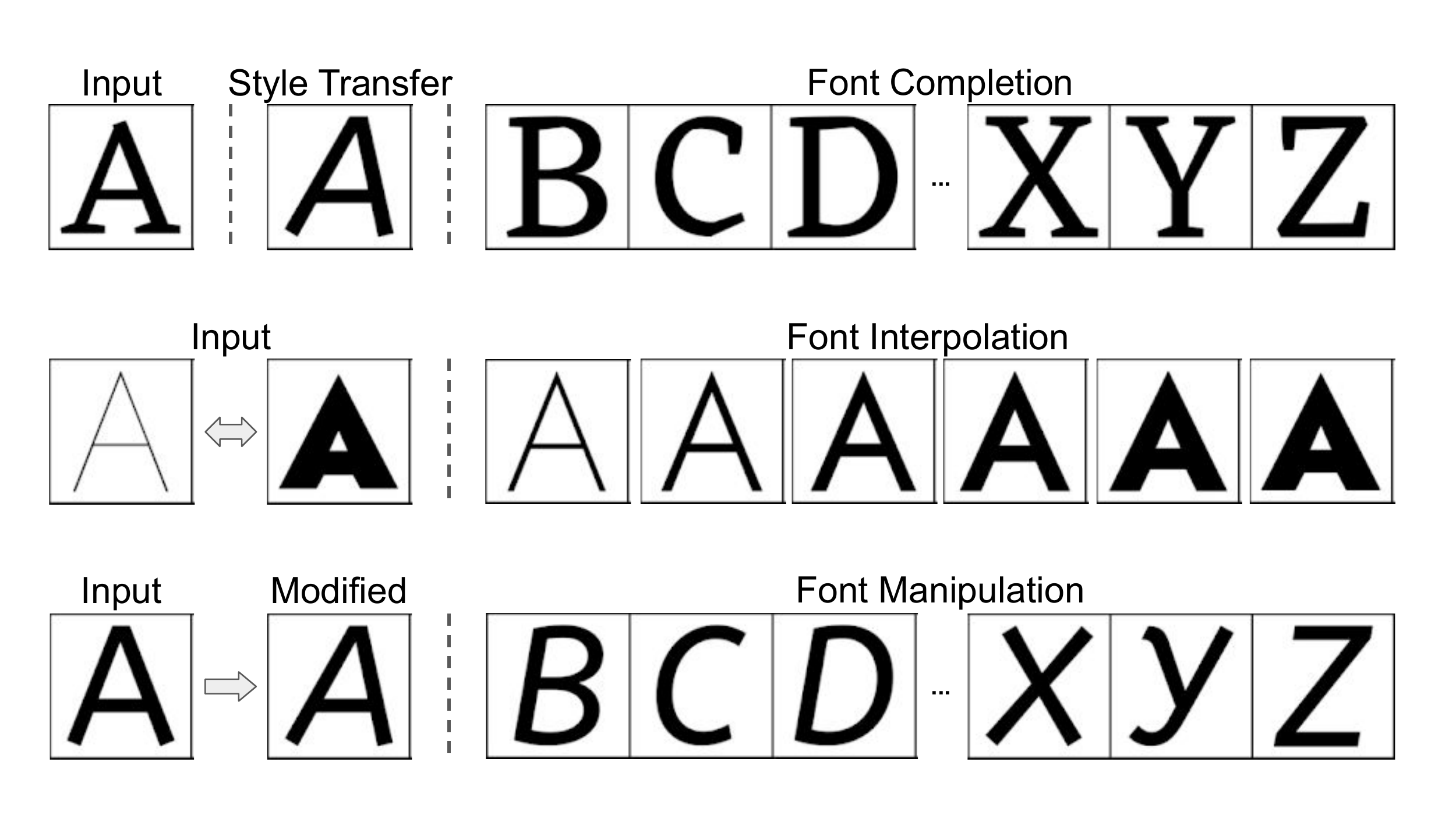}
  \caption{Unlike style transfer that learns a mapping from a single style to another, font completion learns to transfer arbitrary input style and generate new contents for all the glyph set. Our proposed image-to-graph-to-image method (Section \ref{sec:framework}) learns a graph representation (Section \ref{sec:why_graph}) for font glyphs, which facilitates font completion, interpolation and manipulation tasks.}
  \label{fig:task}
\end{figure}

\section{Introduction}

The design of different fonts plays an essential role in multimedia, which can not only convey various moods but also preserve the intellectual property value of designers.
A font is a set of glyphs sharing the same style.
However, to design a font, it is hard to keep the style of all glyphs consistent, especially for languages of significant morphological variations.
Therefore, it is extremely time and resource-consuming for a multimedia designer to manually create new fonts for a set of glyphs.

Due to the success of visual style transfer works \cite{gatys2016image,li2017universal,li2018closed}, more recently the research community shifted attention to address the above problem via the font completion task.
Text glyphs, as a common category of visual signals, have a clear decoupled definition of style (i.e., font) and content (i.e., glyph).
Font completion targets on transferring unseen styles to all glyphs in a font, by using only one or a few sample characters as input. This is different from traditional image style transfer \cite{huang2017arbitrary} which only creates a one-to-one mapping between two domains (Figure \ref{fig:task}). This task has important applications in typography and design, since it can significantly help speed up the process of designing or modifying fonts.

Generating glyphs with the same font style could be seen as an image-to-mage translation problem.
Image-to-image translation \cite{wang2018high} is arguably the most successful framework for image synthesis tasks including photo stylization. Existing image-to-image translation tasks are commonly formulated as extracting a style feature from a reference image, and combining the style feature with the content feature of a target image. Combining the power of deep convolutional neural network (CNN) and generative adversarial network (GAN), image-to-image often produces high-quality stylization results. Unpaired image-to-image \cite{liu2017unsupervised,zhu2017unpaired} further generalizes the framework by relaxing the requirement on paired supervision. Application to font glyphs
\cite{park2018typeface}
shows promising results in transferring the font style in terms of both shape and texture.
However, there are still limitations in the image-to-image framework for font applications. It is observed in our experiments that the supervised image-to-image suffers from poor style/content separation capability.
More importantly, the generated results can only capture the global shape transform but not the high-resolution local details, which is particularly vital to the font style. The generated images often have unsmooth or blurry boundaries.

One key insight from this paper is to advocate that font images form a very restricted and well-defined sub-family of general images. In particular, 
font glyphs are vector objects defined by parametric 2D outlines (e.g. TrueType \cite{ttf}) or strokes (e.g. Metafont \cite{metafont}). These native font formats provide more compact and intrinsic representations from geometric parameter manifolds than pixel representations in the image space. That could be treated as a \underline{strong domain-specific prior} that lays the groundwork for the generation of finer details specific to font images, yet was largely ignored by prior work. To the best of our knowledge, the joint modeling of parametric and perceptual data modalities is still an underexplored area in the research community. In this paper, we aim to investigate how to utilize such \underline{multi-modality representation} to facilitate the font completion task by generating glyphs with both faithful global structure (glyph information) and fine-grained local details (style information).

\begin{figure}[t!]
\begin{center}
    \includegraphics[width=\linewidth]{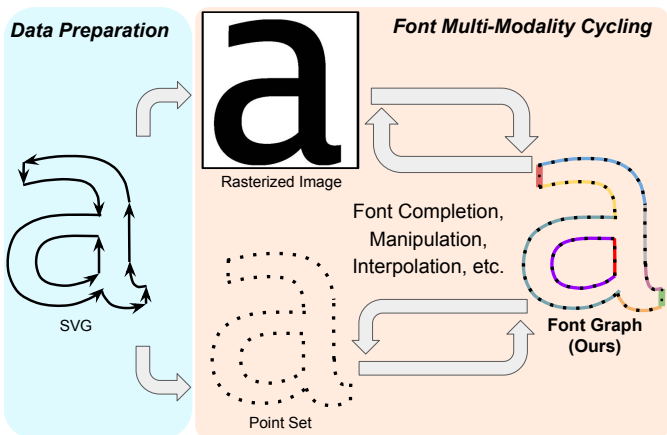}
\end{center}
\caption{We propose to leverage the graph structure of font glyphs as intermediate representation to address the font completion problem. After data preparation from SVG (Section \ref{sec:dataset}), our framework (Section \ref{sec:framework}) can \textbf{cycle between different glyph modalities} and thus enable \textbf{high-quality font completion, manipulation and interpolation}. The proposed graph representation, serving as a transition state between other modalities including point set and rasterized image, can capture both global content and local style features. It is also resolution-free and allows easy manipulation by designers. The graph has a hierarchical structure consisting of strokes (colored curves) and point nodes (dots).}
\label{cycle_architecture}
\end{figure}

To this end, we propose a \textbf{graph-based glyph representation} as an intermediate stage to improve conventional image-based glyph style transfer. As shown in Figure~\ref{cycle_architecture}, the graph has a hierarchical structure, consisting of primary nodes segmenting glyph strokes, secondary nodes within each stroke, and edges connecting the entire glyph contour. The primary nodes control the global glyph structure and the secondary nodes (dense sample points) describe the local region details. Each node has both coordinates and tangents information so that the original curve segments can be fully captured. Our graph can express any connection between strokes. It is a \textbf{scalable} representation similar to Scalable Vector Graphics (SVG) format, and also a computational representation \textbf{easy to integrate with deep learning models}. Furthermore, the graph representation generated by our model also provides an \textbf{intuitive intermediate interface} for designers to do local editing and manipulation on existing font glyphs.

Specifically, we leverage four representations of the font glyphs in our framework: SVG curves, rasterized images, point set (coordinates and tangents), and graph (with each point in point set as a node and the connections along contour as edges). We adopt a cross-modality auto-encoder framework to learn font style features and transitions between different representations. For font completion (Figure~\ref{fig:completion_model_structure}), given a reference source glyph, we first extract the font style vector with an image encoder. Then a graph constructor for the target glyph synthesizes the stylized target glyph in its graph representation, which is further converted to the pixel output by a graph transformer-based image renderer. On the Google Fonts dataset \cite{googlefonts}, our method attains noticeable improvement over baseline and previous method in both quantitative and qualitative evaluation.

To summarize, our contributions include:
\begin{enumerate}
    \item We design a graph representation for font glyphs to better capture style and glyph information for font completion problem (Section \ref{sec:why_graph}). The graph representation is scalable and serves as a pivot to train conversion to other formats (rasterized images, point set, etc.).
    \item We design a cross-modality auto-encoder framework towards high-quality font completion (Section \ref{sec:framework}).
    \item We collected and processed a font dataset that includes all four modalities for the first time (Section \ref{sec:dataset}).
    \item With the superior properties of our graph representation, we further enable diverse font tasks, including font interpolation (\ref{sec:interpolation}) and local manipulation (\ref{sec:manipulation}).
\end{enumerate}

\section{Related Work}

\subsection{Font Completion and Stylization}

Several methods have been proposed to complete a typeface from a small number of example glyphs. Initial work here focused on analyzing geometric features and learning a non-linear mapping to interpolate and extrapolate existing fonts~\cite{suveeranont2010example,campbell2014learning}. Early work using deep networks used a VAE encoder to learn a font embedding that could be decoded into each glyph~\cite{upchurch2016z}. More recent deep learning methods are mostly based on image-to-image translation adversarial networks~\cite{isola2017image}.
\cite{azadi2018multi} proposed to transfer Latin font styles between rasterized images by using a glyph completion GAN followed by a stylization network. This has been generalized to Chinese typefaces also with a GAN-based architecture~\cite{jiang2017dcfont}. A shape-matching GAN has been developed to help control the stylization of the generated glyphs~\cite{yang2019controllable}. To support the automatic generation of target typographic font with a small subset of target glyph images, \cite{miyazaki2019automatic} extracted the natural strokes from glyph images, and constructs glyphs by deploying the appropriate strokes onto skeletons.

TCN~\cite{park2018typeface} utilized all the existing fonts to train a model that can extract content features and style features from a rasterized image, and then replace the style feature to reconstruct images of new glyphs of the same style. StarFont~\cite{yan2019starfont} focused on Chinese font completion and treats each glyph as a class and abandons reconstruction loss to reduce the total number of example glyphs needed.

\subsection{Vector Font Learning}

A small number of font learning methods operated in the vector domain, such as SVG-VAE~\cite{lopes2019learned}, a generative model that takes a rasterized image of icons as input, and outputs the corresponding icons in scalable vector graphics (SVGs) format. A method for finding correspondences between glyphs has been developed that leverages an unsupervised Bezier representation~\cite{smirnov2020deep}.
We build on these approaches to demonstrate the effectiveness of a model able to jointly operate in both the raster and vector domains.

\subsection{Graph Representation Learning}

In computer vision and graphics, graph-structured representations have attained widespread use to efficiently represent compositional scenes in large-scale images.
Scene graphs have been used for image retrieval \cite{johnson2015image} and generation \cite{johnson2018image}, where nodes are objects and edges give relationships between objects.
Structured representations of visual scenes can also explicitly encode certain types of properties, such as attributes, object co-occurrence, or spatial relationships between pairs of objects~\cite{fisher2011characterizing}. 3D shape segmentation is addressed in \cite{yi2017syncspeccnn} by focusing on learning graph convolutions and using a graph spectral embedding via eigenvector decomposition.
In the context of spectral clustering, \cite{tian2014learning} learned an auto-encoder that maps the rows of a graph Laplacian matrix onto the corresponding spectral embedding. We build upon these approaches and leverage the graph structures of font glyphs for both cross-modal style translation and to represent font style and content.

\section{Problem Definition and Motivation}

\subsection{Font Completion}

Font completion is to transfer the font style of one or several glyphs to the others in a glyph set (see top of Figure~\ref{fig:task}). For the ease of discussion, assume we are given a single input glyph image $\bm{x}_{s,c^{\mathrm{in}}}$ of a specific font style $s$ and glyph content $c^{\mathrm{in}}$, and the font completion task requires completing the remaining glyphs in the glyph set $X_s=\{\bm{x}_{s,1}, \bm{x}_{s,2}, \cdots, \bm{x}_{s,C}\}$, where $C$ is the number of predefined glyphs (e.g., $C{=}26$ for Roman capital letters).
Taking $\bm{x}_{s,c^{\mathrm{in}}}$ as input, our model generates the remaining glyphs $\{ \hat{\bm{x}}_{s,c} | c \neq c^{\mathrm{in}} \}$ sharing the same font style $s$ as $\bm{x}_{s,c^{\mathrm{in}}}$.

The objective of our learning process is to obtain a model $f$ parameterized by $\theta$ that minimizes the difference between $\hat{\bm{x}}_{s,c}$ and $\bm{x}_{s,c}$ ($ c \in [1, C] \setminus \{c^{\mathrm{in}}\}$).
The overall formula of this problem is
\begin{equation}
\theta^* = \operatorname*{argmin}_\theta \sum_{s=1}^S \sum_{c^{\mathrm{in}}=1}^C \sum_{c \neq c^{\mathrm{in}}} d(\bm{x}_{s,c}, f(\bm{x}_{s,c^{\mathrm{in}}}, c; \theta)),\label{eq:font_completion_objective}
\end{equation}
where $d$ is the distance measure between two images, $s \in [1, S]$ and $c \in [1, C]$ are the style and content index of a glyph $\bm{x}_{s,c}$, respectively, and $S$ is the total number of fonts in training data.
We mainly use $\texttt{PSNR}$ (Peak Signal-to-Noise Ratio) and $\texttt{SSIM}$ (Structural SIMilarity) between predicted and ground truth images as evaluation metrics. With the above optimal $\theta^*$, we can obtain $\hat{\bm{x}}_{s,c} = f(\bm{x}_{s,c^{\mathrm{in}}}, c; \theta^*)$ as the prediction of $\bm{x}_{s,c}$.

Generating font glyph images with the same style could also be seen as an image-to-image translation problem, where the style feature from a reference image is extracted and combined with the content features of target images. Existing single image-to-image translation methods \cite{gatys2016image,huang2017arbitrary,li2017universal,li2018closed} only learn a single modality translation. However, as will be shown in Figure~\ref{fig:img2img_failure_case}, the visual quality of direct image-to-image translation is low when only a single input glyph is given. This motivates us to look for another font modality with which the intrinsic structure of font glyph can be better captured.

\subsection{Why Graph Font Representation is Necessary?} \label{sec:why_graph}

In font completion, the input image $\bm{x}_{s,c^{\mathrm{in}}}$ and target image $\bm{x}_{s,c}$ share the same font style but different glyph contents.
We can infer that model $f$ must learn meaningful style features of input glyphs and content features of target glyphs. Different from the image-to-image translation task, we have multiple modalities to choose from to represent the target glyph, which would be finally converted to glyph image $\hat{\bm{x}}_{s,c}$.
\ul{However, it is not immediately clear which font modality is the most suitable one for modeling.} To facilitate the learning process of Eq.~\ref{eq:font_completion_objective}, we therefore make the following key observations for the font completion task.

\noindent \textbf{Disentanglement and characterization of style and content features.} The model $f$ will need to extract style feature from an input glyph $\bm{x}_{s,c^{\mathrm{in}}}$, and combine the style feature with the content features of the target glyph $\bm{x}_{s,c}$. Thus, the font completion task will benefit from learning decoupled global content and local style features of glyphs. Another key property of font ignored by most previous work in font generation is that font glyphs are initially designed as vector objects defined by 2D curves and shapes, which means they have more intrinsic representation in geometric parameter space than the perceptual representation in the image space. By design, the geometric parameter space can better capture local patterns of glyph strokes, such as serif, terminal, bowl, and ear, which are the key to distinguish different font styles \cite{fontanatomy}.
Therefore, we expect a good font representation to disentangle content and style information, and also faithfully capture local style details.

\noindent \textbf{Scalability and manipulability of learned font representation.} The learned representation will be decoded or rendered into a rasterized font glyph image. Thus, for practical usage, we also need $f$ to learn a representation of good scalability, i.e., can be easily rendered to glyph images of various resolutions. Moreover, since human designers usually create new style designs by referring to existing ones, it is also beneficial to have good manipulability for the learned representation, i.e., can be easily modified by designers with interpretable operations.

Based on the above requirements for font completion, we compare different font representations available to us as summarized in Table~\ref{table:modalities}.
Specifically, we have three existing representations of font glyphs: SVG curves, rasterized images, and point set sampled along the contour (with coordinates and tangents). Each of these representation has its pros and cons: SVG vector fonts are scalable and resolution-free, but are not good candidates for representation learning as they cannot be uniquely expressed with Bezier curves; rasterized fonts are more perceptual and good for visual learning, but are limited by fixed resolution and therefore may not capture enough style details; sampled point set can be easily manipulated and converted from other formats, but lacks global structure information for glyph content.

\begin{table}[h!]
\caption{Comparison of different font modalities for the needs of font completion task.}
\centering
\footnotesize
\begin{tabular}{cccccc}
\toprule
Modality & \begin{tabular}[c]{@{}c@{}} Content \\ (global)\end{tabular} & \begin{tabular}[c]{@{}c@{}} Style \\ (local)\end{tabular} & Scalable & Manipulable \\ \midrule
SVG curves &  &  & \checkmark & \checkmark \\
Raster image & \checkmark &  &  &  \\
Point set &  & \checkmark & \checkmark & \checkmark \\
Graph (proposed) & \checkmark & \checkmark  & \checkmark & \checkmark \\
\bottomrule
\end{tabular}\label{table:modalities}
\end{table}

\subsection{Font in Graph: Decoupled Style/Content with Scalability \& Manipulation}

Motivated by the limitations of existing representations, we want to design a new font representation that can capture both local and global glyph appearance, scale to any resolution, and deform easily with the designer's manipulation. To this end, we propose a graph-based font glyph representation which serves as an intermediate stage to facilitate the font completion task.

In our graph formulation, we have a set of nodes corresponding to a sampled point set whose attributes capture the local font styles, and the edge connections between nodes define the global glyph structure.
As shown in Figure~\ref{cycle_architecture}, the graph has a hierarchical structure, consisting of primary nodes segmenting glyph strokes, secondary nodes describing local details within each stroke, and edges connecting the entire glyph contour. 
Each node has attributes containing both coordinates and tangents information so that the original vector curve segments can be fully reconstructed. These attributes also expose a user-friendly interface for manual editing. The graph can express any connection between strokes. When the graph is partially connected, it can also represent glyphs with disjoint strokes, e.g., the letter ``i''.

As summarized in the last row of Table~\ref{table:modalities}, our graph representation satisfies all the desired properties for font completion. It can be regarded as an intermediate state between vectorized SVG curves and a rasterized image. Moreover, with the recent advance of graph deep learning models such as Graph Transformer Networks \cite{graphtrans_nips2019} and Graph Convolutional Networks \cite{gcn_2018modeling}, the graph representation is also computationally tractable to integrate with other deep learning modules.
With more intrinsic font glyph information represented in our graph, we will show how it can help regularize the conventional image style transfer problem with domain knowledge from a complementary modality, and facilitate tasks including font completion, interpolation, and manipulation (see Figure~\ref{fig:task}).

\section{Font Completion: Cycling between Multi-modality Representations} \label{sec:framework}

\begin{figure}[ht!]
\begin{center}
    \includegraphics[width=1\linewidth]{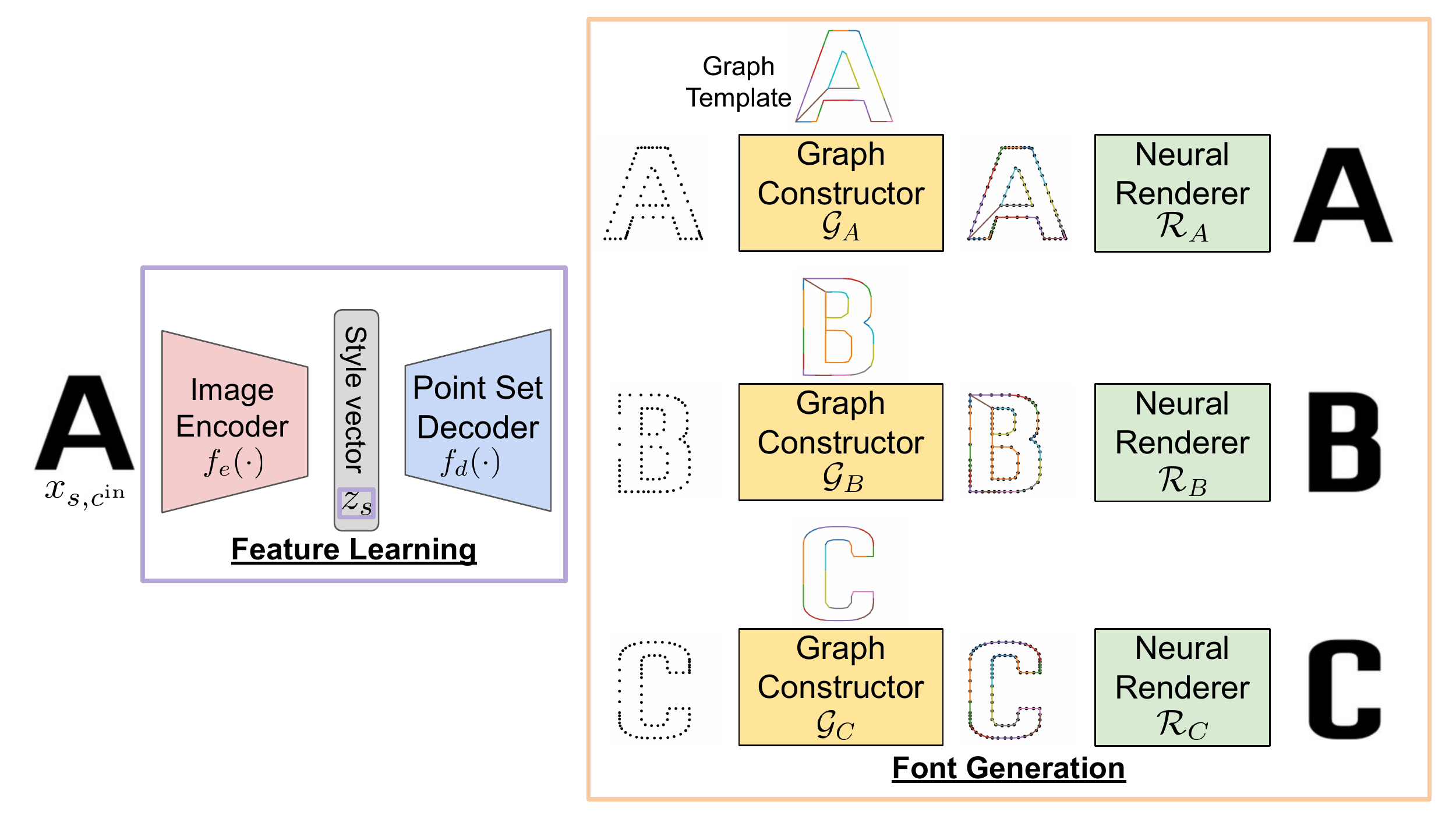}
\end{center}
\caption{Our image-to-graph-to-image method for font completion. Given a reference source glyph, we will generate all unseen glyphs of the same font style. To leverage multi-modal font representations, we design a cross-modality encoder-decoder to transit between different representations. The image encoder first extracts the font style vector from the source glyph. Then a point-set decoder together with a graph constructor synthesizes the stylized target glyph in its graph representation, which is further converted to pixel output by a graph transformer-based image renderer. }
\label{fig:completion_model_structure}
\end{figure}

\noindent \textbf{Framework Design Philosophy}
Given an input image $ \bm{x}_{s,c^{\mathrm{in}}} $ of glyph $ c^{\mathrm{in}} $ with style label $s$, the goal of font completion is to generate glyph images $ \bm{x}_{s,c} $ for any other glyph $c$ of the same font style $s$.
To leverage the graph-based intermediate representation (Section \ref{sec:why_graph}), we decompose the overall framework into two stages: 1) the input glyph image is transformed to glyph graph representations, and 2) the graph feature is rendered as completed glyph images with the transferred font style. In short, the intermediate graph representation connects the input and the glyphs to be completed, and regulates the transferring process.

As shown in Figure~\ref{fig:completion_model_structure}, our deep network model $f$ consists of four integral components: an image encoder $f_e$, a point set decoder $f_d$, a graph constructor $\mathcal{G}$, and a neural renderer $\mathcal{R}$, which gives
\begin{equation}
\label{eq:framework}
    \hat{\bm{x}}_{s,c} = f(\bm{x}_{s,c^{\mathrm{in}}}, c; \theta)
    = \mathcal{R} \circ \mathcal{G} \circ f_d  \circ f_e(\bm{x}_{s,c^{\mathrm{in}}}, c; \theta) .
\end{equation}
We first extract font style feature with image encoder $f_e$, based on which a number of sampled points of target glyph are generated with point set decoder $f_d$. The discrete point set is further mapped to graph representation with graph constructor $\mathcal{G}$, which is finally converted to a rasterized image with neural renderer $\mathcal{R}$.
In this way, we establish the two-way connection between the image modality and graph modality of font glyphs. The graph representation serves as an intermediate representation to efficiently capture the structure of glyph and the style of font.

\noindent \textbf{Image encoder $f_e$.}
The image encoder $f_e$ extracts a latent vector $\bm{z}_s$ from a source glyph image $\bm{x}_{s,c^{\mathrm{in}}}$ that represents the common style shared all the glyphs of style $s$:
\begin{equation}
    \bm{z}_s = f_e(\bm{x}_{s,c^{\mathrm{in}}}; \theta_e) \label{eq:v_s},
\end{equation}
where $f_e$ is designed as a convolutional network with parameter $\theta_e$.
The objective of learning this image encoder is to disentangle font style from glyph content,
and faithfully capture local style details.
To make the learned feature distinctive of font style, we apply a linear classifier $h$ on $\bm{z}_s$, which guides feature learning with a font classification loss:
\begin{align}
    \mathcal{L}_\mathrm{cls} &
    = - \mathop{\mathbb{E}}_{s,c} \left[ \mathrm{log} p(s|\bm{x}_{s,c}) \right] \\
    & = - \mathop{\mathbb{E}}_{s,c} \left[ \mathrm{log} h_s \circ f_e(\bm{x}_{s, c}) \right] ,
\label{eq:loss_encoder_classification}
\end{align}
where $h_s$ is the normalized classification scoring function for style class $s$.

\noindent \textbf{Point set decoder $f_d$.}
Instead of directly generating image or graph of target glyph $c$ from the style feature $\bm{z}_s$, we first employ a point set decoder $f_d$ to output a fixed number of $m$ sample points on the contour of target glyph $c$:
\begin{equation}
    \mathbf{P}_{s,c} = f_d(\bm{z}_s, c; \theta_d),
\end{equation}
where $\mathbf{P}_{s,c}{=}[\bm{p}_1, \cdots, \bm{p}_m]$, and a point feature $\bm{p}_i$ of dimension $p$ contains attribute values such as coordinates and tangents.
$f_d$ is designed as a multi-layer 1D convolutional network with parameter $\theta_d$. The generated points are densely distributed over the entire glyph contour, and therefore can serve as candidate node locations for graph construction in the next step. Such a design reduces the complexity of graph constructor which otherwise has to build the entire graph from raw style feature.

\noindent \textbf{Graph constructor $\mathcal{G}$.}
The goal of a graph constructor $\mathcal{G}$ is to select graph nodes from the point set $\mathbf{P}_{s,c}$ and predict their connectivity. Suppose there are $n_1$ primary nodes each of which is associated with $n_2$ secondary nodes, we then have $n {=} n_1{\times}n_2$ actual points to be mapped from the $m$ points in $\mathbf{P}_{s,c}$, assuming $m{\geq}n$. The mapping is represented by a $m{\times}n$ binary matrix $\mathbf{M}$. As the $n_2$ secondary nodes associated with each primary node are defined to be sequentially connected, the order of the secondary nodes within each stroke is implied by the mapping matrix $\mathbf{M}$. The remaining topology variation from the primary nodes can be modeled by a $n_1{\times}n_1$ adjacency matrix $\mathbf{A}$, with $A(i,j){=}1$ indicating a directed edge is presented from primary node $i$ to $j$ (i.e. their corresponding curve segments are connected on the contour), and $A(i,j){=}0$ otherwise. Note directed graph is necessary to model the closed loop of the inner and outer outline of a stroke. When there are two primary nodes on an outline loop, they are connected with a bi-directional edge $A(i,j){=}A(j,i){=}1$. When there is only one primary node on an outline loop, a self-linkage $A(i,i){=}1$ is added.

Now we define a directed graph $\bm{g}_{s,c}$ as the graph representation of glyph $\bm{x}_{s,c}$, whose structure is predicted with graph constructor $\mathcal{G}$:
\begin{equation}
\label{eq:g_sc}
    \bm{g}_{s,c} = (\mathbf{M}_{s,c}, \mathbf{A}_{s,c}) \approx \mathcal{G}(\mathbf{P}_{s,c}; \theta_g, \phi_c),
\end{equation}
where $\mathbf{M}_{s,c}$ and $\mathbf{A}_{s,c}$ are the mapping and adjacency matrices for glyph $\bm{x}_{s,c}$. The model $\mathcal{G}$ has a backbone network with parameter $\theta_g$ shared by all the glyphs, and two head networks for each glyph predicting $\mathbf{M}_{s,c}$ and $\mathbf{A}_{s,c}$ with parameter $\phi_c$.
The shared backbone network is a single layer Transformer \cite{transformer_nips2017} considering the non-local relationship among points $\mathbf{P}_{s,c}$. The Transformer maps $\mathbf{P}_{s,c}$ of size $m{\times}p$ to a feature matrix $\mathbf{Q}_{s,c}$ of size $m{\times}n$, which is fed into the two head networks.
The mapping matrix head consists of $n$ parallel linear layers followed by softmax operation over $m$-dimensional outputs. The adjacency matrix head applies average pooling of size $n_2$ along the column of $\mathbf{Q}_{s,c}$, followed by a multilayer perceptron (MLP) and a sigmoid activation.
Both of the two predicted matrices $\hat{\mathbf{M}}_{s,c}$ and $\hat{\mathbf{A}}_{s,c}$ of the inferred graph $\hat{\bm{g}}_{s,c}$ are probabilistic version of their binary format. To learn the graph constructor, we use point reconstruction loss $\mathcal{L}_\mathrm{rec}$ for the mapped points and log likelihood loss $\mathcal{L}_\mathrm{adj}$ for all the entries of $\mathbf{A}_{s,c}$:
\begin{align}
    \mathcal{L}_\mathrm{rec} = & \mathop{\mathbb{E}}_{s,c} \left[ \norm{ \hat{\mathbf{M}}_{s,c}^T \mathbf{P}_{s,c} - \mathbf{N}_{s,c}}_2
    \right] \label{eq:loss_rec}\\
    \mathcal{L}_\mathrm{adj} = & - \Sigma_{ij}\mathop{\mathbb{E}}_{s,c} \left[ A_{s,c}(i,j)\mathrm{log}\hat{A}_{s,c}(i,j) \nonumber \right. \\
    & \left. + (1{-}A_{s,c}(i,j))\mathrm{log}(1{-}\hat{A}_{s,c}(i,j)) \right] ,
\end{align}
where $\mathbf{N}_{s,c}$ denotes the aligned node points on the ground truth graph $\bm{g}_{s,c}$. We cannot apply loss on $\hat{\mathbf{M}}_{s,c}$ since it is an operation on the point set $\mathbf{P}_{s,c}$ with no predefined order.

\noindent \textbf{Neural renderer $\mathcal{R}$.}
Lastly, the rasterized image of target glyph can be generated by a neural renderer $\mathcal{R}$ conditioned on the predicted graph $\hat{\bm{g}}_{s,c}$:
\begin{equation}
    \hat{\bm{x}}_{s,c} = \mathcal{R}(\hat{\bm{g}}_{s,c}; \theta_{r,c}). \label{eq:render}
\end{equation}
The neural renderer $\mathcal{R}$ is designed as a multi-graph transformer network~\cite{multigraph_arxiv2019} with glyph-specific parameter $\theta_{r,c}$. The adjacency matrix $\hat{\mathbf{A}}_{s,c}$ defines a global graph of primary nodes and the sequentially connected secondary nodes implied by $\hat{\mathbf{M}}_{s,c}$ produce a local graph for each glyph stroke. The graph attention enforced on different scales helps to generate glyph images with a smooth global structure as well as detailed local patterns.
The neural renderer is trained by minimizing the pixel-wise MSE loss:
\begin{equation}
\label{eq:loss_img}
    \mathcal{L}_\mathrm{img} = \mathop{\mathbb{E}}_{s,c} \left[ \norm{\hat{\bm{x}}_{s,c} - \bm{x}_{s,c}}_2 \right].
\end{equation}
Note that it is also possible to use a conventional graphics renderer as a surrogate for the neural renderer by fitting smooth Bezier curves to vectorize the outline connecting the nodes in $\hat{\bm{g}}_{s,c}$.

\noindent \textbf{Model training}
There are multiple components ($f_e$, $f_d$, $\mathcal{G}$, $\mathcal{R}$) in the overall font completion model $f$, and some components even have individual parameter instance for each glyph. Therefore, training all the model parameters jointly is computationally prohibitive. In practice, we first pre-train $f_e$ with $\mathcal{L}_\mathrm{cls}$ loss. Then we jointly train $f_e$, $f_d$ and $\mathcal{G}$ with additional losses of $\mathcal{L}_\mathrm{rec}$ and $\mathcal{L}_\mathrm{adj}$. Finally, $\mathcal{R}$ is independently trained with loss $\mathcal{L}_\mathrm{img}$ for each glyph $c$. During the training of $\mathcal{G}$, we can load and optimize a subset of glyph-specific parameters $\phi_c$ at each time to make the memory usage independent of glyph set size. This training strategy works well in our experiments and is scalable to newly added glyphs.

\section{Experiments}

\subsection{Dataset} \label{sec:dataset}

All our experiments are conducted on the Google Fonts dataset \cite{googlefonts}, and we follow similar pre-processing steps as in~\cite{lopes2019learned} to filter out bad glyphs, render images and extract font metadata. Glyphs with irregular designs such as symbols and ornaments are removed to simplify our graph template design (see below). We consider all the 26 Roman capital letters in the experiments. In total, we have 2693/55554 valid fonts/glyphs.
We use 95\% fonts for train/validation and 5\% fonts for testing.

\noindent \textbf{Graph Template.}
The image, SVG curves, and sample points of a glyph can be obtained using standard procedures. To train our model, the ground truth for the graph modality $\bm{g}_{s,c}$ still needs to be created.

For each glyph, we design an automatic procedure to build templates for all font styles which specifies the number of primary nodes $n_1$, the number of corresponding secondary nodes $n_2$, and the connections among all the primary nodes $\mathbf{A}_c$.
Following~\cite{smirnov2020deep}, we decompose each glyph into three levels of hierarchy: contour, stroke, and node. Each stroke corresponds to a primary node and the nodes on the stroke are secondary nodes. All the primary nodes on the same contour loop are sequentially connected in the counterclockwise direction. The primary nodes of different contours are not connected.
We keep the same total number of nodes as 150 for all the glyphs and distribute the nodes to strokes according to the number of contours and the relative curve length from their vector representations.

To obtain the node attribute matrix $\mathbf{N}_{s,c}$, we need to align all the nodes from the graph template for $c$ to the SVG curves of the glyph, and then calculate the exact coordinates and tangents. The same node should be aligned to the corresponding position of all glyphs $c$ with different styles $s$.
To do that, we first use the keypoint detection model of~\cite{smirnov2020deep} to roughly segment all the strokes based on the glyph image. The nearest points on SVG curves to those detected points are used as the locations of primary nodes. Then all the secondary nodes are found by uniformly sampling SVG curves between each pair of connected primary nodes.
Figure~\ref{graph_construction} shows all the font glyph modalities in our dataset.

\begin{figure}[th]
\begin{center}
    \includegraphics[width=\linewidth, trim=0 30 0 0, clip=true]{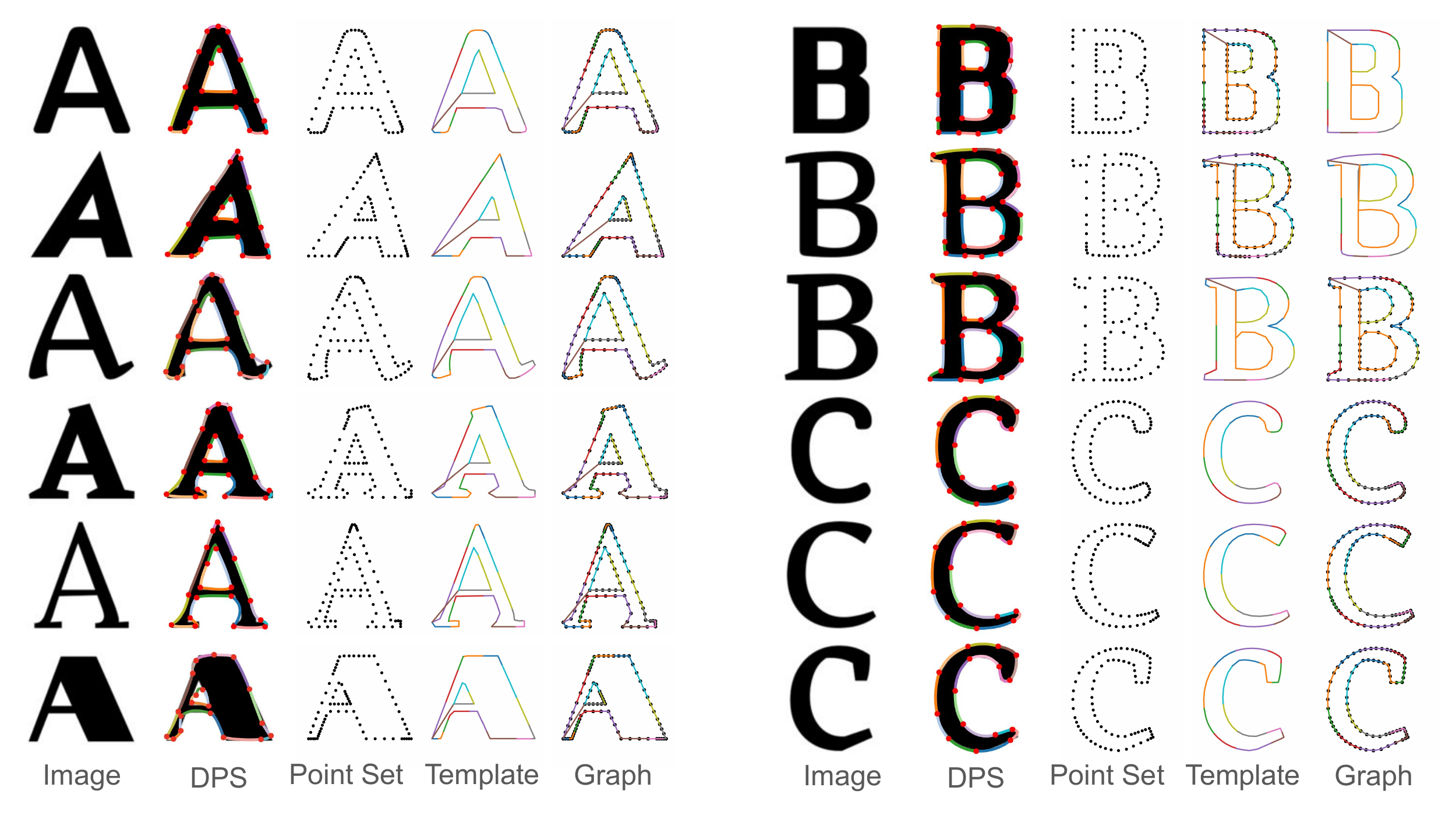}
\end{center}
\caption{Different data modalities of font glyphs. For each example from left to right: image, keypoints, point set, graph template and directed graph representation.}
\label{graph_construction}
\end{figure}

\subsection{Implementations}
\noindent \textbf{Encoder/Decoder Learning.}
The image encoder $f_e$ takes a $128{\times}128$ grayscale image as input and encodes a latent vector $\bm{z}_s{\in}\mathds{R}^{128}$ using five Conv2d-BatchNorm-ReLU layers. A linear mapping with sigmoid function is used for font classifier $h$ in Eq.~\ref{eq:loss_encoder_classification}. The point set decoder $f_d$ uses four glyph-conditional TranposeConv1D-BatchNorm-ReLU layers with the first layer shared by all glyphs.

The image-to-image baseline (Table~\ref{table:results_ablation} and Figure~\ref{fig:img2img_failure_case}) has the same encoder structure as $f_e$, and its decoder uses five glyph-conditional TranposeConv2D-BatchNorm-ReLU layers with the first layer shared by all glyphs. The last layer is followed by a sigmoid function which generates an output image with the same size as input.

\noindent \textbf{Graph Constructor.}
We define graph node attributes $\mathbf{N}_{s,c}$ with 2-dim coordinates and 2-dim tangents of the corresponding points on SVG curves. The adjacency matrix $\mathbf{A}_{s,c}$ reduces to $\mathbf{A}_c$ in our case as the same graph template is used for any given $c$. In this way, the loss term $\mathcal{L}_\mathrm{adj}$ is not activated.
The mapping $\mathbf{M}_{s,c}$ is still learned to align point set with graph template, from which the relative position of a node on the stroke/contour can be determined.

\noindent \textbf{Neural Renderer.}
The neural renderer takes the node attributes $\mathbf{N}_{s,c}$ (coordinates and tangents) and graph structure $\mathbf{A}_c$ (adjacency matrix) as input. The coordinates and tangents are translated to a 256-dim feature through a linear function. As $\mathbf{A}_c$ is fixed for any given glyph, we use a dictionary lookup table to implement the graph attentions over strokes and contours defined by $\mathbf{A}_c$, which results in another 256-dim feature.
The node feature and edge feature are concatenated and fed into a four-layer Conv1D network to produce a 128-dim embedding, which further passes through five TranposeConv2D layers to reconstruct the output image.

We train all the models on a single GeForce RTX 2080 Ti GPU for 100 epochs, using Adam with learning rate $1{\times}10^{-3}$, betas $(0.9, 0.999)$ and batch size $64$.

\subsection{Results}

\subsubsection{Font Completion}

\noindent \textbf{Quantitative Measure.} We first quantitatively validate the contribution from our proposed graph font representation and $\mathcal{L}_\mathrm{cls}$.
Specifically, we calculate $\mathrm{MSE}$, $\mathrm{PSNR}$ and $\mathrm{SSIM}$ against the ground truth glyph images, as listed in Table~\ref{table:results_ablation}. By optimizing the graph reconstruction loss, our method outperforms image-to-image translation by 35.3\%, 29.4\%, 8.8\% on $\mathrm{MSE}$, $\mathrm{PSNR}$ and $\mathrm{SSIM}$, respectively.
This validates the \textbf{necessity of introducing our intermediate font graph representation}.
The performance can be further improved by $\mathcal{L}_\mathrm{cls}$, i.e., with the learned font style feature disentangled from glyph content.

\begin{table}[ht!]
\centering
\footnotesize
\caption{Quantitative evaluation of font completion task for img2img baseline and our graph based method using different training losses.
$\mathcal{L}_\mathrm{img}$ for img2img is the image reconstruction loss defined similarly as our renderer loss in Eq.~\ref{eq:loss_img}.
$\mathcal{L}_\mathrm{rec}$ is our graph reconstruction loss in Eq.~\ref{eq:loss_rec}. The classification loss $\mathcal{L}_\mathrm{cls}$ in Eq.~\ref{eq:loss_encoder_classification} makes the learned style feature more distinctive of font style.
}\vspace{-0.5em}
\begin{tabular}{ccccc}
\toprule
Method & Loss  & $\mathrm{MSE} \downarrow$ & $\mathrm{PSNR} \uparrow$ & $\mathrm{SSIM} \uparrow$ \\ \midrule
img2img & $\mathcal{L}_\mathrm{img}$  & 0.17 & 8.14 & 0.57 \\
\midrule
\multirow{2}{*}{\makecell{img2\textbf{graph}2img \\ (ours)}}
 &  $\mathcal{L}_\mathrm{rec}$  & 0.11 & 10.53 & 0.62 \\
 &  $\mathcal{L}_\mathrm{rec}$ + $\mathcal{L}_\mathrm{cls}$ & \textbf{0.10} & \textbf{10.90} & 0.62 \\
\bottomrule
\end{tabular}
\label{table:results_ablation}
\end{table}

\begin{figure}[!ht]
\begin{subfigure}[b]{0.48\textwidth}
    \includegraphics[width=\textwidth]{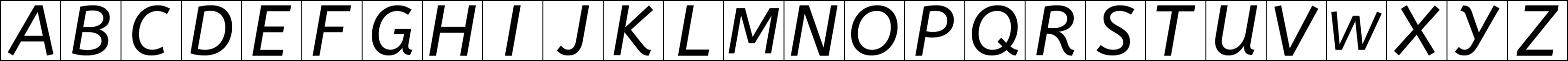}
    \includegraphics[width=\textwidth]{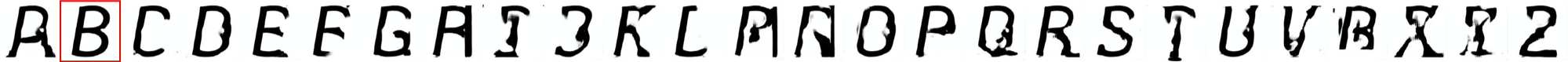}
    \includegraphics[width=\textwidth]{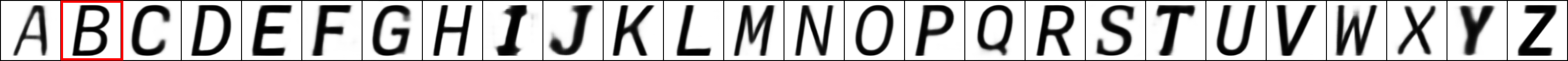}
    \caption{ABeeZee-Italic}
\end{subfigure}
\begin{subfigure}[b]{0.48\textwidth}
    \includegraphics[width=\textwidth]{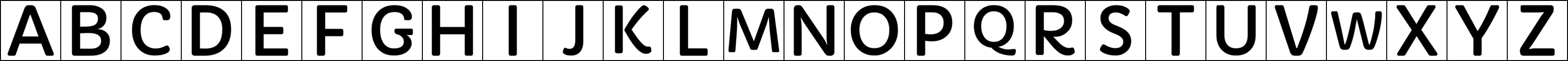}
    \includegraphics[width=\textwidth]{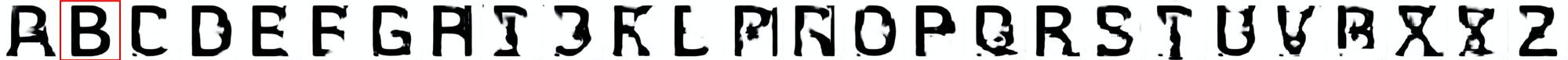}
    \includegraphics[width=\textwidth]{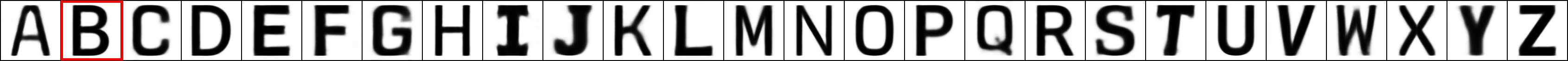}
    \caption{Capriola-Regular}
\end{subfigure}
\begin{subfigure}[b]{0.48\textwidth}
    \includegraphics[width=\textwidth]{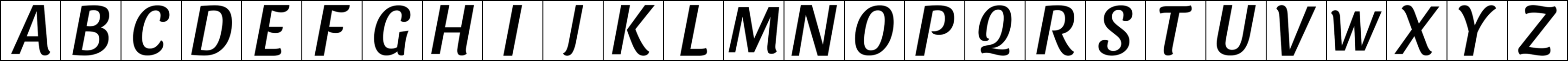}
    \includegraphics[width=\textwidth]{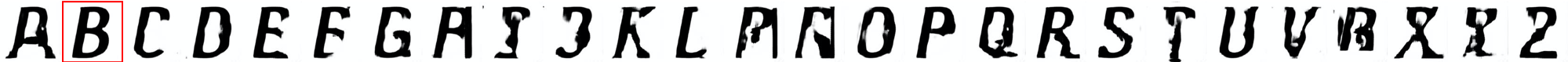}
    \includegraphics[width=\textwidth]{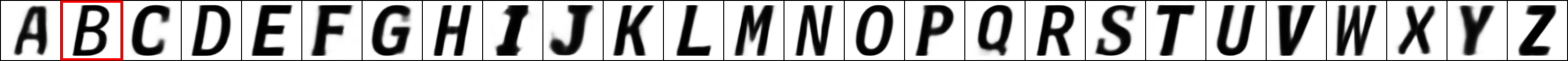}
    \caption{Sansita-Italic}
\end{subfigure}
\begin{subfigure}[b]{0.48\textwidth}
    \includegraphics[width=\textwidth]{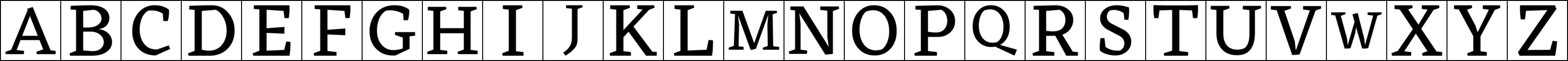}
    \includegraphics[width=\textwidth]{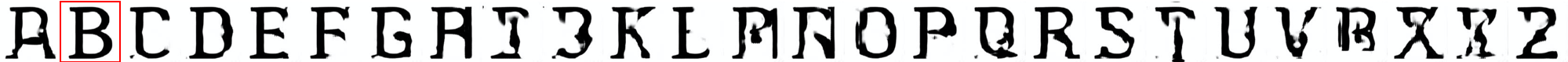}
    \includegraphics[width=\textwidth]{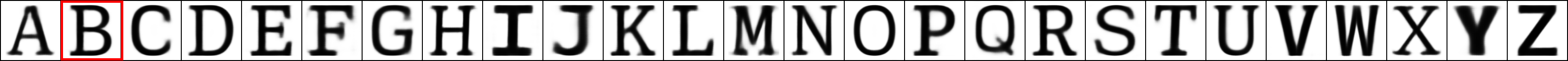}
    \caption{Eczar-Regular}
\end{subfigure}
\begin{subfigure}[b]{0.48\textwidth}
    \includegraphics[width=\textwidth]{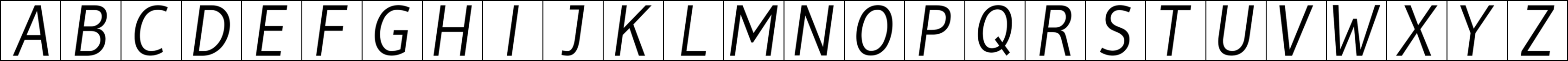}
    \includegraphics[width=\textwidth]{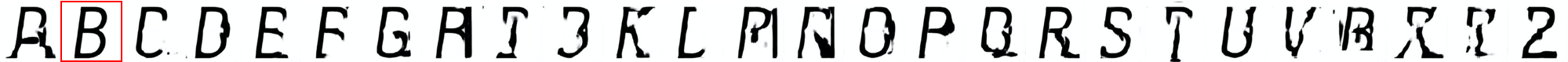}
    \includegraphics[width=\textwidth]{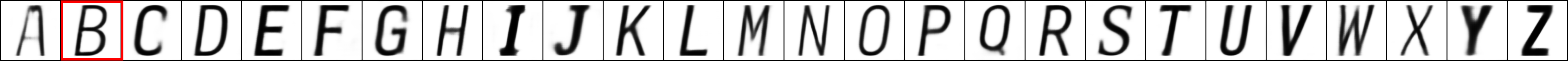}
    \caption{Gudea-Italic}
\end{subfigure}
\caption{Font completion visualization. Three rows in each style panel: ground truth glyphs (top), results of \textbf{TCN} \cite{park2018typeface} (middle), and our image-to-graph-to-image approach (bottom). Red boxes indicate input glyphs. Our method completes glyphs with clear appearance and consistent style with the target glyphs.
}
\label{fig:img2img_failure_case}
\end{figure}

\noindent \textbf{Visual Quality.} As demonstrated in Figure~\ref{fig:img2img_failure_case} (one style for each panel), comparing with previous \textbf{TCN} \cite{park2018typeface} method,
our method (the third row in each panel) produces better completed glyphs with both clear appearance and consistent style with the target glyphs (first row in each panel), even if the targets are unseen during training. In contrast, TCN suffers from serious blur, artifacts, broken glyph structure, and fails to maintain style information from the input glyph.

\subsubsection{Font Manipulation} \label{sec:manipulation}

The goal of font manipulation is to make small modifications to the font style of a source glyph, and transfer the changes to other glyphs of this font. Considering that modifying several points on a graph is much easier than changing a large number of pixels on an image, our proposed graph representation and graph constructor can serve as powerful tools for human designers to easily change typography properties like serif, italic, and weight.

After modifying an original graph $\bm{g}_{s,c^{in}}$ to a modified one $\Tilde{\bm{g}}_{s,c^{in}}$, we can transfer the changes to another glyph image $\Tilde{\bm{x}}_{s,c}$ by inferring an updated style feature $\Tilde{\bm{z}}_s$ with either forward or backward propagation:
\begin{align}
    \Tilde{\bm{z}}_s &= f_e(\mathcal{R}(\Tilde{\bm{g}}_{s,c^{in}}; \theta_{r,c^{in}}); \theta_e), \label{eq:manipulate0} \\
    \Tilde{\bm{z}}_s &= \operatorname*{argmin}_{\bm{z}_s} \norm{\mathbf{M}^T_{s,c^{in}} f_d(\bm{z}_s, c^{in}; \theta_d) - \Tilde{\mathbf{N}}_{s,c^{in}}}_2.
    \label{eq:manipulate}
\end{align}
In the back propagation of Eq.~\ref{eq:manipulate}, we assume the manipulated graph $\Tilde{\bm{g}}_{s,c}$ keeps the same node mapping relation $\mathbf{M}_{s,c^{in}}$ and only updates the node attributes $\Tilde{\mathbf{N}}_{s,c^{in}}$.
In our experiment, the optimization is solved by Adam optimizer with learning rate $10^{-4}$ for $10^4$ steps.

\begin{figure}[!h]
\centering
    \includegraphics[width=0.5\textwidth]{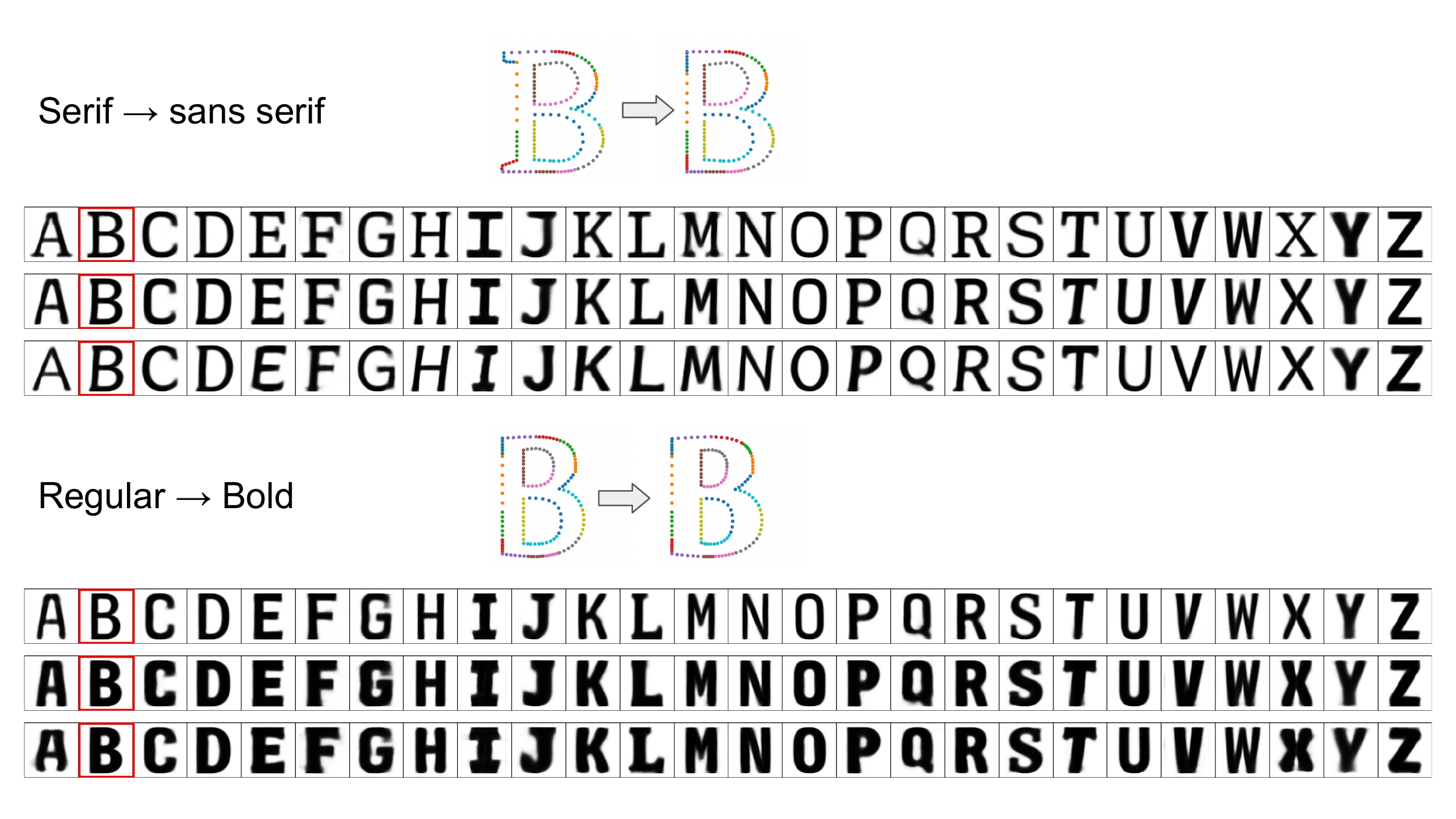}
    \includegraphics[width=0.5\textwidth]{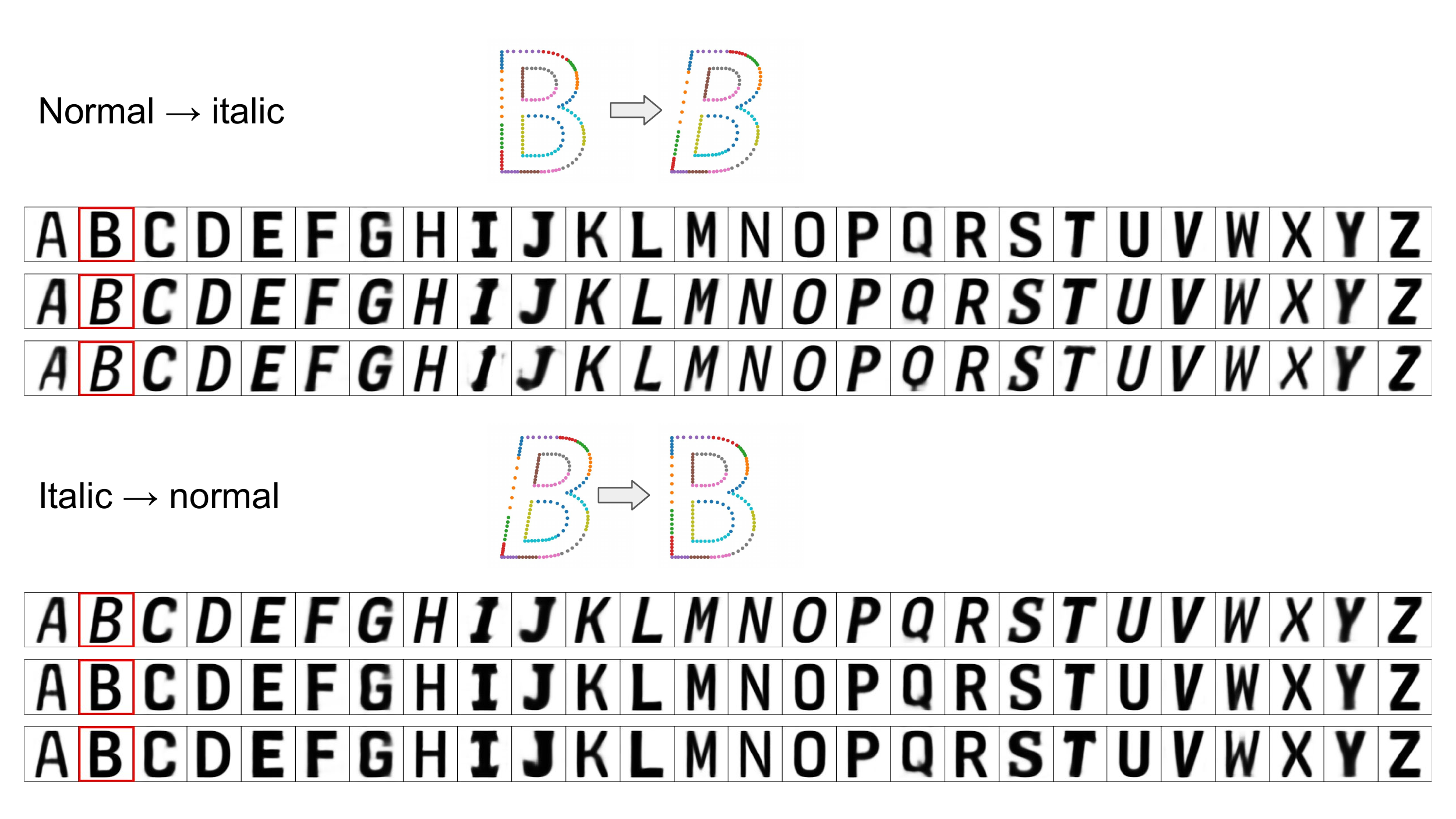}
\caption{Font manipulation results for editing properties of serif, italic and weight. Three rows in each style panel: ground truth before manipulation (top), forward-propagation results (middle), back-propagation results (bottom). Red boxes indicate input glyphs.}
\label{fig:font_manipulation_res}
\end{figure}

The updated style feature $\Tilde{\bm{z}}_s$ will then be used to complete the glyph set (Eq.~\ref{eq:g_sc} and Eq.~\ref{eq:render}). As illustrated in Figure~\ref{fig:font_manipulation_res}, both forward and backward methods can transfer the manipulated font style to other glyphs successfully. This demonstrates the capability of the proposed graph representation to capture even small local style variations, and also to effectively transfer the style changes to image modality.

\subsubsection{Font Interpolation} \label{sec:interpolation}

\begin{figure}[h]
\begin{center}
    \includegraphics[width=\linewidth]{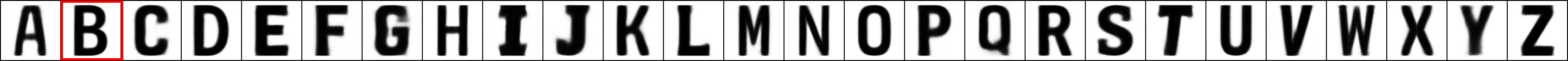}
    \includegraphics[width=\linewidth]{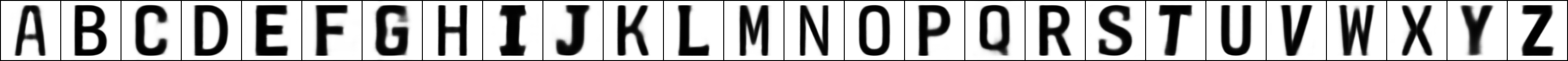}
    \includegraphics[width=\linewidth]{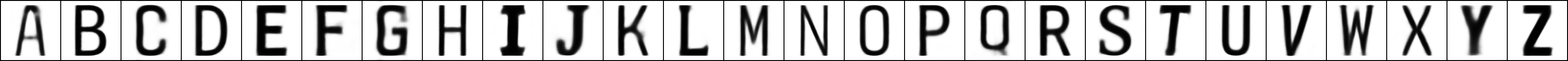}
    \includegraphics[width=\linewidth]{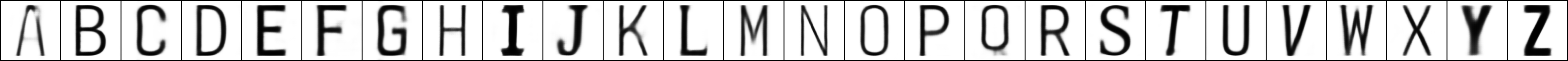}
    \includegraphics[width=\linewidth]{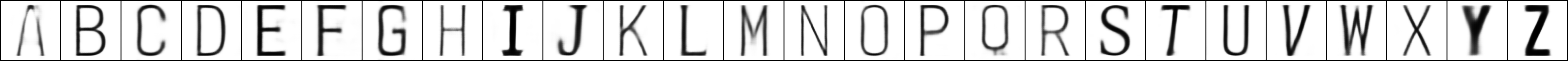}
    \includegraphics[width=\linewidth]{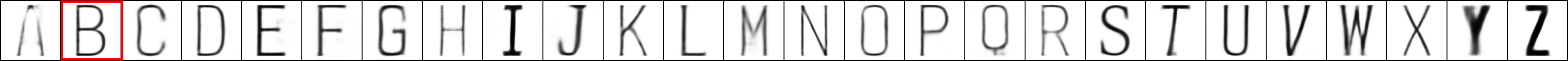}
\end{center}
\caption{Font interpolation results. The input glyphs $\bm{x}_{{s_1},c^{\mathrm{in}}}$ and $\bm{x}_{{s_2},c^{\mathrm{in}}}$ are highlighted by red boxes in the top and bottom rows. The middle rows are interpolated with $\lambda=0.8, 0.6, 0.4, 0.2$ (from top to bottom). The top and bottom rows also show font completion results from the input glyphs.}
\label{fig:font_interpolation_res}
\end{figure}

Font interpolation is performed to validate the effectiveness of the learned latent feature $\bm{z}_s$ in controlling glyph styles.
Given two input glyphs $\bm{x}_{{s_1},c^{\mathrm{in}}}$ and $\bm{x}_{{s_2},c^{\mathrm{in}}}$ of the same content but different styles, we first extract their font features $\bm{z}_{s_1}$ and $\bm{z}_{s_2}$ with $f_e$. With linear interpolation between the two features $\lambda \bm{z}_{s_1} + (1{-}\lambda) \bm{z}_{s_2}$ for $\lambda \in [0,1]$, we can generate all the glyph set for each interpolation coefficient $\lambda$ as shown in Figure~\ref{fig:font_interpolation_res}.
The smooth transition from one style to another demonstrates that our proposed graph representation can effectively capture local glyph styles, and the linear transition in the latent feature space can also be transferred to the image space.

\section{Conclusion}
In this work, motivated by the limitation of image modality for font generation tasks, we proposed a new graph representation for font glyphs. To better capture both global content and local details, we proposed a cross-modality auto-encoder framework to leverage the graphs as an intermediate representation for conversion between different font modalities. We build the first graph representation dataset for font glyphs by transforming SVG curves and point sets to hierarchical graphs.
Our method achieves significant performance improvements on font completion in both visual quality and quantitative metrics, compared with the previous image-to-image translation methods.
Our graph representation also exhibits high scalability and convenience for manual manipulation, which was demonstrated in font manipulation and interpolation tasks. In summary, our graph representation and cross-modality framework provide the font community with a new learning strategy as well as a new benchmark.


%

\appendices

\section{Graph Templates of All Glyphs}

As listed in Table~\ref{tab:graphdesign}, we keep the same total number of nodes for all the glyphs and distribute the nodes to strokes according to the number of contours and the relative curve length from their vector representations. We also includes all designed templates in Fig.~\ref{graph_construction}. The templates are consistent among different font styles.

\begin{table}[h]
\centering
\footnotesize
\caption{Decomposition of contours, strokes and nodes in our designed graph templates for all the Roman capital letters.}\vspace{-0.5em}
\begin{tabular}{ccccc}
\toprule
 Glyphs & Contours & \makecell{\#strokes \\ per contour} & \makecell{\#nodes \\ per stroke} & \makecell{Total \\ \#nodes}\\ \midrule
A,P,R & \makecell{outer \\ inner} & \makecell{15 \\ 3} & \makecell{8 \\ 10} & 150 \\ \midrule
D,O,Q & \makecell{outer \\ inner} & \makecell{15 \\ 4} & \makecell{6 \\ 15} & 150 \\ \midrule
B & \makecell{outer \\ inner upper \\ inner bottom} & \makecell{15 \\ 3 \\ 3} & \makecell{6 \\ 10 \\ 10} & 150 \\ \midrule
others & outer & 15 & 10 & 150 \\ \bottomrule
\end{tabular}
\label{tab:graphdesign}
\end{table}

\section{Dataset Statistics}\vspace{-1em}
\begin{figure}[h!]
\begin{center}
\vspace{-0.5em}
    \includegraphics[width=0.9\linewidth]{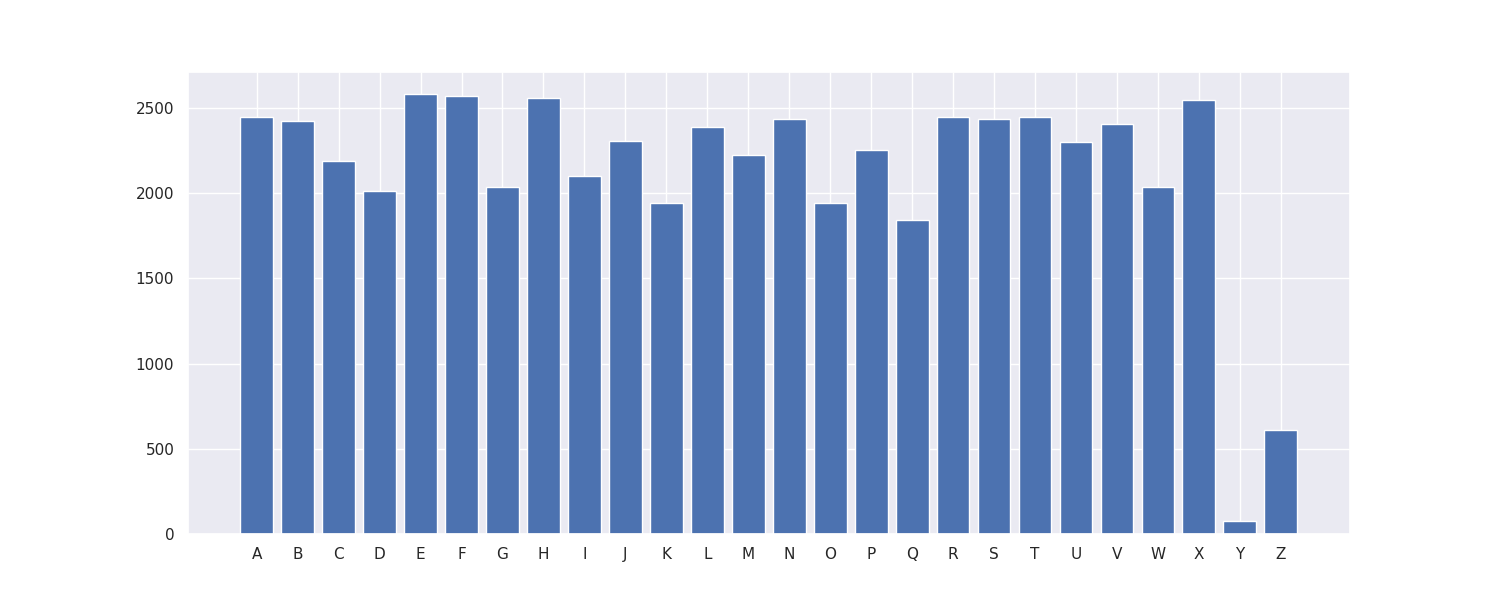}
    \includegraphics[width=0.25\linewidth]{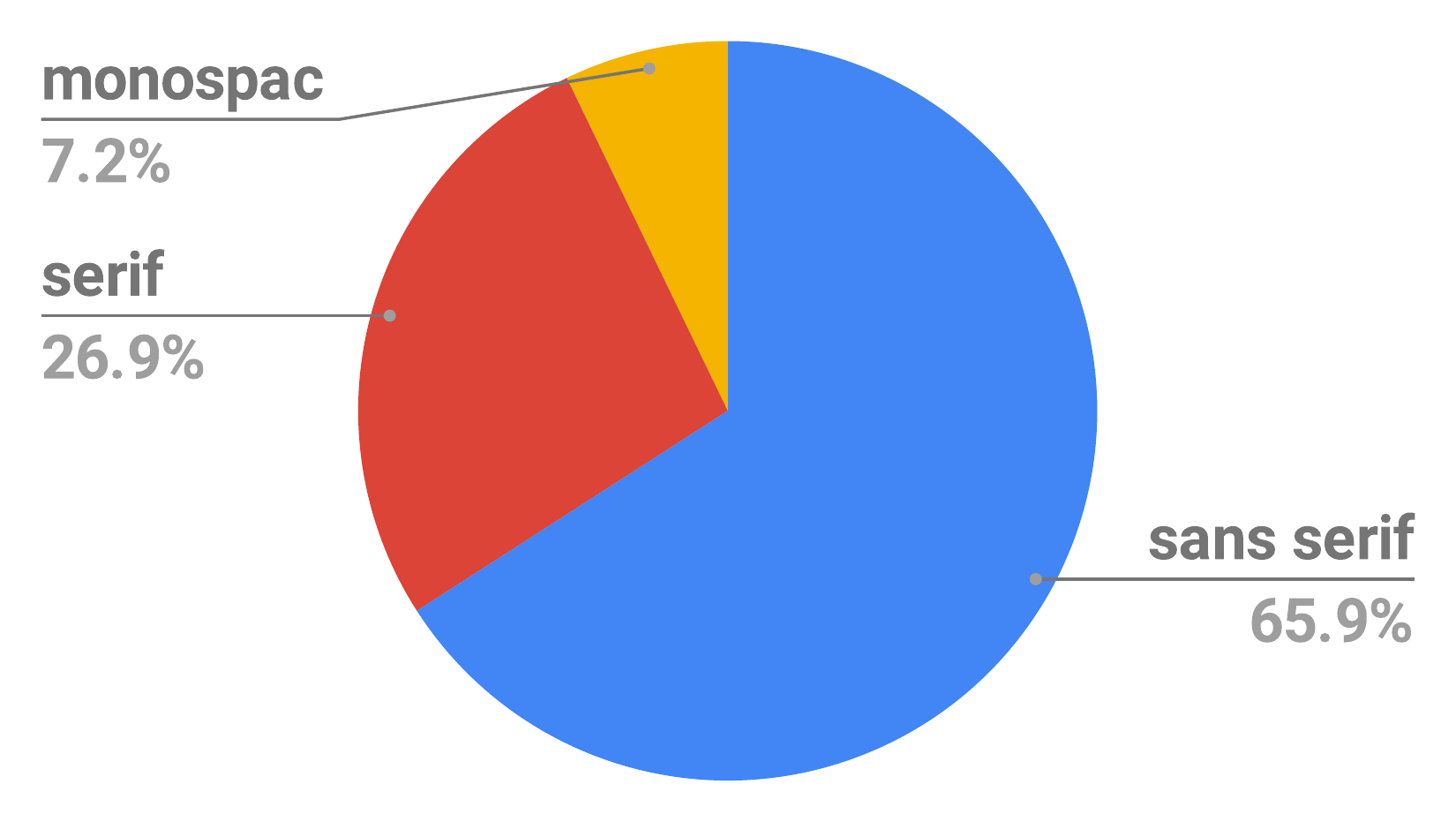}
    \includegraphics[width=0.25\linewidth]{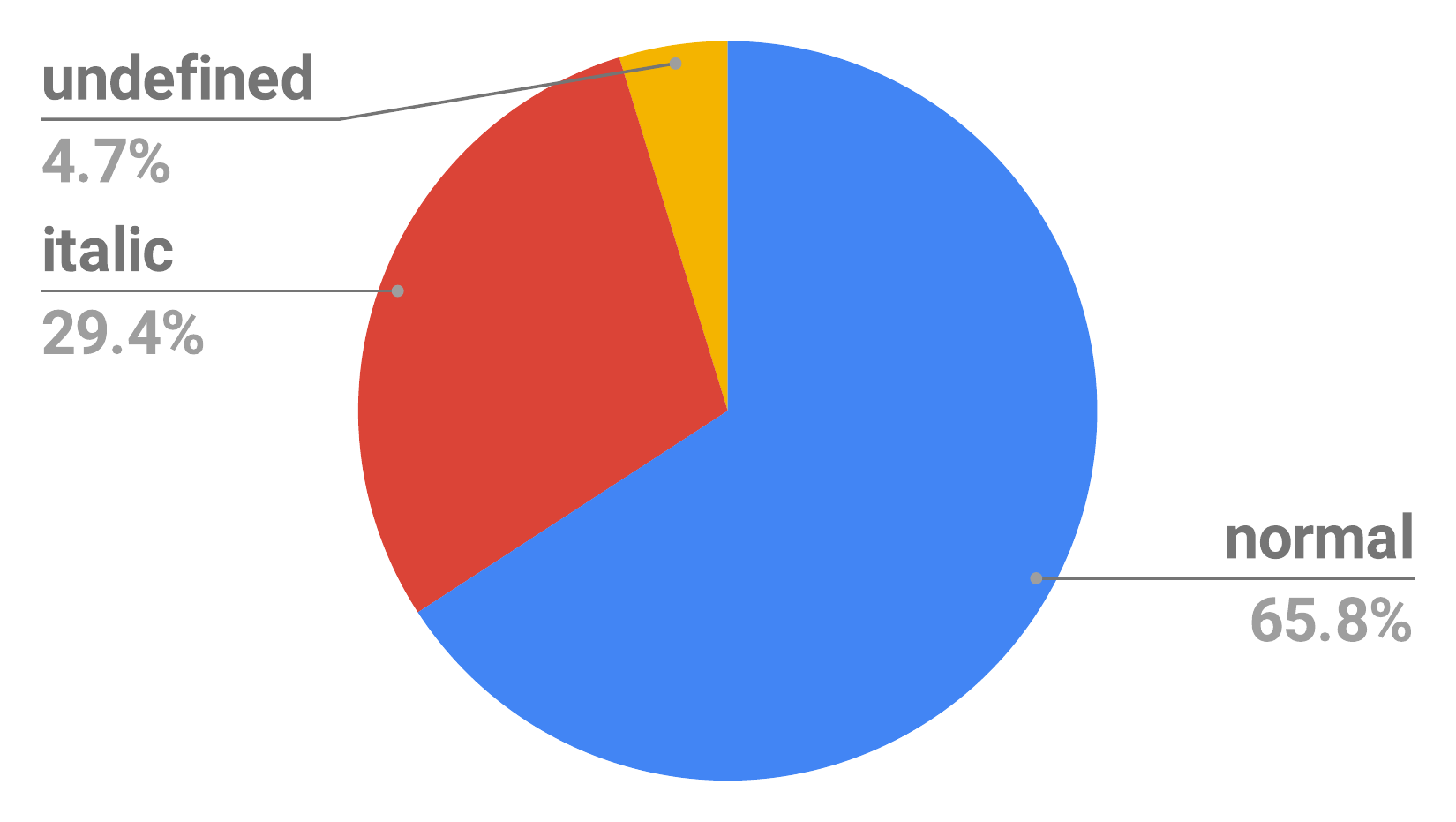}
    \includegraphics[width=0.25\linewidth]{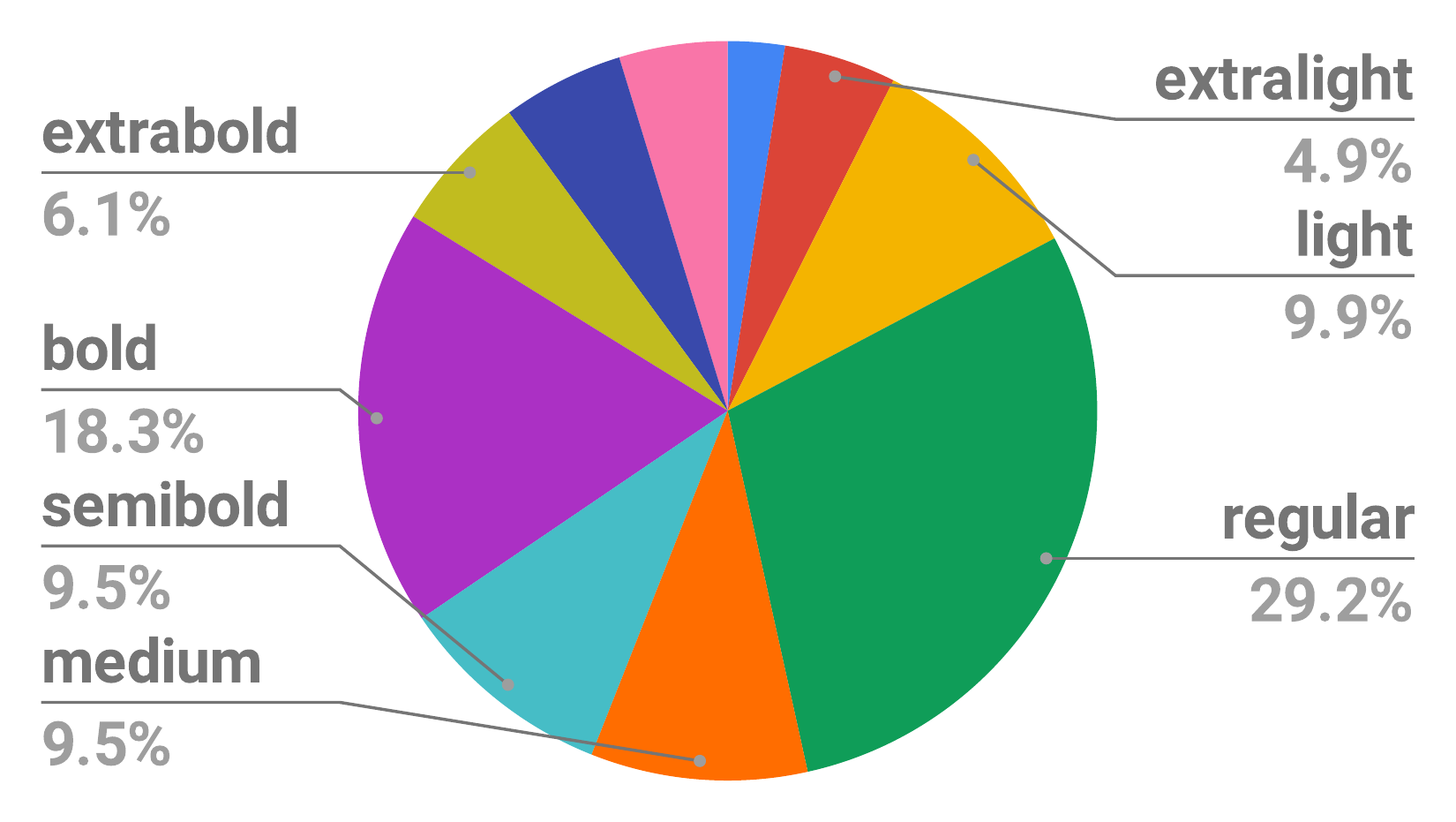}
\end{center}
\caption{Google Fonts dataset: the number of fonts per letter, distribution of font categories, styles, and different stroke widths.}
\label{data_distribution}
\end{figure}

After data preprocess, the dataset has an imbalance distribution as shown in Fig.~\ref{data_distribution}. For some characters (``x'', ``y'', ``z'', etc.) there are not many glyphs due to the failure in keypoint detection during template mapping. Besides, since glyphs of sans serif, normal, and regular stroke width dominate the dataset, font completion performance of these styles are slightly better than the others.

\section{Additional Results}

\begin{figure*}[!h]
\begin{subfigure}[b]{\textwidth}
    \includegraphics[width=\textwidth]{figures/completion_2/gt/ofl_abeezee_ABeeZee-Italic.png}
    \includegraphics[width=\textwidth]{figures/completion_2/img2seq2img/B_ofl_abeezee_ABeeZee-Italic_all.png}
    \includegraphics[width=\textwidth]{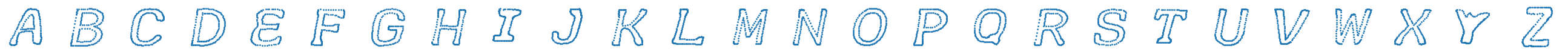}\vspace{-0.2em}
    \caption{ABeeZee-Italic}
\end{subfigure}
\begin{subfigure}[b]{\textwidth}
    \vspace{0.3em}\includegraphics[width=\textwidth]{figures/completion_2/gt/ofl_capriola_Capriola-Regular.png}
    \includegraphics[width=\textwidth]{figures/completion_2/img2seq2img/B_ofl_capriola_Capriola-Regular_all.png}
    \includegraphics[width=\textwidth]{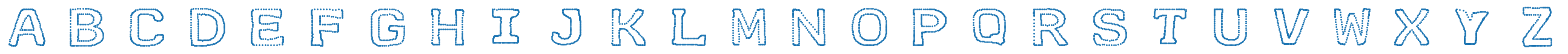}\vspace{-0.2em}
    \caption{Capriola-Regular}
\end{subfigure}
\begin{subfigure}[b]{\textwidth}
    \vspace{0.3em}\includegraphics[width=\textwidth]{figures/completion_2/gt/ofl_sansita_Sansita-Italic.png}
    \includegraphics[width=\textwidth]{figures/completion_2/img2seq2img/B_ofl_sansita_Sansita-Italic_all.png}
    \includegraphics[width=\textwidth]{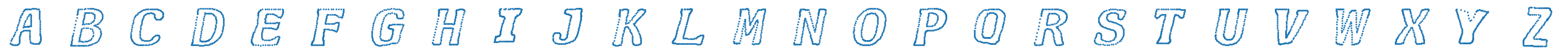}\vspace{-0.2em}
    \caption{Sansita-Italic}
\end{subfigure}
\begin{subfigure}[b]{\textwidth}
    \vspace{0.3em}\includegraphics[width=\textwidth]{figures/completion_2/gt/ofl_eczar_Eczar-Regular.png}
    \includegraphics[width=\textwidth]{figures/completion_2/img2seq2img/B_ofl_eczar_Eczar-Regular_all.png}
    \includegraphics[width=\textwidth]{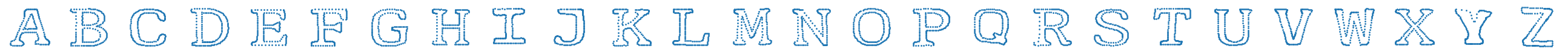}\vspace{-0.2em}
    \caption{Eczar-Regular}
\end{subfigure}
\begin{subfigure}[b]{\textwidth}
    \vspace{0.3em}\includegraphics[width=\textwidth]{figures/completion_2/gt/ofl_capriola_Capriola-Regular.png}
    \includegraphics[width=\textwidth]{figures/completion_2/img2seq2img/B_ofl_capriola_Capriola-Regular_all.png}
    \includegraphics[width=\textwidth]{figures/completion_2/img2seq/B_ofl_capriola_Capriola-Regular_seq.png}\vspace{-0.2em}
    \caption{Capriola-Regular}
\end{subfigure}
\begin{subfigure}[b]{\textwidth}
    \vspace{0.3em}\includegraphics[width=\textwidth]{figures/completion_2/gt/ofl_gudea_Gudea-Italic.png}
    \includegraphics[width=\textwidth]{figures/completion_2/img2seq2img/B_ofl_gudea_Gudea-Italic_all.png}
    \includegraphics[width=\textwidth]{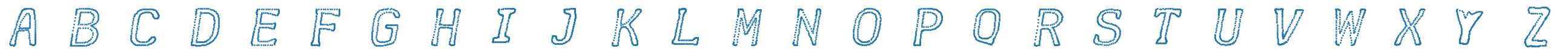}
    \caption{Gudea-Italic}
\end{subfigure}
\caption{Font completion visualization. Three rows in each style panel: ground truth glyphs (top), results of our image-to-graph-to-image approach (middle), and intermediate graph visualization of our image-to-graph-to-image approach (bottom). Red boxes indicate input glyphs. Our method completes glyphs with clear appearance and consistent style with the target glyphs.
}
\label{fig:img2seq2img_compeletion}
\end{figure*}

\begin{figure*}[!h]
    \includegraphics[width=\textwidth]{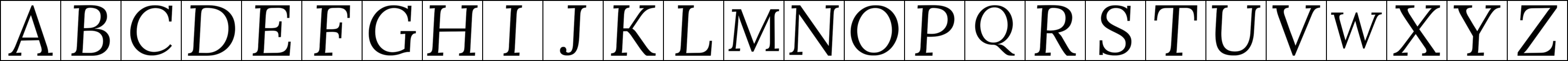}
    \includegraphics[width=\textwidth]{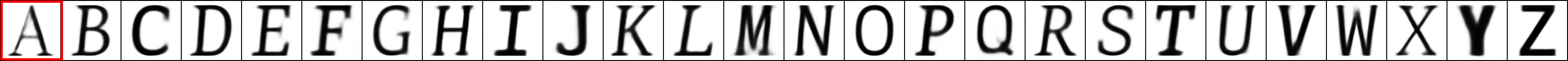}
    \includegraphics[width=\textwidth]{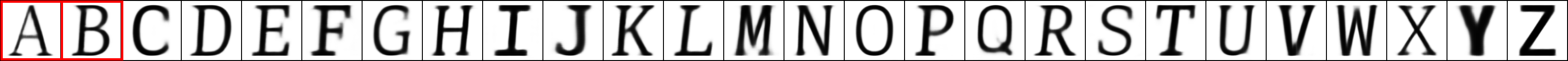}
    \includegraphics[width=\textwidth]{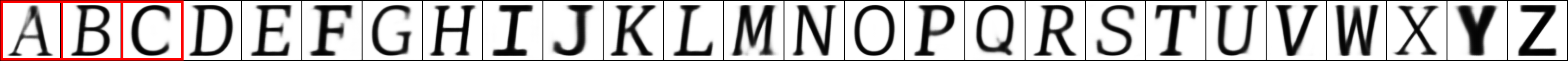}
    \includegraphics[width=\textwidth]{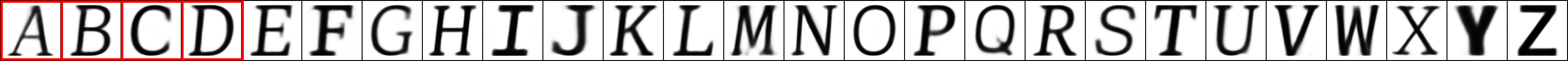}
    \includegraphics[width=\textwidth]{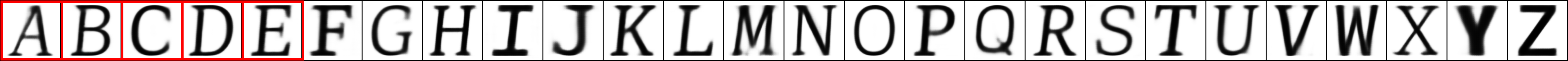}
    \includegraphics[width=\textwidth]{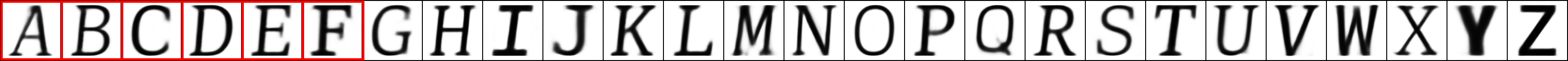}
    \includegraphics[width=\textwidth]{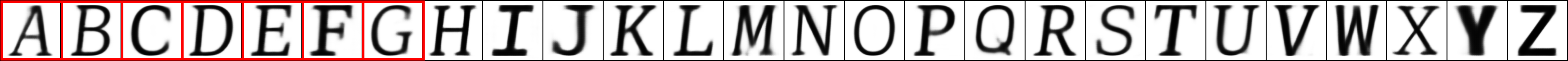}
    \includegraphics[width=\textwidth]{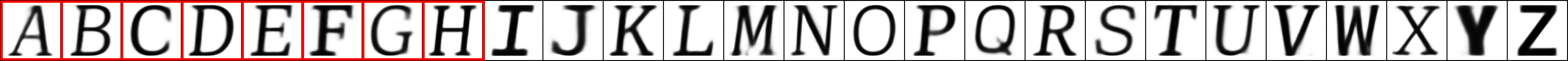}
    \includegraphics[width=\textwidth]{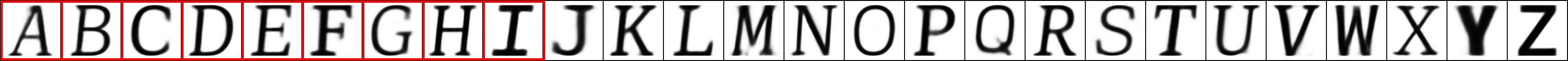}
    \includegraphics[width=\textwidth]{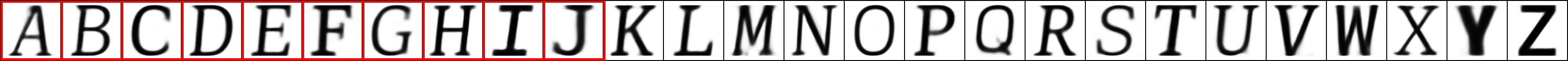}
    \includegraphics[width=\textwidth]{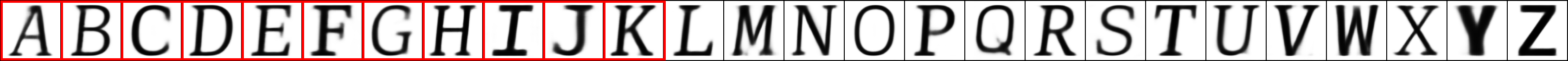}
    \includegraphics[width=\textwidth]{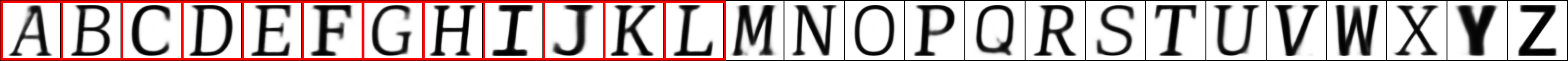}
    \includegraphics[width=\textwidth]{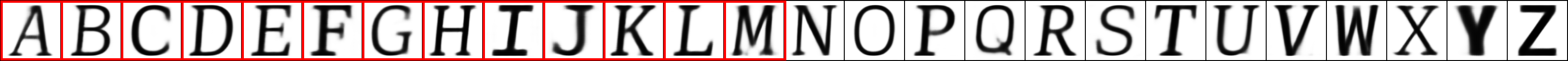}
    \includegraphics[width=\textwidth]{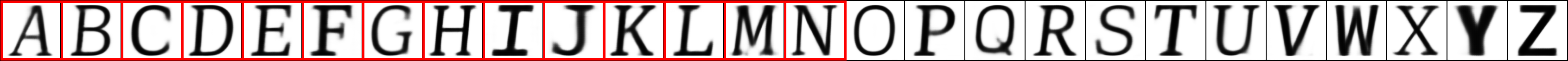}
    \includegraphics[width=\textwidth]{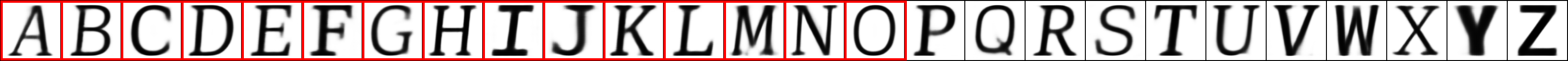}
    \includegraphics[width=\textwidth]{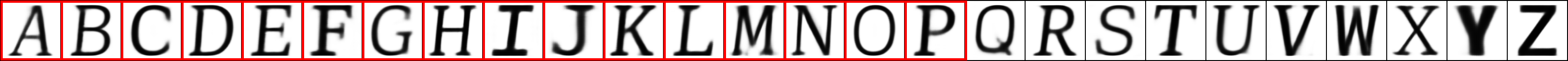}
    \includegraphics[width=\textwidth]{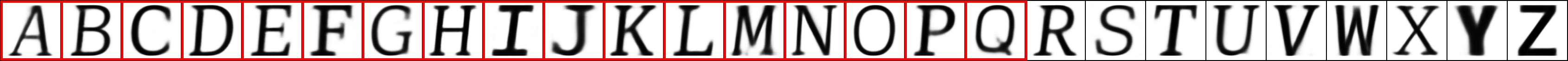}
    \includegraphics[width=\textwidth]{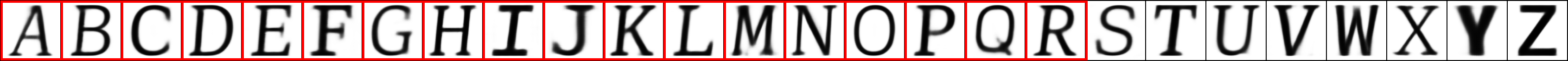}
    \includegraphics[width=\textwidth]{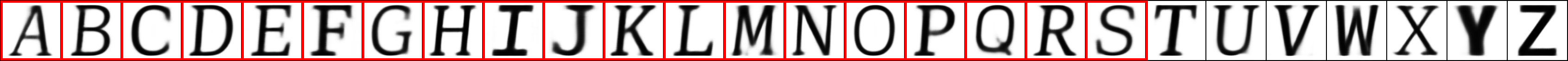}
    \includegraphics[width=\textwidth]{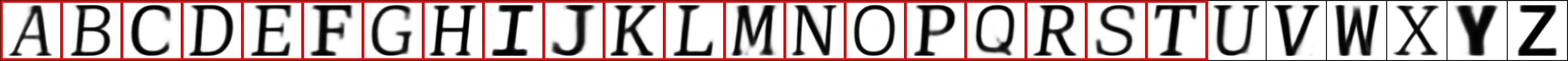}
    \includegraphics[width=\textwidth]{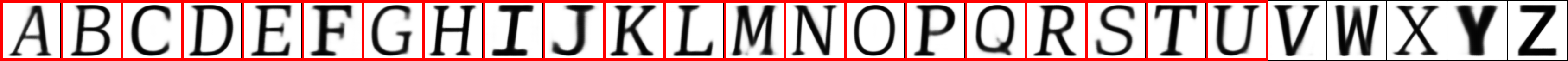}
    \includegraphics[width=\textwidth]{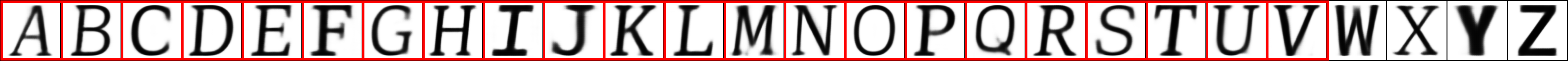}
    \includegraphics[width=\textwidth]{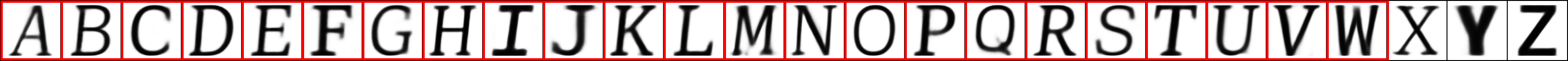}
    \includegraphics[width=\textwidth]{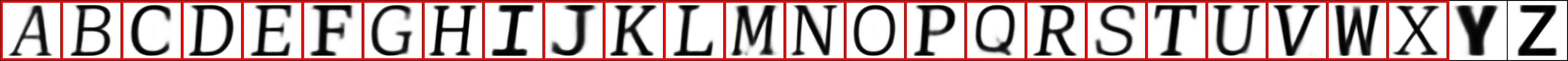}
    \includegraphics[width=\textwidth]{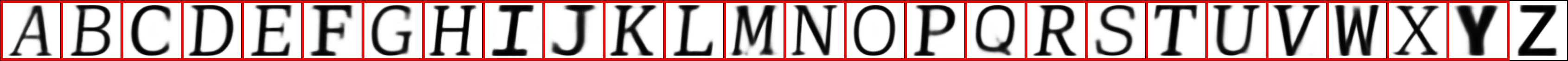}
    \includegraphics[width=\textwidth]{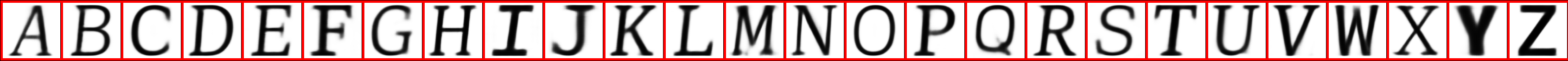}
\caption{Multi-input inference of our image-to-graph-to-image method, where we use averaged style feature of all input to complete the remaining glyphs.}
\label{multi_infer}
\end{figure*}

\begin{figure*}[!h]
    \includegraphics[width=\textwidth]{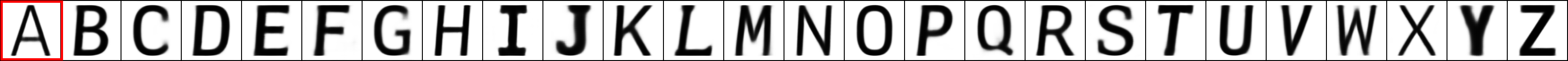}
	\includegraphics[width=\textwidth]{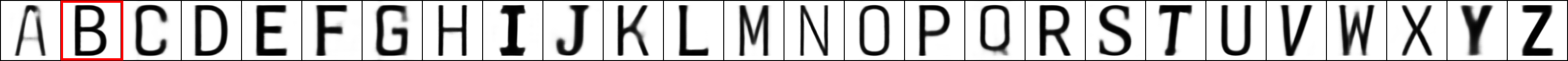}
	\includegraphics[width=\textwidth]{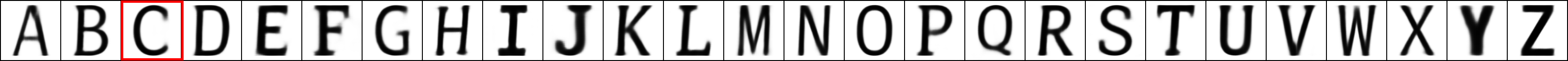}
	\includegraphics[width=\textwidth]{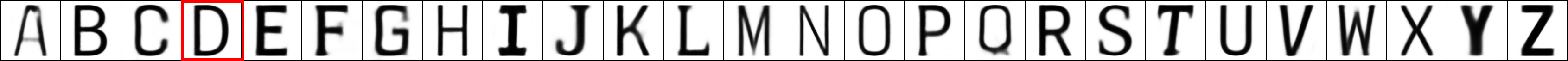}
	\includegraphics[width=\textwidth]{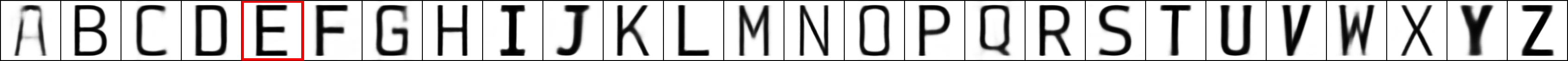}
	\includegraphics[width=\textwidth]{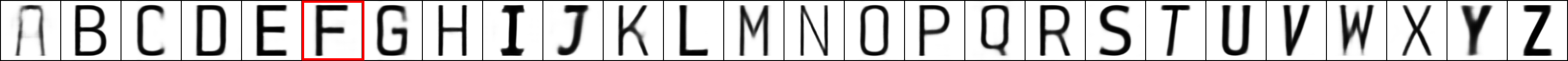}
	\includegraphics[width=\textwidth]{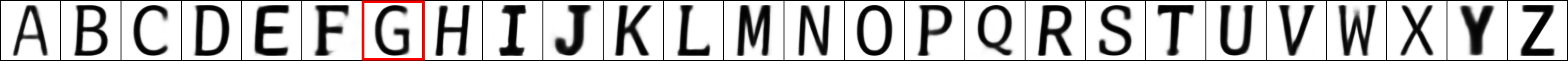}
	\includegraphics[width=\textwidth]{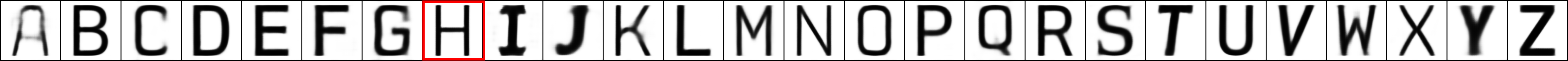}
	\includegraphics[width=\textwidth]{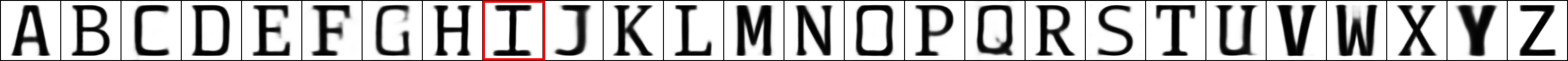}
	\includegraphics[width=\textwidth]{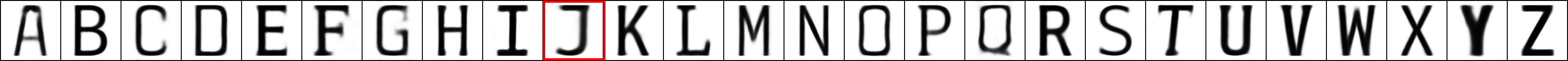}
	\includegraphics[width=\textwidth]{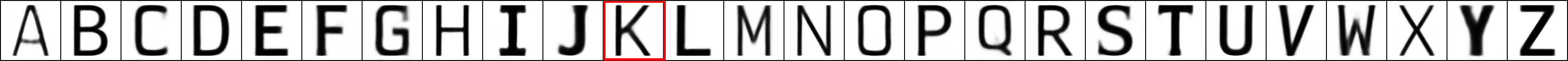}
	\includegraphics[width=\textwidth]{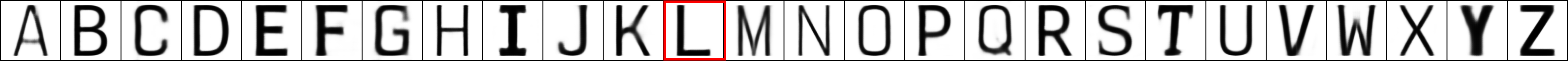}
	\includegraphics[width=\textwidth]{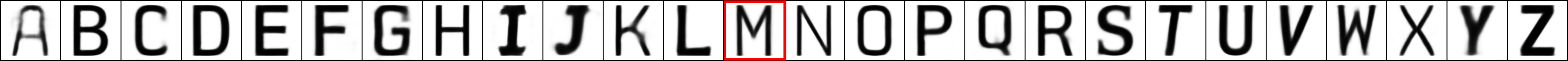}
	\includegraphics[width=\textwidth]{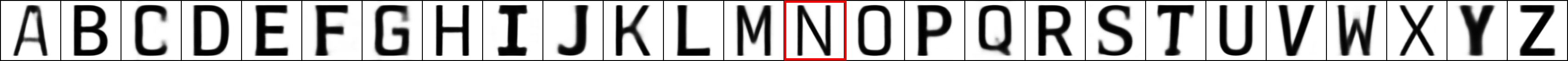}
	\includegraphics[width=\textwidth]{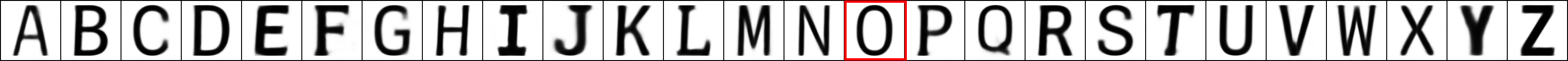}
	\includegraphics[width=\textwidth]{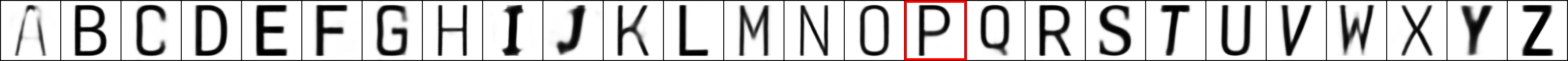}
	\includegraphics[width=\textwidth]{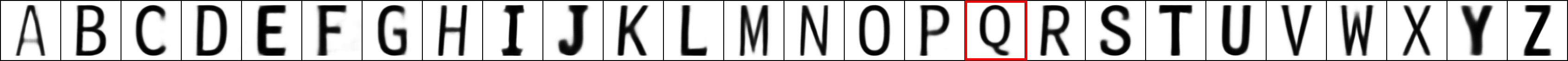}
	\includegraphics[width=\textwidth]{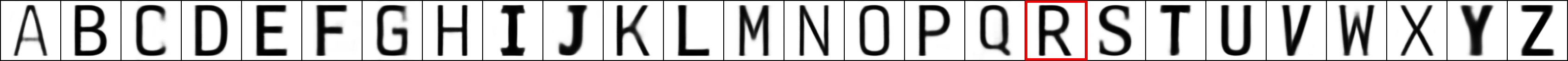}
	\includegraphics[width=\textwidth]{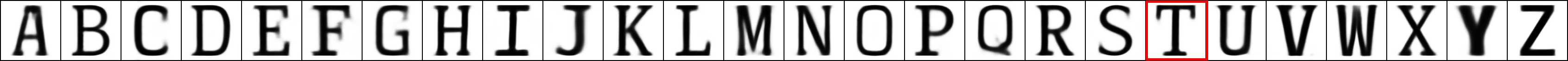}
	\includegraphics[width=\textwidth]{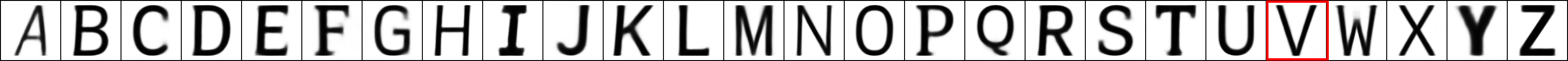}
	\includegraphics[width=\textwidth]{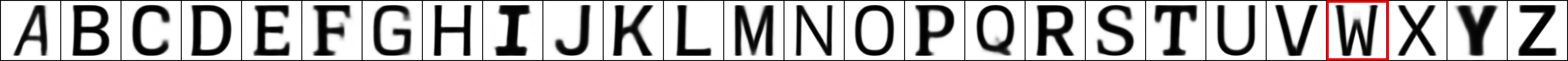}
	\includegraphics[width=\textwidth]{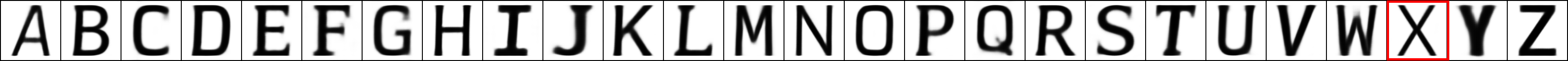}
\caption{Failure case of image-to-graph-to-image (font style: ptmono-PTM55FT).}
\label{failure_case}
\end{figure*}

The image completion and graph completion visualizations in Fig.\ref{fig:img2seq2img_compeletion} also indicate that part of the font completion error comes from the graph renderers, and the final completion performance can be improved by switching to a better graph renderer.
We also add the visualizations of the intermediate graph representation in font completion task Fig.~\ref{fig:img2seq2img_compeletion}. We also compare the single glyph font completion and  multi-glyph font completion performance. As we could see from Fig.~\ref{multi_infer}, the single glyph font completion has a comparable performance to multi-graph completion. However, for some font, i.e. ptmono-PTM55FT, single glyph could not represent the style, therefore font completion using different glyph as input results in inconsistent results (Fig.~\ref{failure_case}).

\begin{figure}[h!]
\centering
    \includegraphics[width=0.45\linewidth]{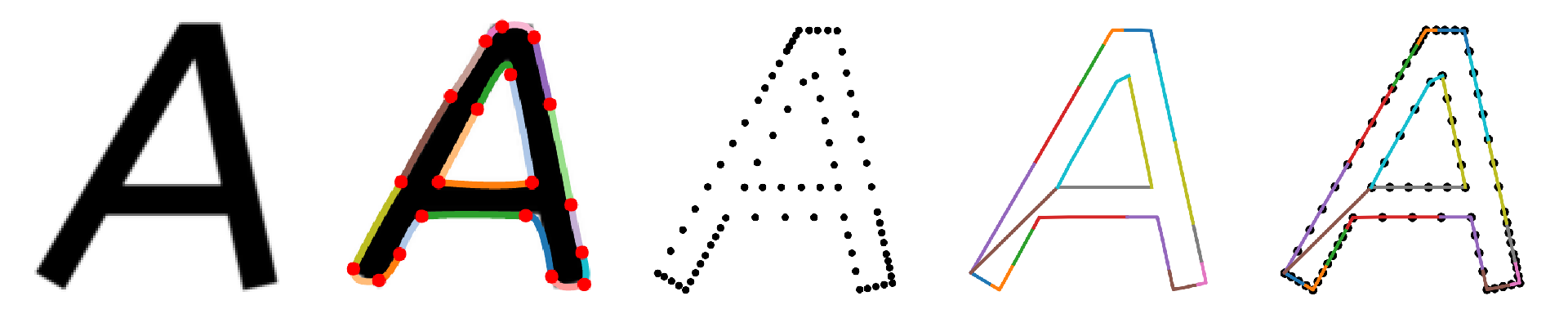} 
    \includegraphics[width=0.45\linewidth]{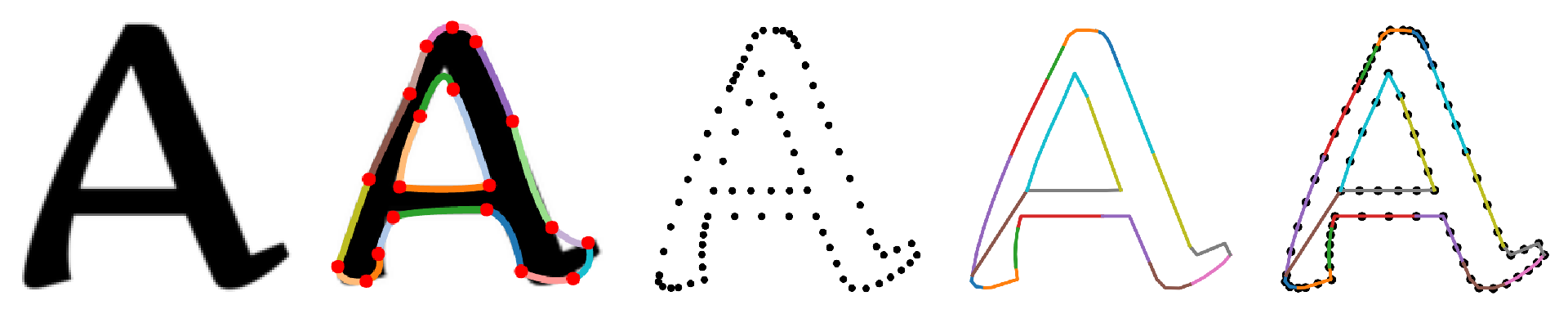} 
    \includegraphics[width=0.45\linewidth]{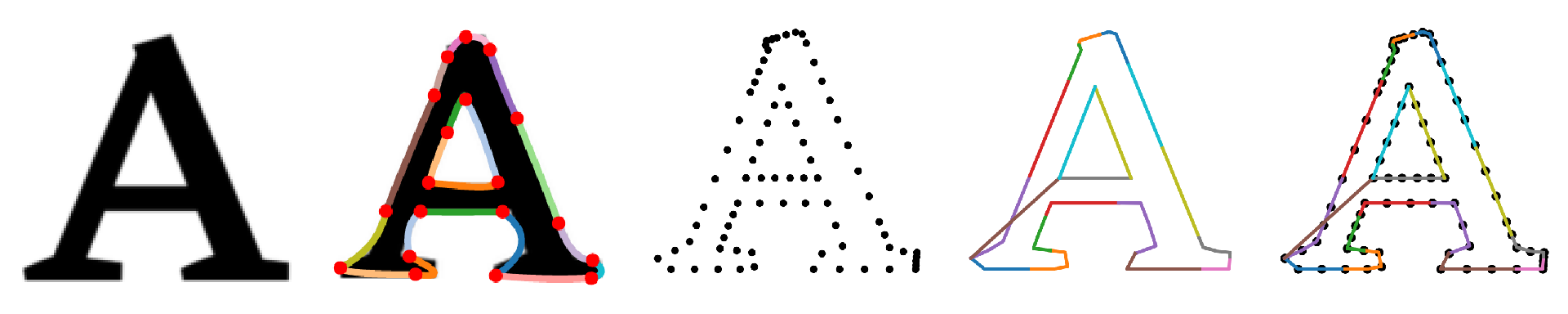} 
    \includegraphics[width=0.45\linewidth]{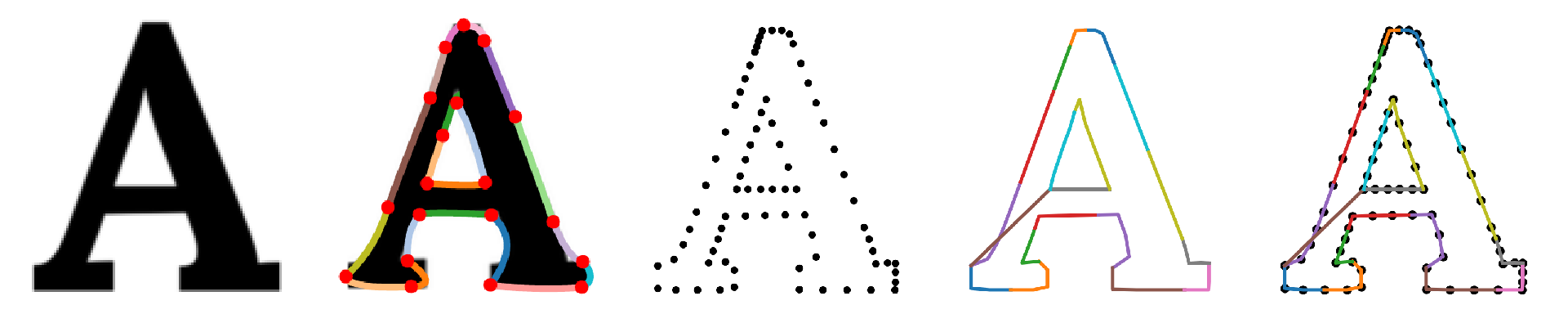} 
    \includegraphics[width=0.45\linewidth]{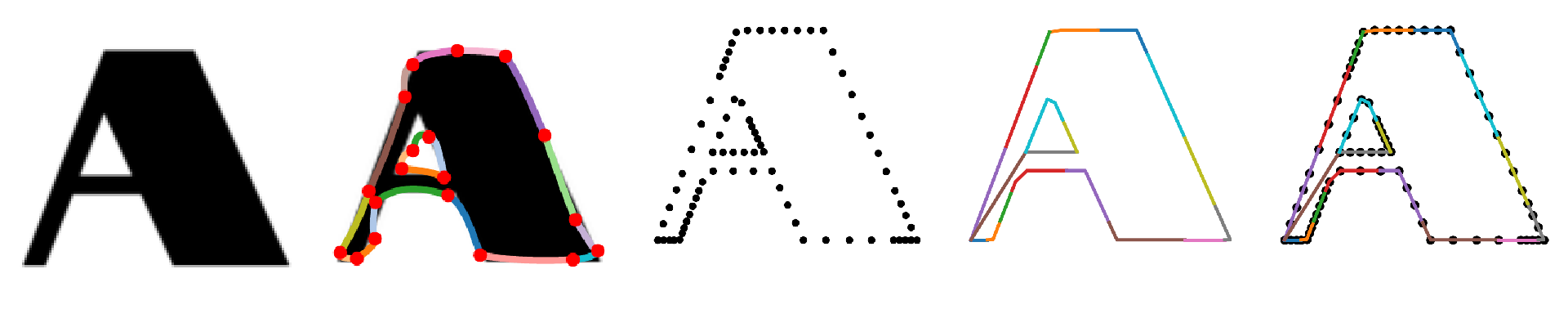} 
    \includegraphics[width=0.45\linewidth]{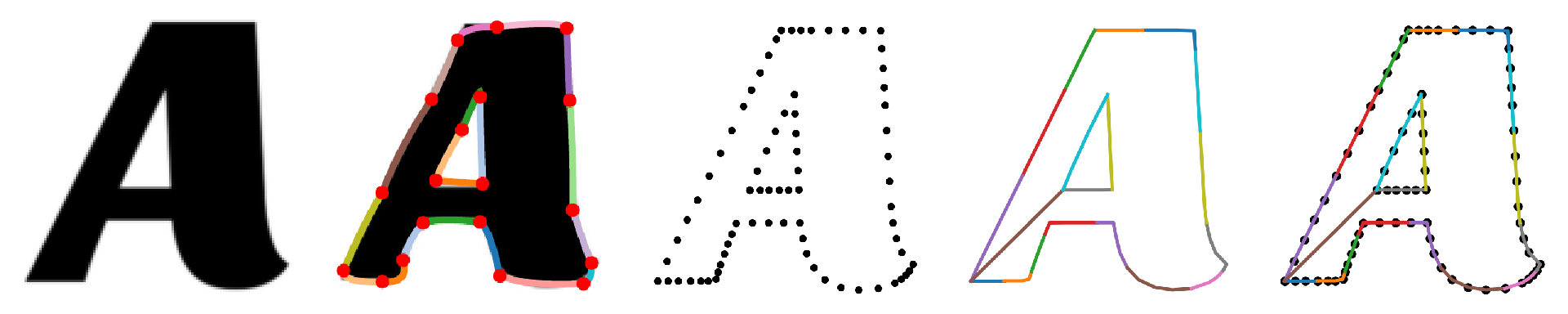}
    \includegraphics[width=0.45\linewidth]{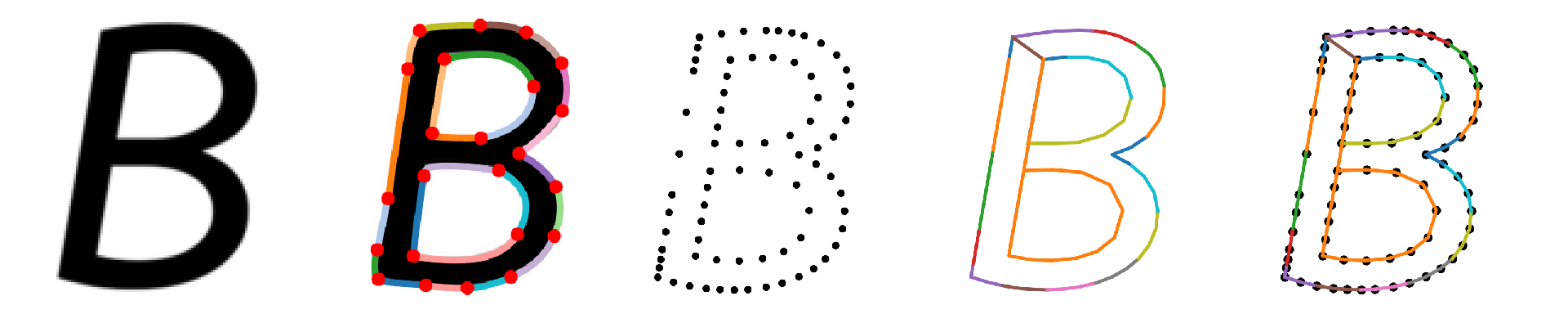} 
    \includegraphics[width=0.45\linewidth]{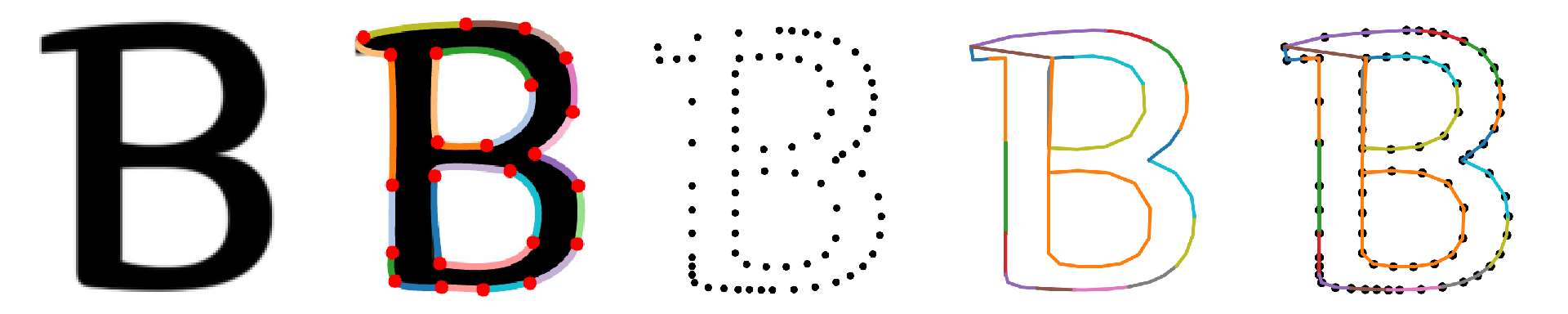} 
    \includegraphics[width=0.45\linewidth]{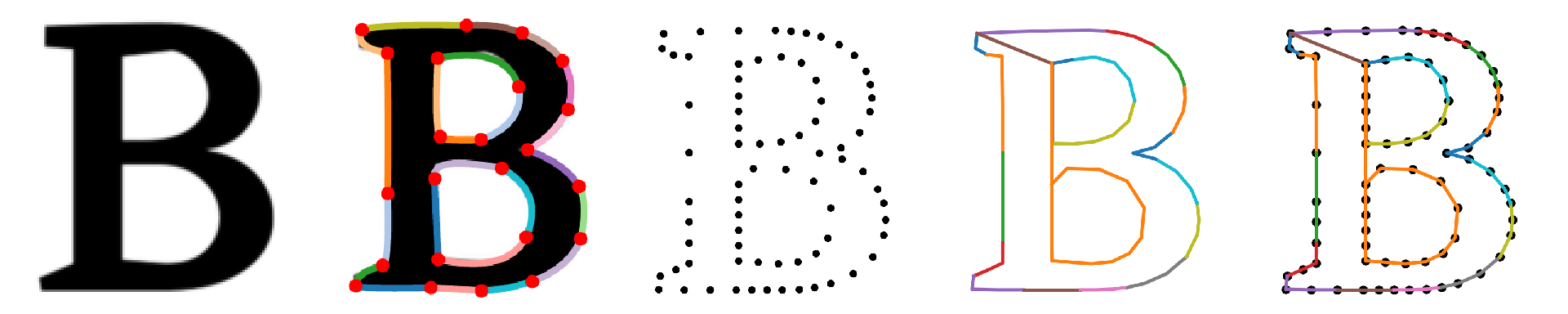} 
    \includegraphics[width=0.45\linewidth]{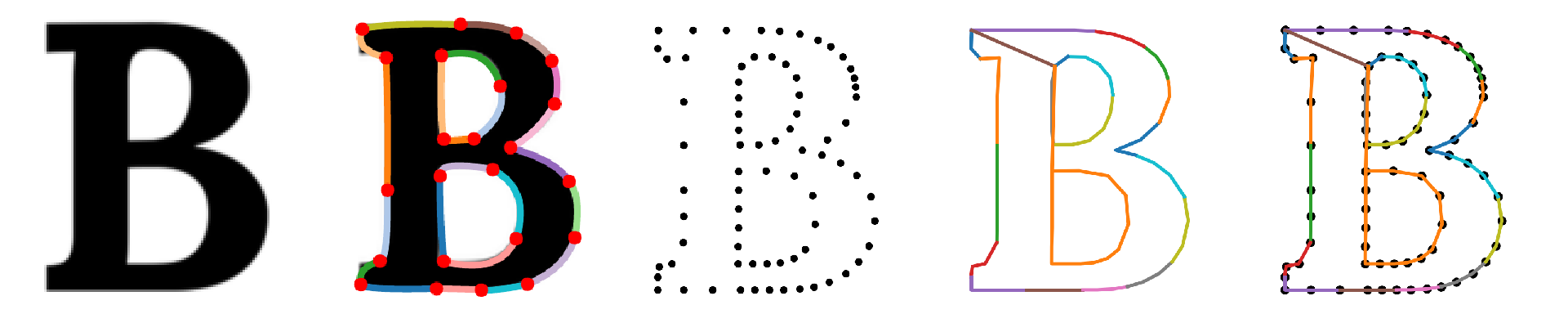} 
    \includegraphics[width=0.45\linewidth]{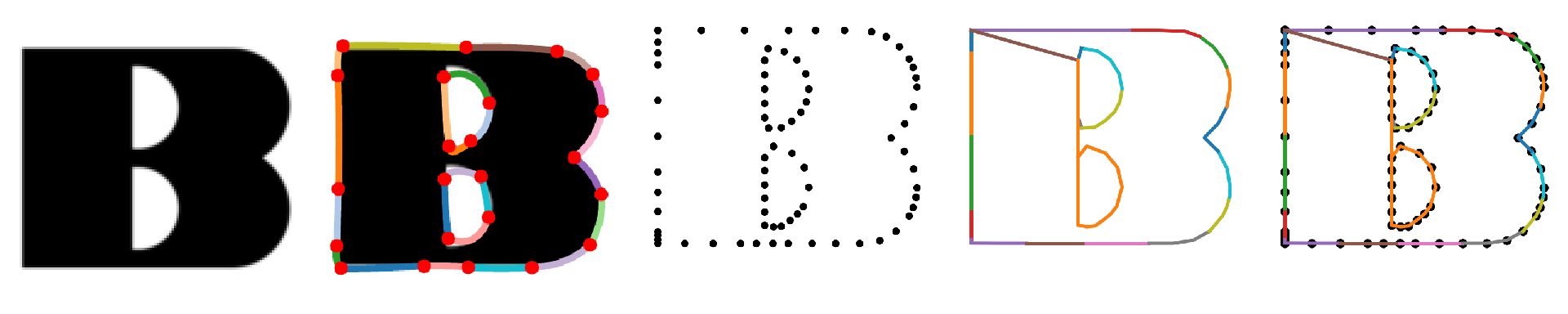} 
    \includegraphics[width=0.45\linewidth]{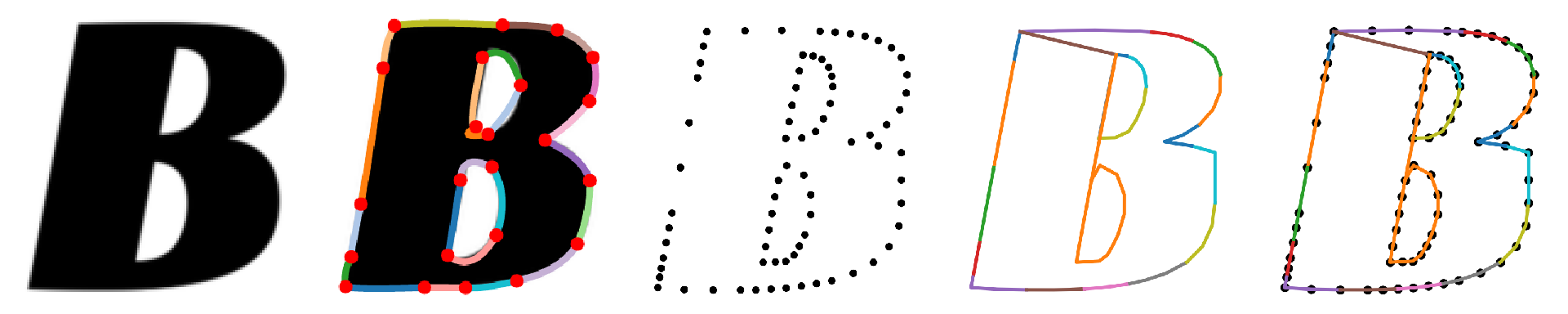} 
    \includegraphics[width=0.45\linewidth]{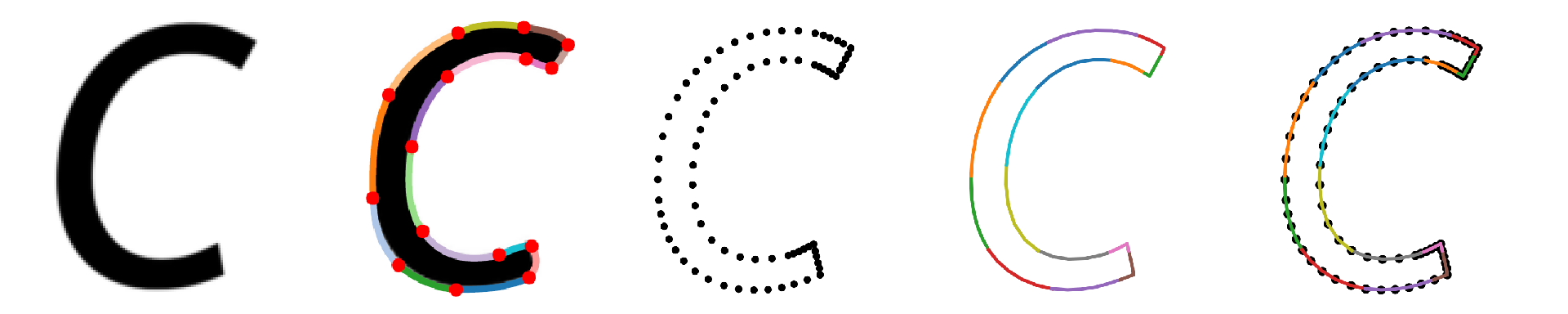} 
    \includegraphics[width=0.45\linewidth]{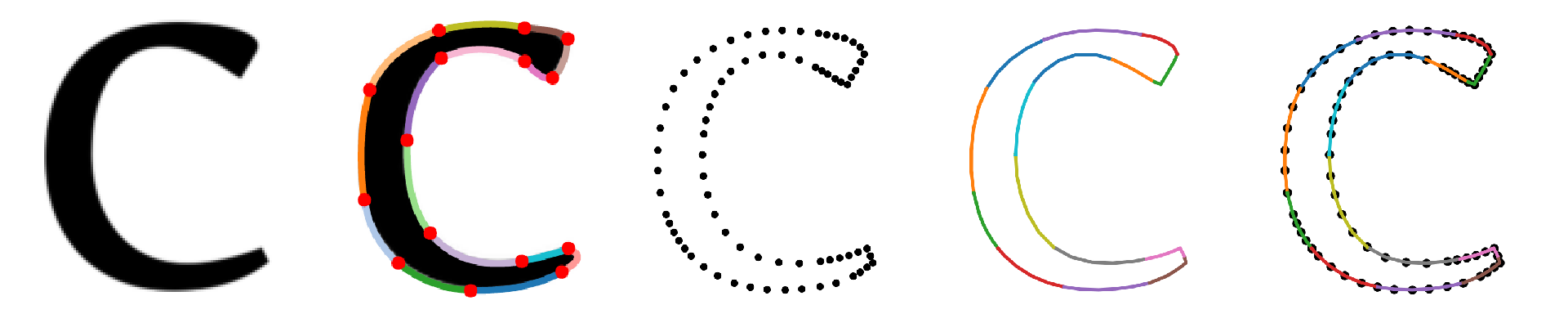} 
    \includegraphics[width=0.45\linewidth]{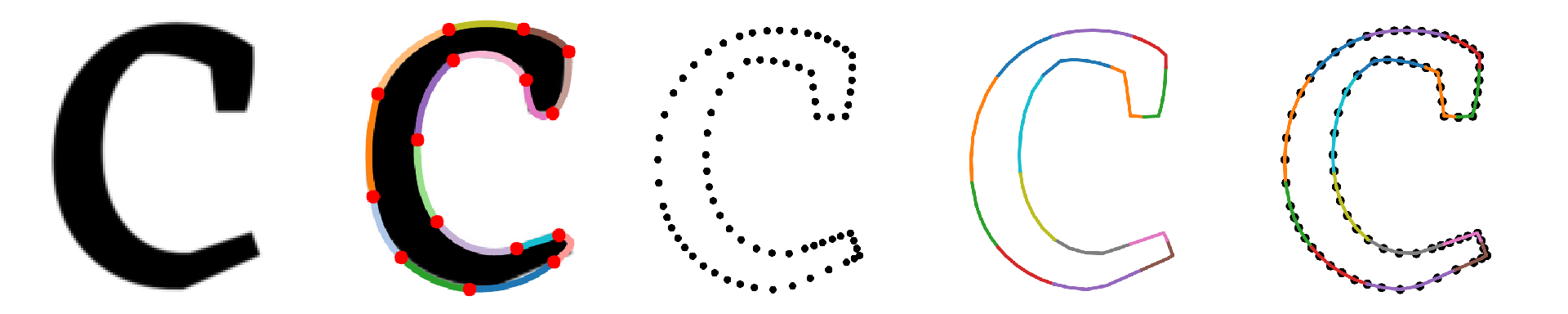} 
    \includegraphics[width=0.45\linewidth]{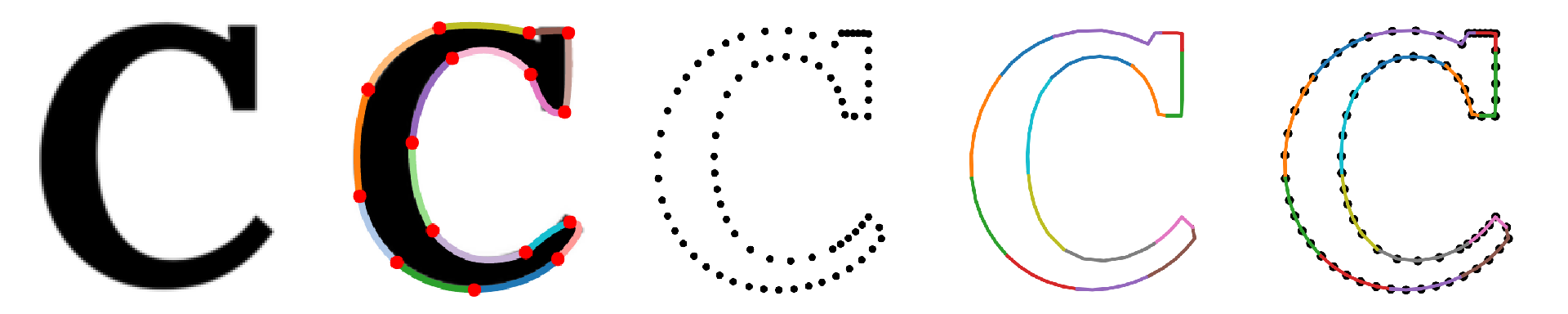} 
    \includegraphics[width=0.45\linewidth]{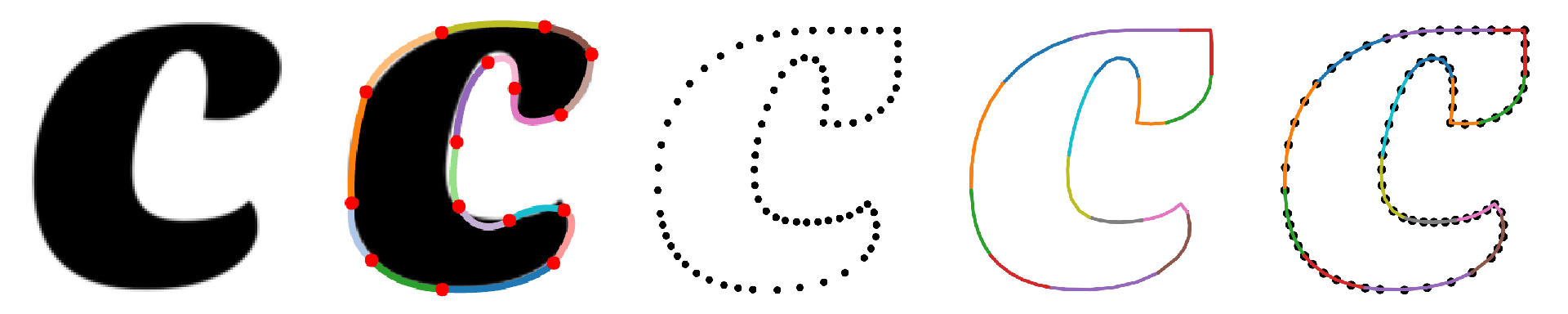} 
    \includegraphics[width=0.45\linewidth]{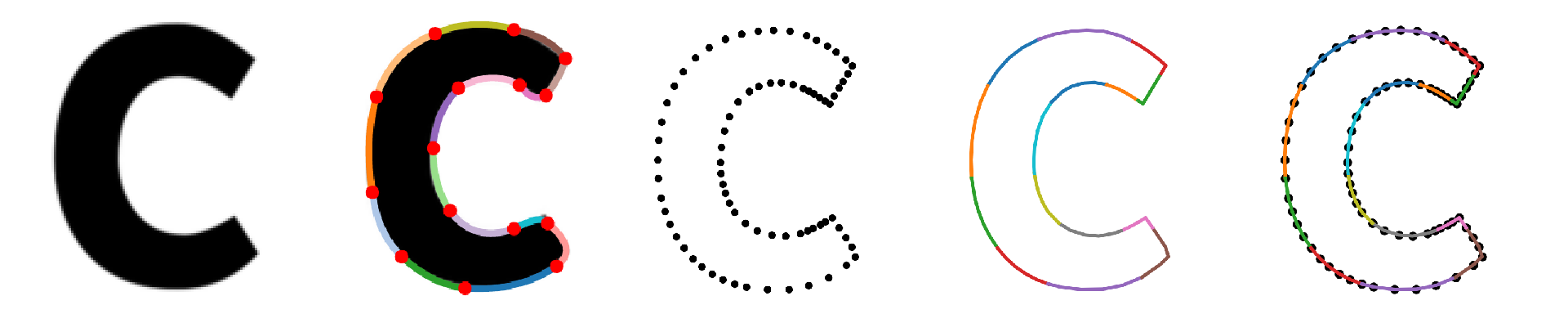} 
    \includegraphics[width=0.45\linewidth]{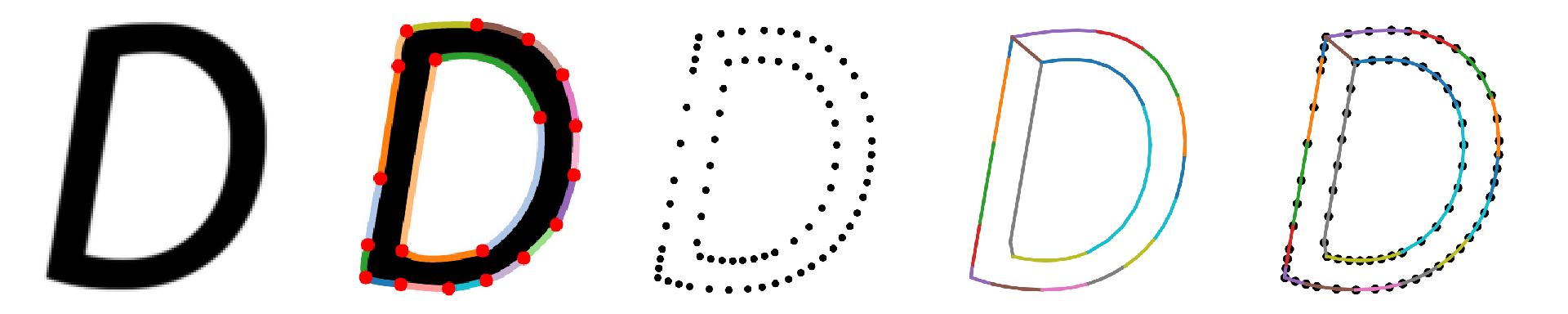} 
    \includegraphics[width=0.45\linewidth]{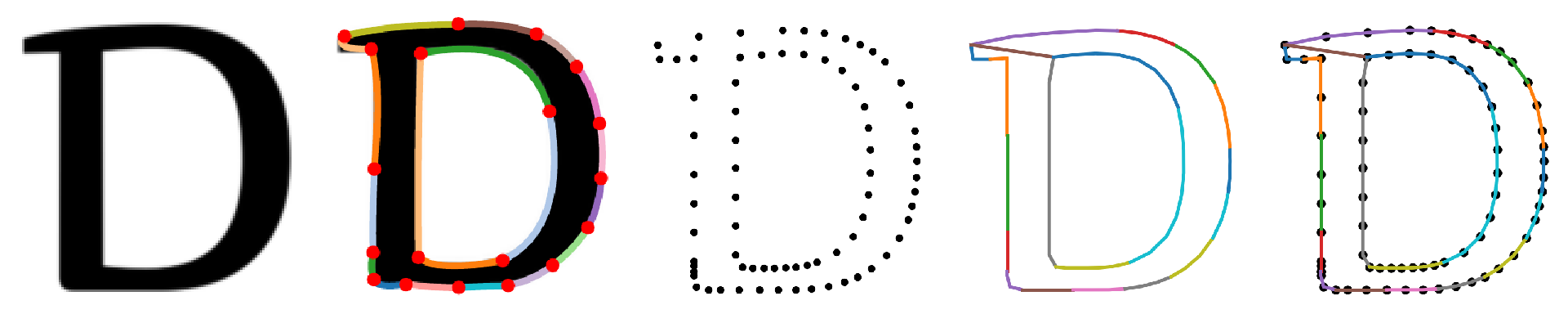} 
    \includegraphics[width=0.45\linewidth]{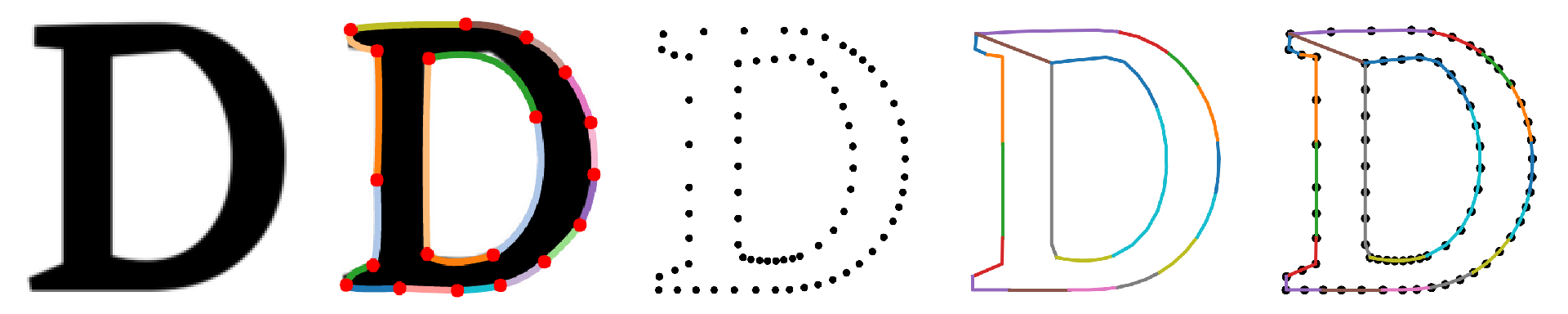} 
    \includegraphics[width=0.45\linewidth]{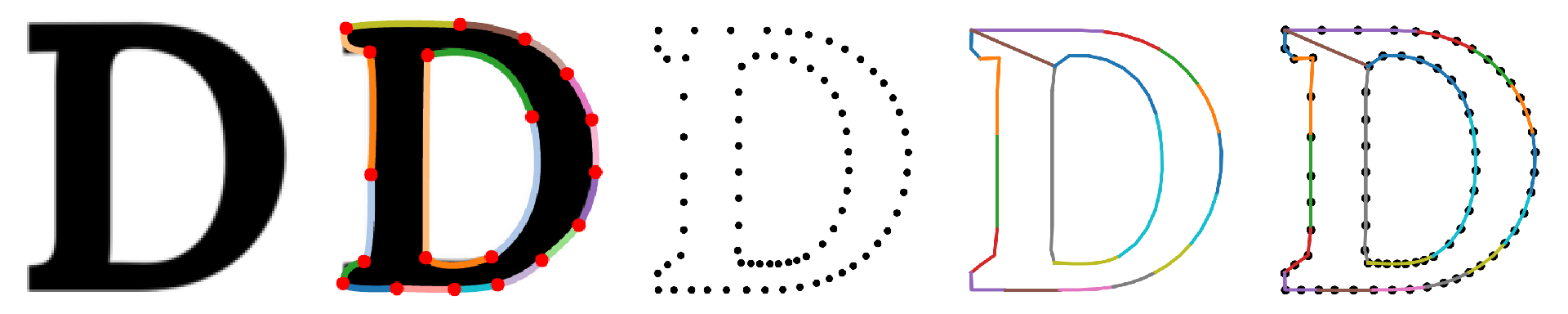} 
    \includegraphics[width=0.45\linewidth]{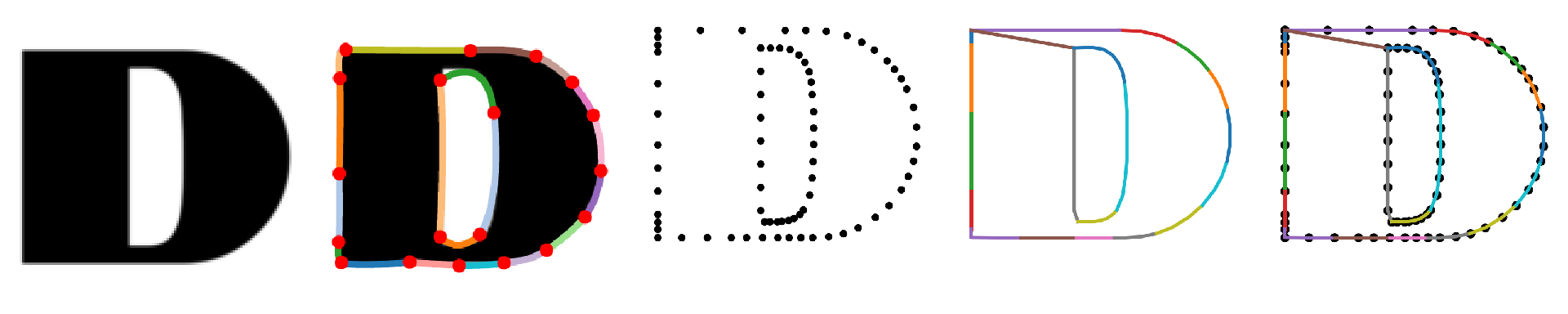} 
    \includegraphics[width=0.45\linewidth]{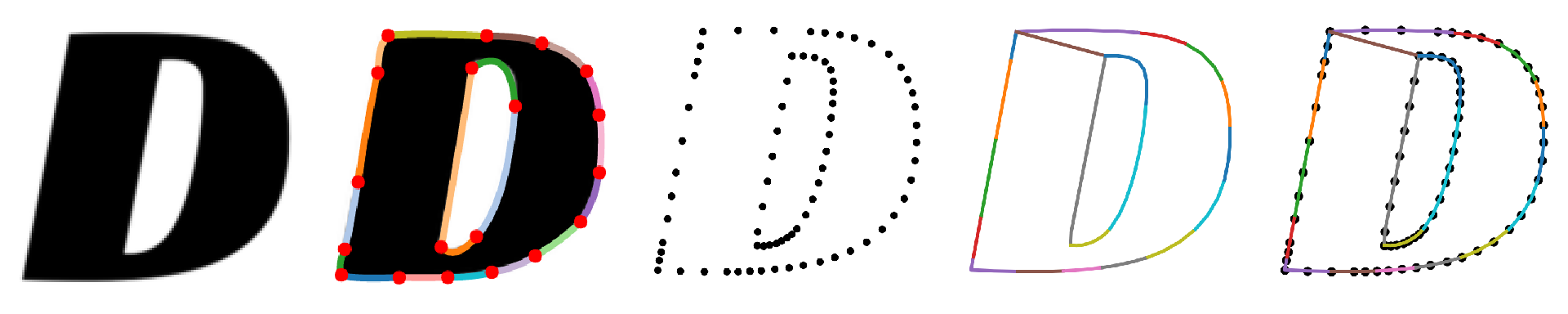}
    \includegraphics[width=0.45\linewidth]{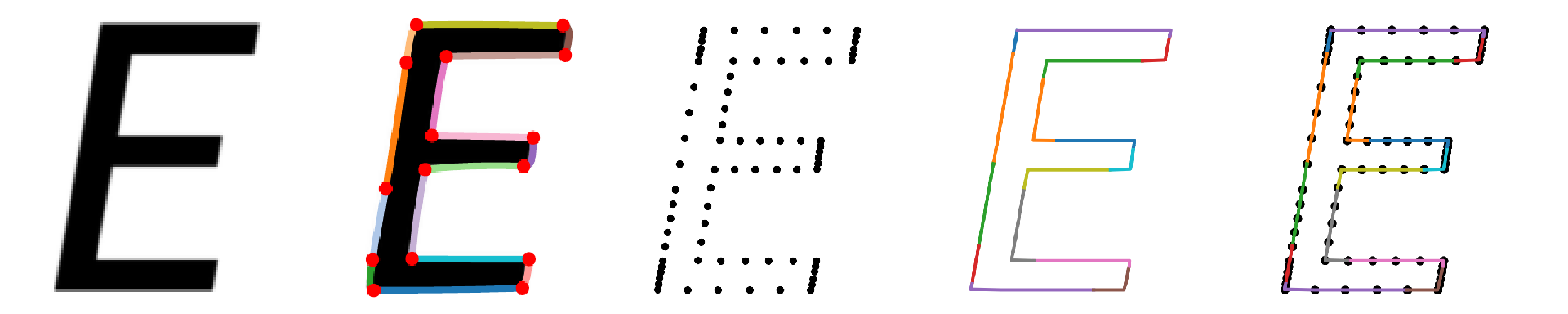} 
    \includegraphics[width=0.45\linewidth]{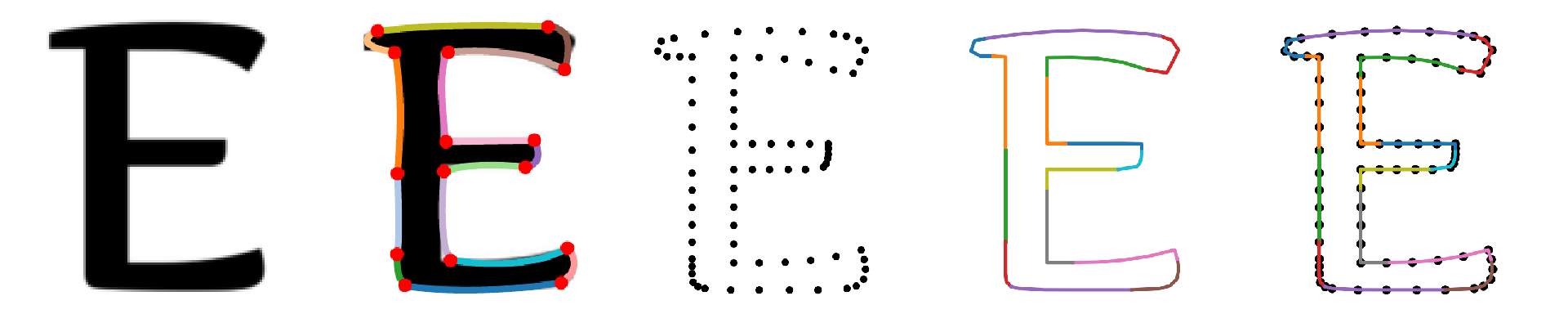} 
    \includegraphics[width=0.45\linewidth]{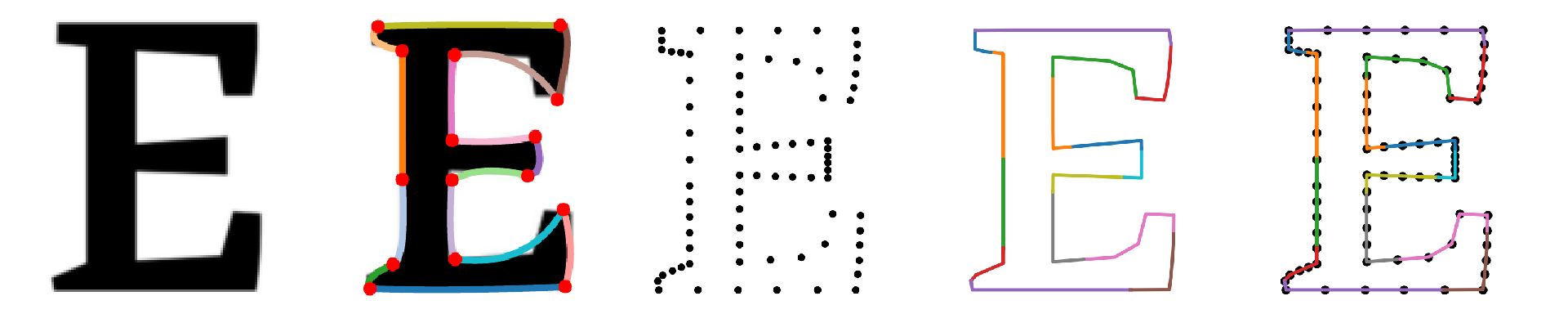} 
    \includegraphics[width=0.45\linewidth]{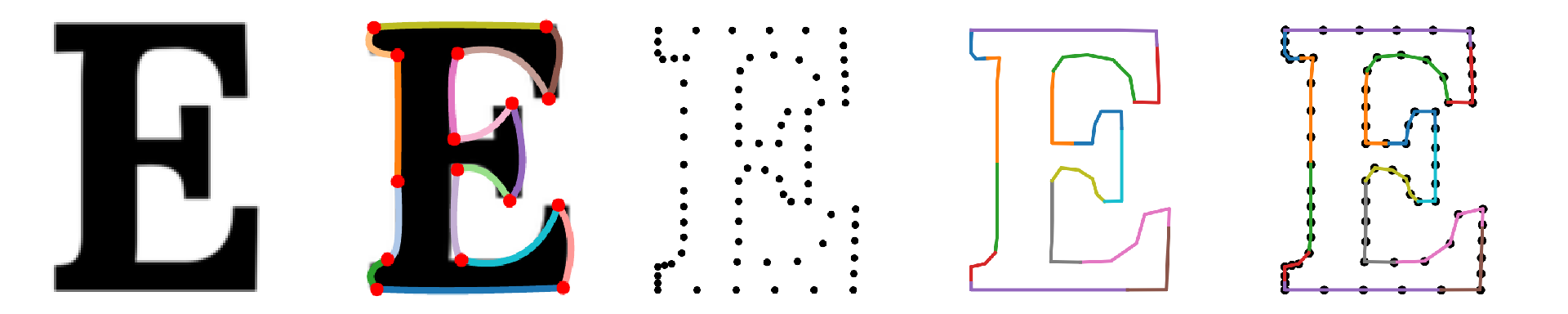} 
    \includegraphics[width=0.45\linewidth]{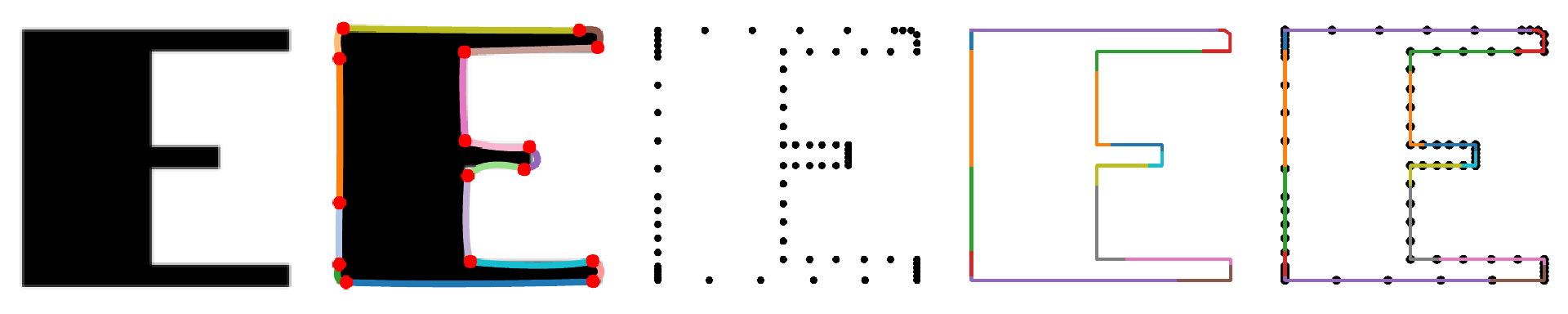} 
    \includegraphics[width=0.45\linewidth]{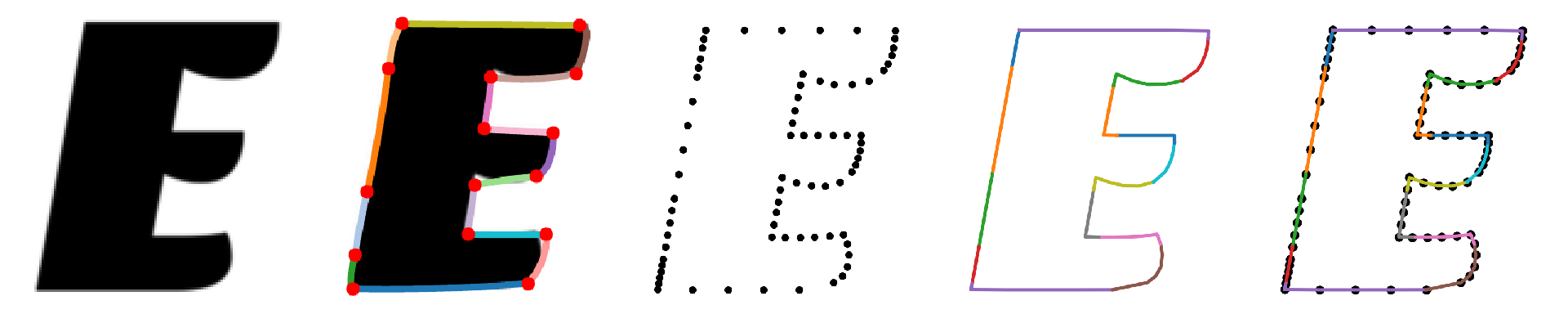} 
    \includegraphics[width=0.45\linewidth]{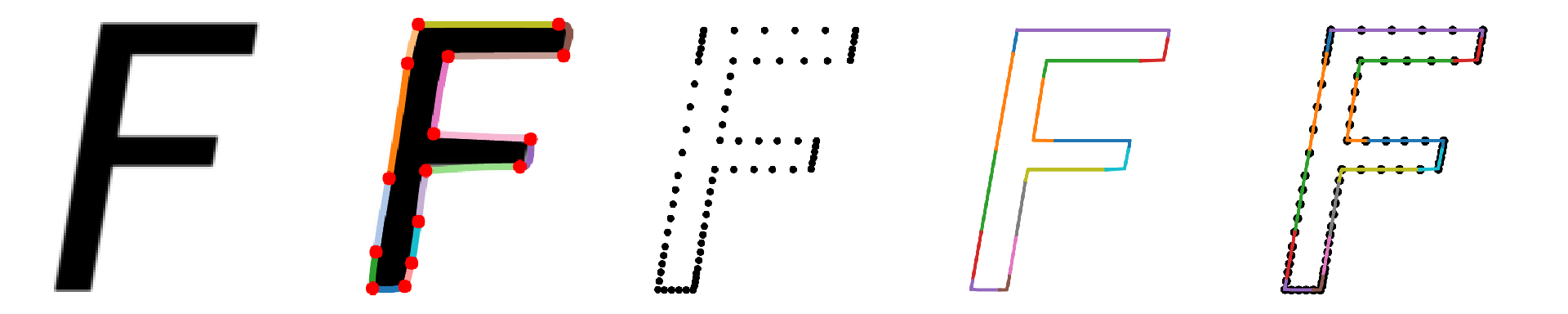} 
    \includegraphics[width=0.45\linewidth]{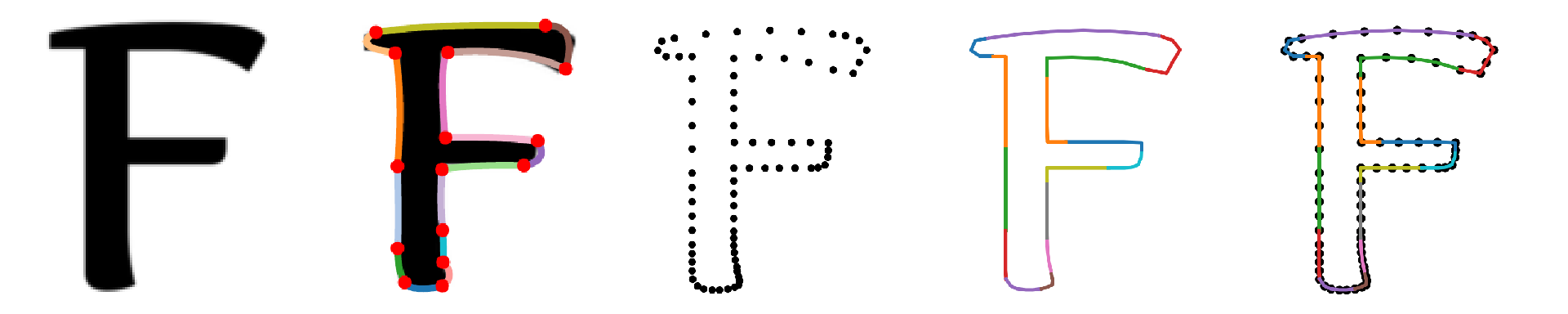} 
    \includegraphics[width=0.45\linewidth]{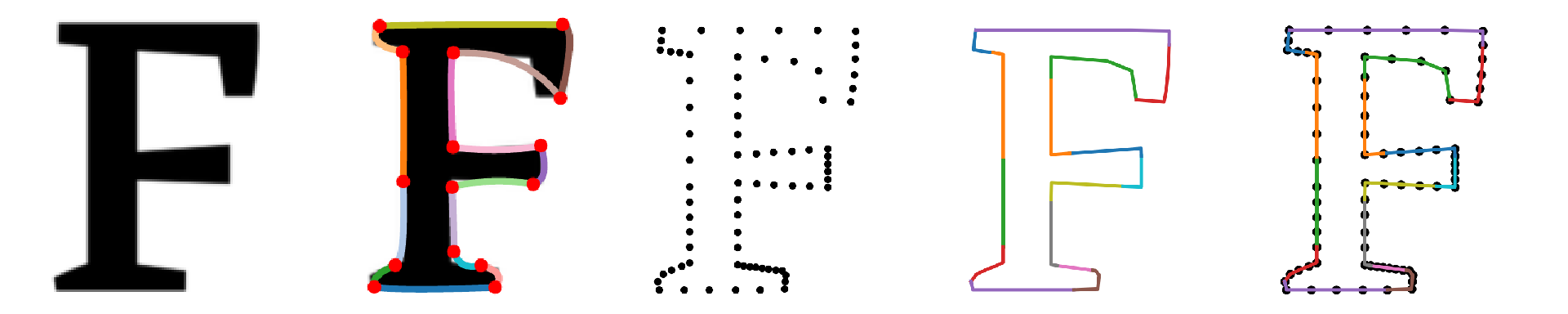} 
    \includegraphics[width=0.45\linewidth]{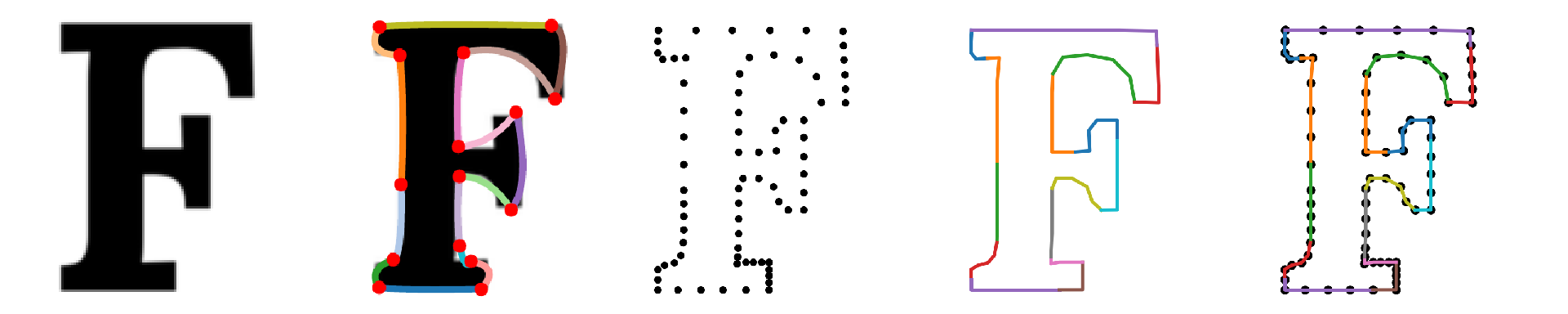} 
    \includegraphics[width=0.45\linewidth]{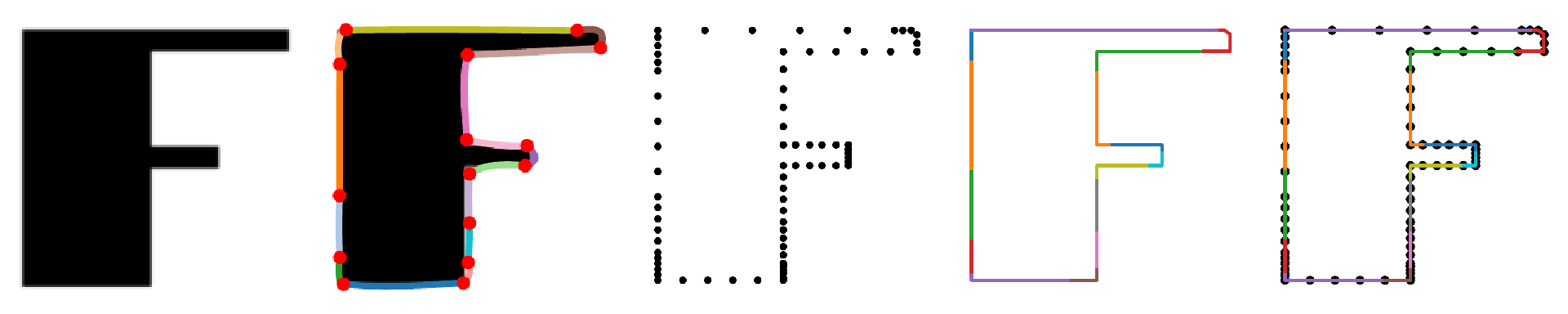} 
    \includegraphics[width=0.45\linewidth]{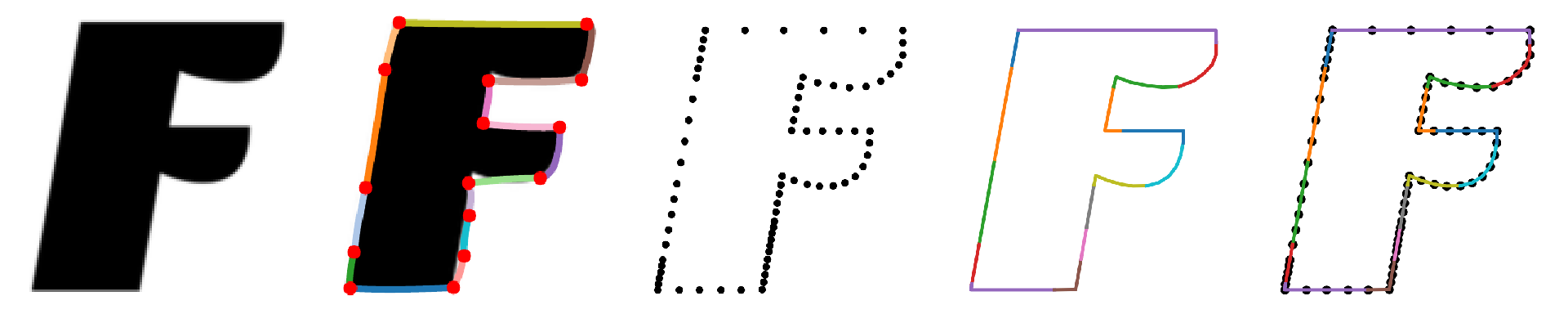} 
\caption{Different data modalities of font glyphs. For each example from left to right: image, keypoints, point set, graph template and directed graph representation. (To be continued.)}
\label{graph_construction}
\end{figure}

\begin{figure}[h!]\ContinuedFloat
\vspace{-0.5em}
\begin{center}
    \includegraphics[width=0.45\linewidth]{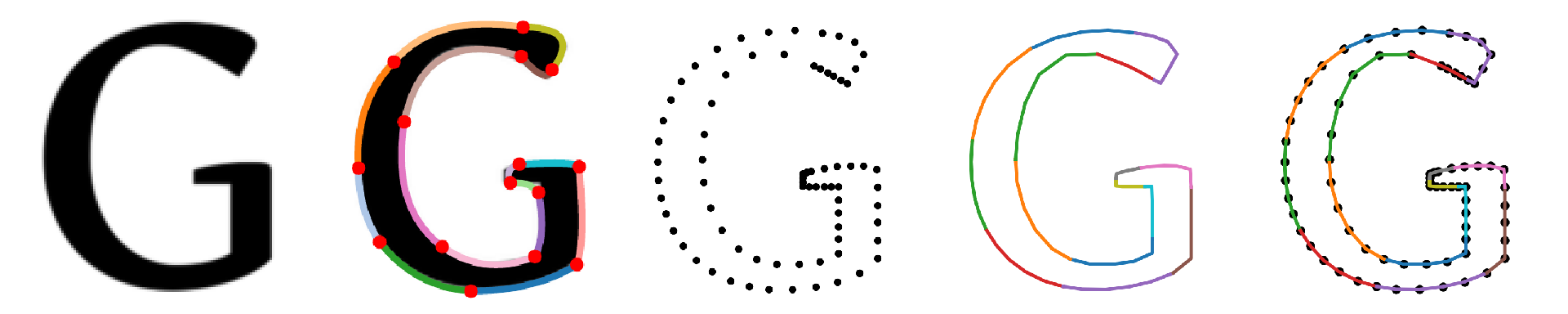} 
    \includegraphics[width=0.45\linewidth]{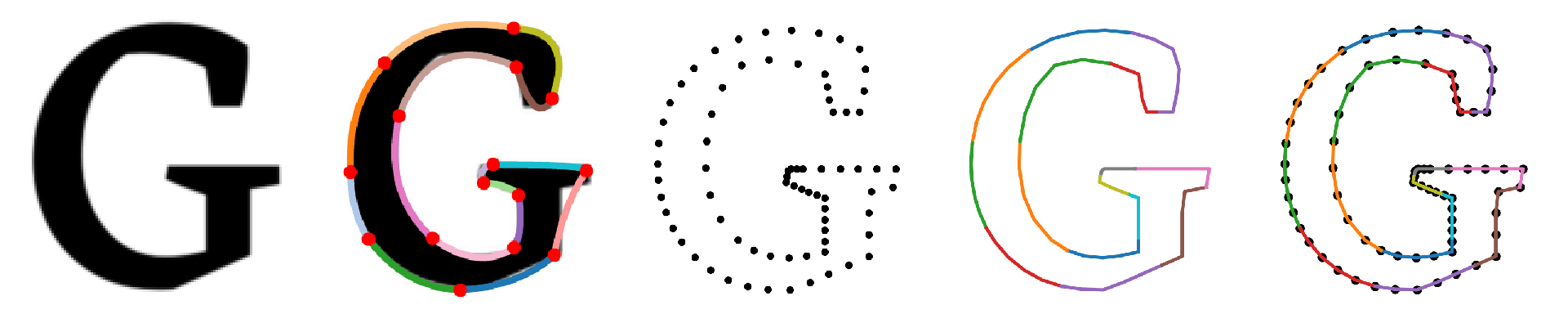} 
    \includegraphics[width=0.45\linewidth]{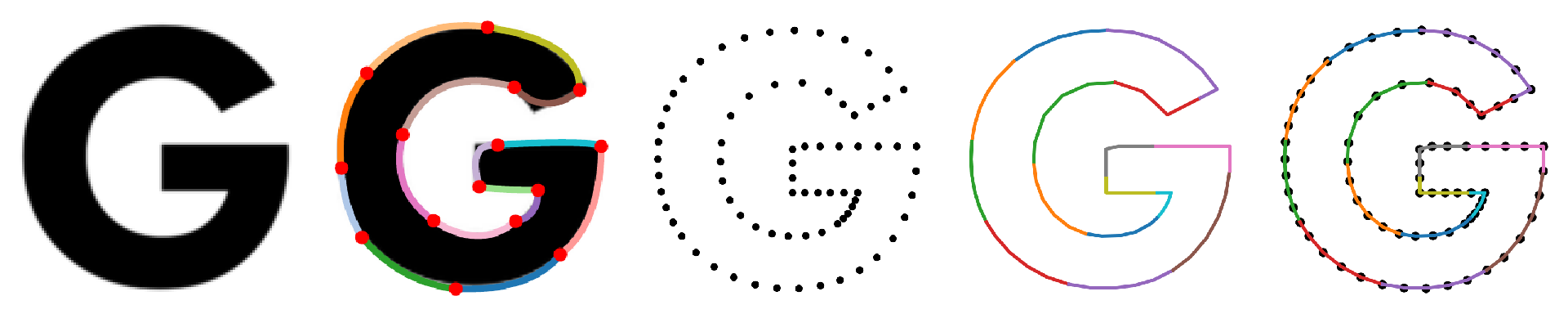} 
    \includegraphics[width=0.45\linewidth]{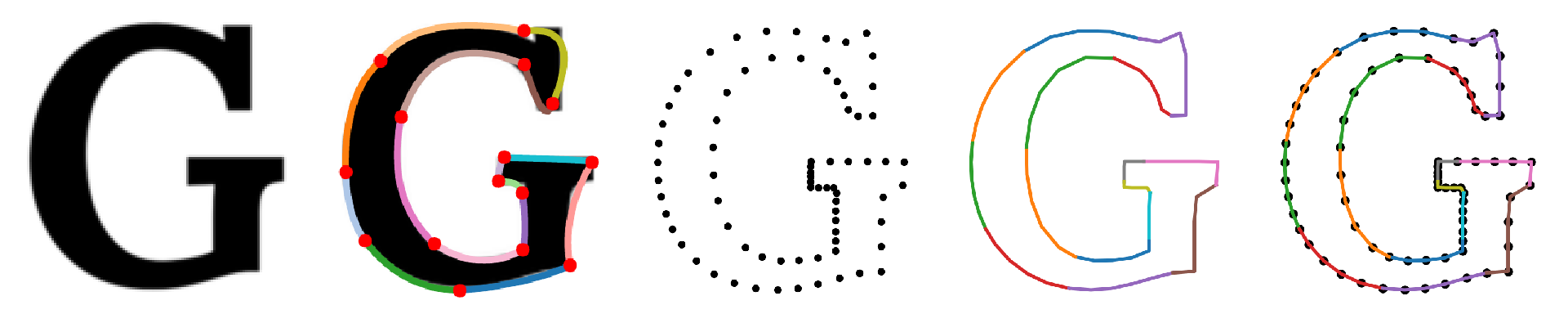} 
    \includegraphics[width=0.45\linewidth]{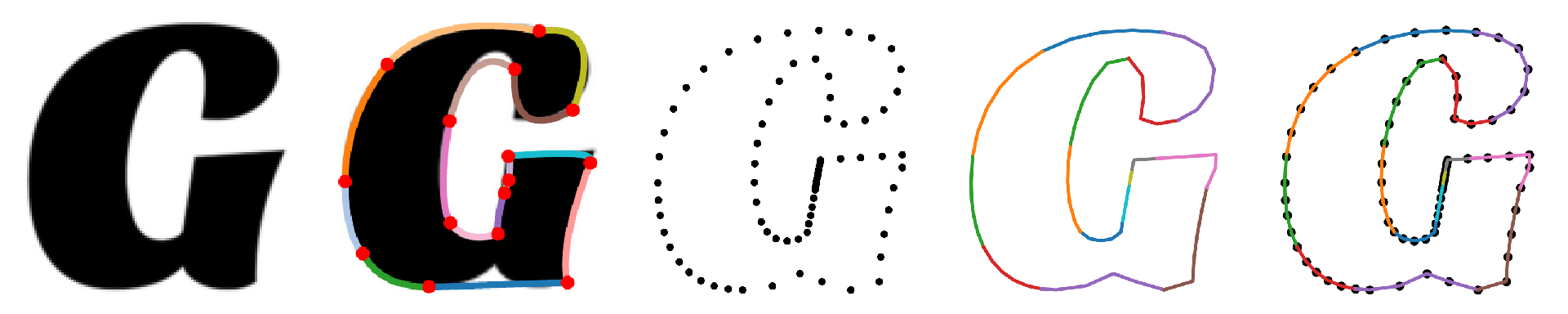} 
    \includegraphics[width=0.45\linewidth]{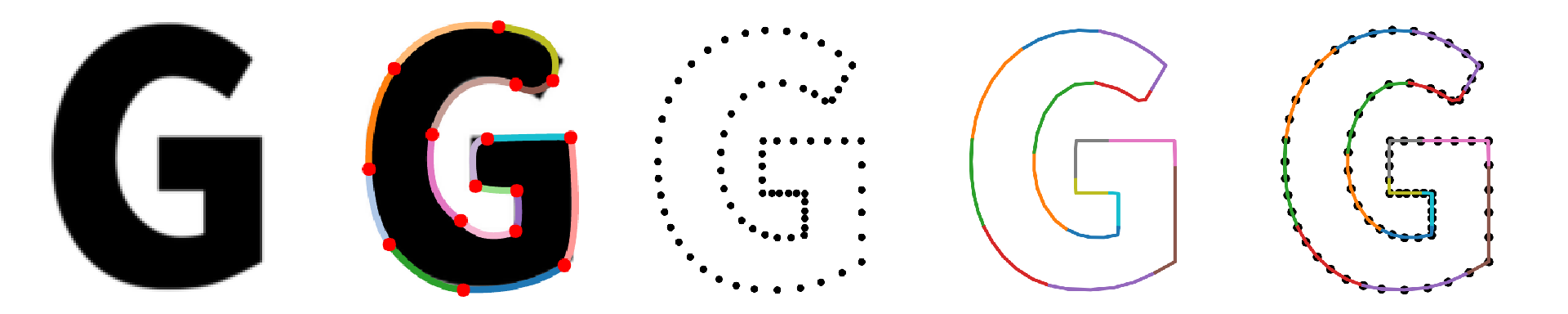} 
    \includegraphics[width=0.45\linewidth]{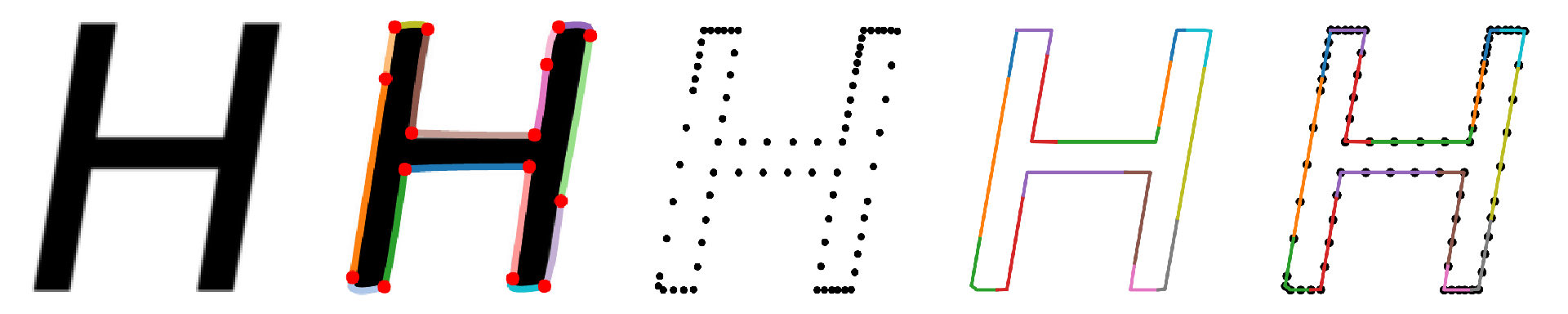} 
    \includegraphics[width=0.45\linewidth]{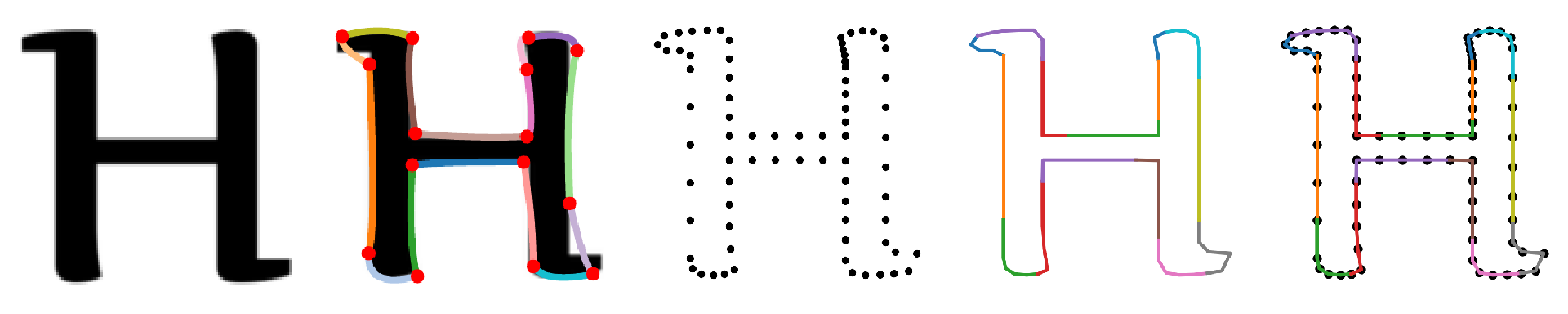} 
    \includegraphics[width=0.45\linewidth]{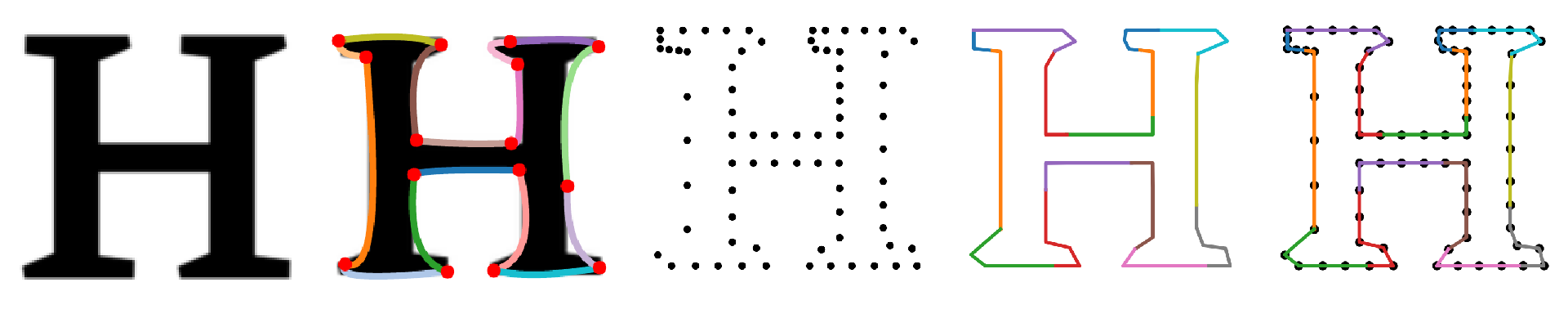} 
    \includegraphics[width=0.45\linewidth]{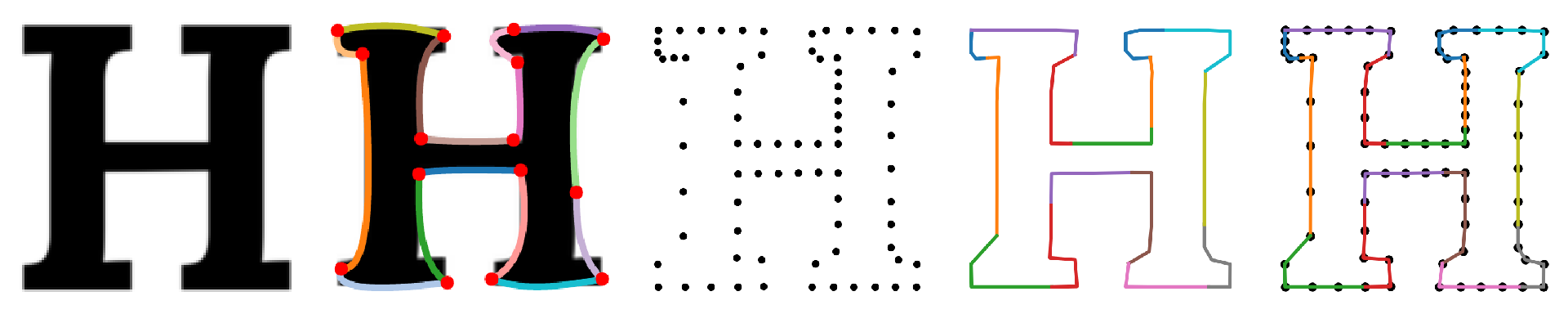} 
    \includegraphics[width=0.45\linewidth]{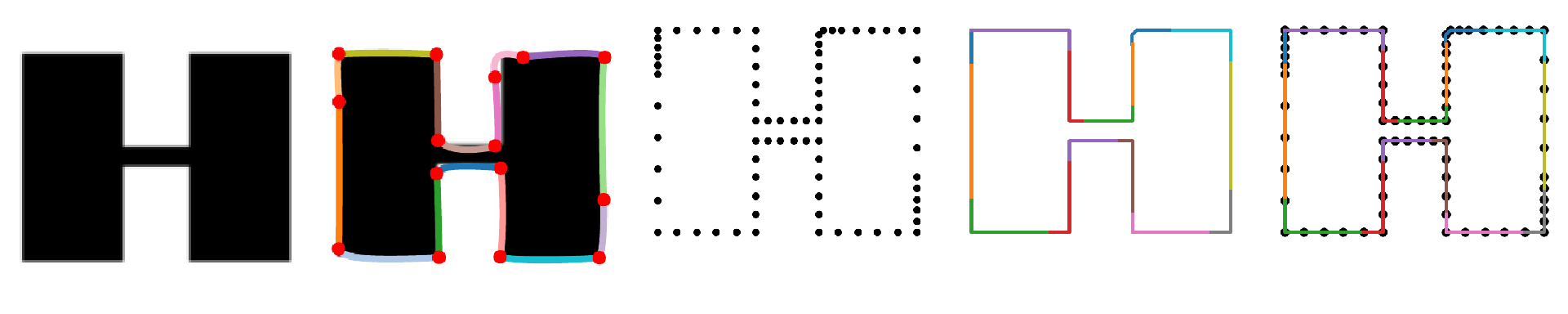} 
    \includegraphics[width=0.45\linewidth]{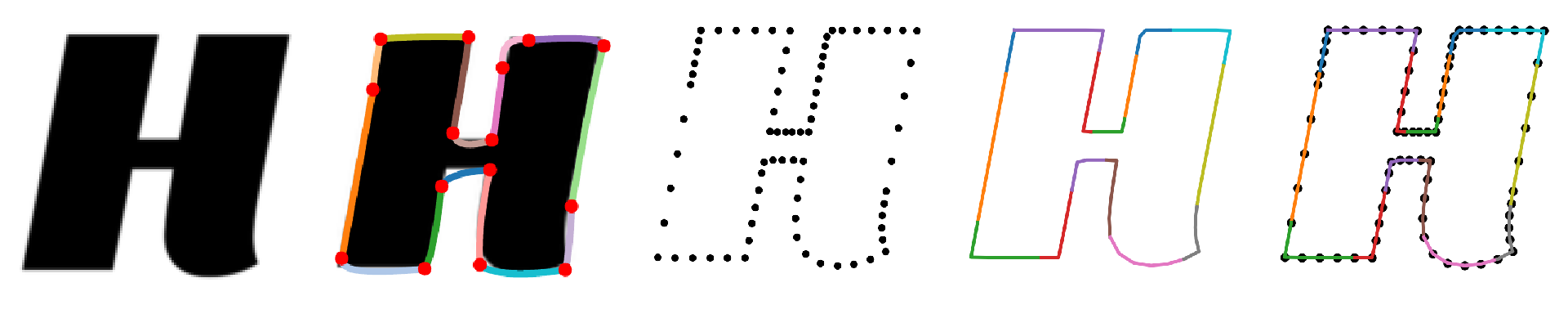} 
    \includegraphics[width=0.45\linewidth]{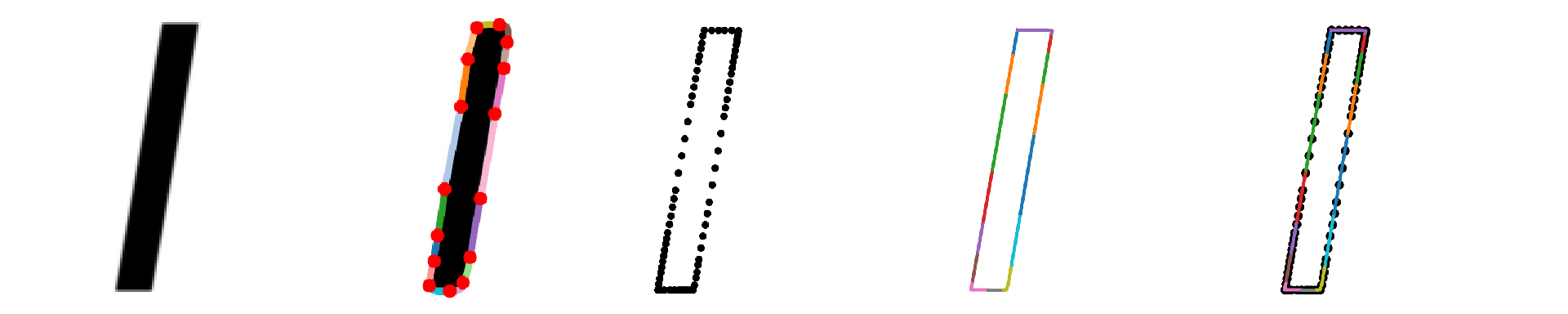} 
    \includegraphics[width=0.45\linewidth]{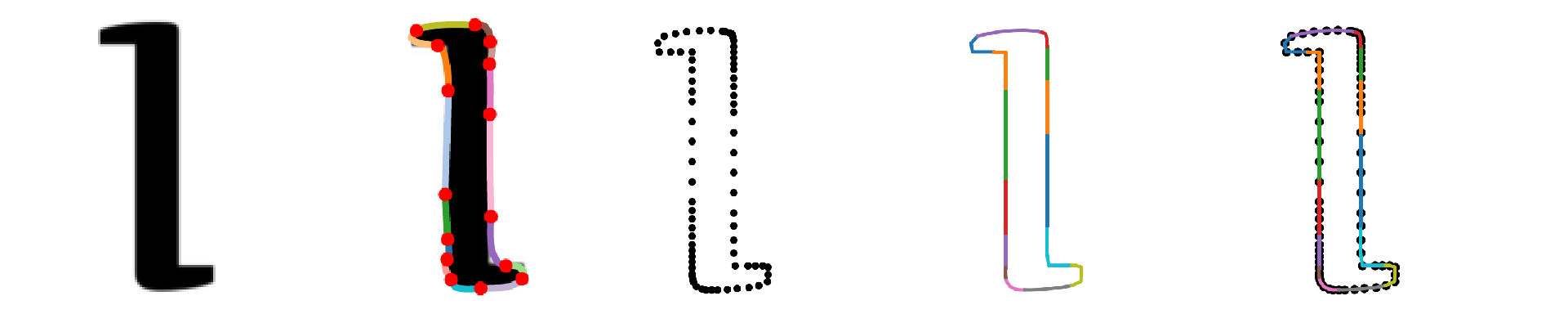} 
    \includegraphics[width=0.45\linewidth]{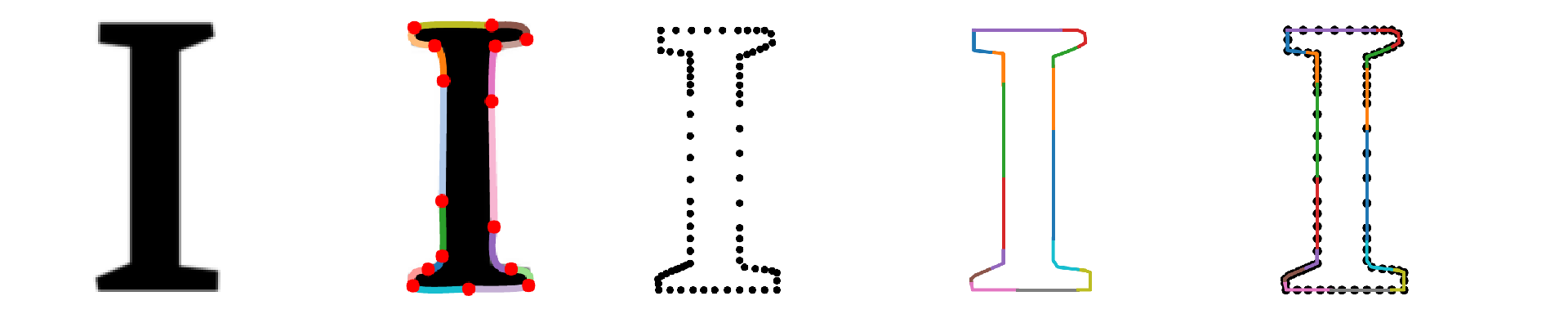} 
    \includegraphics[width=0.45\linewidth]{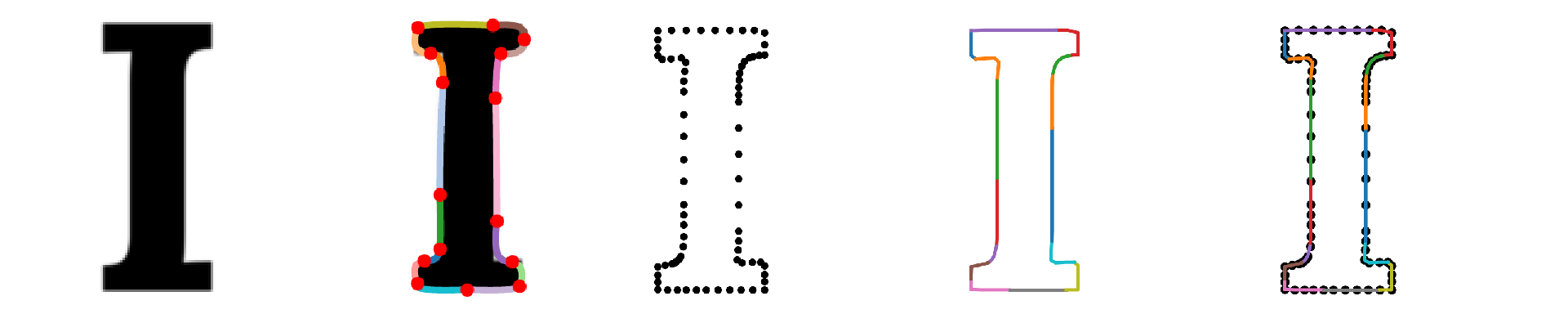} 
    \includegraphics[width=0.45\linewidth]{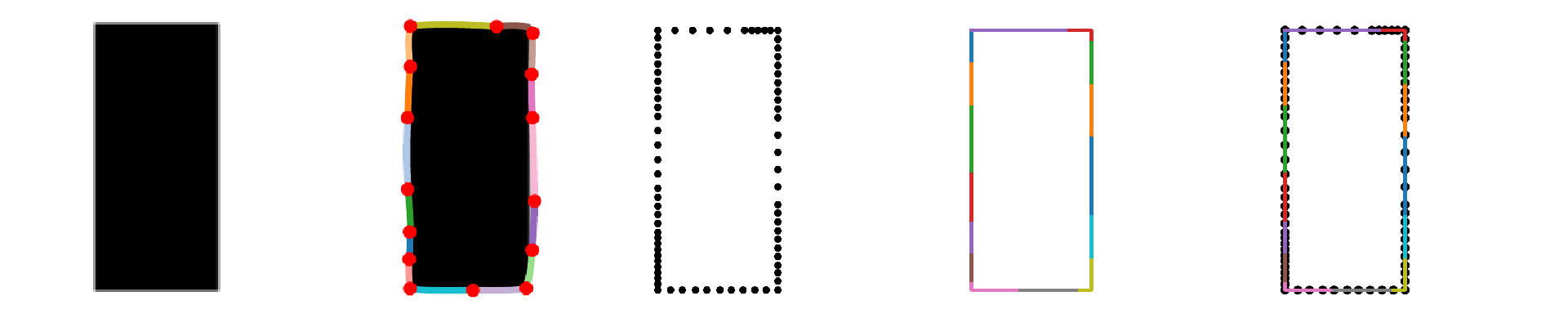} 
    \includegraphics[width=0.45\linewidth]{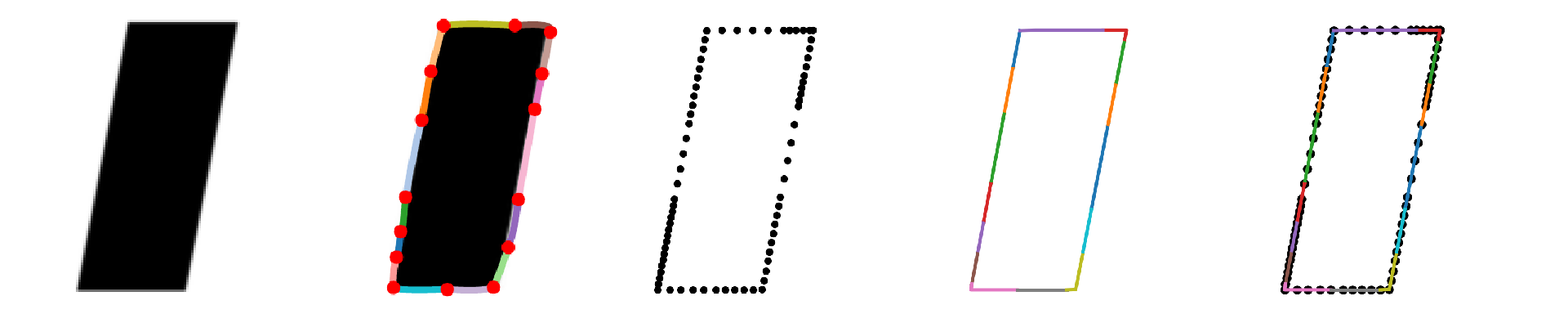} 
    \includegraphics[width=0.45\linewidth]{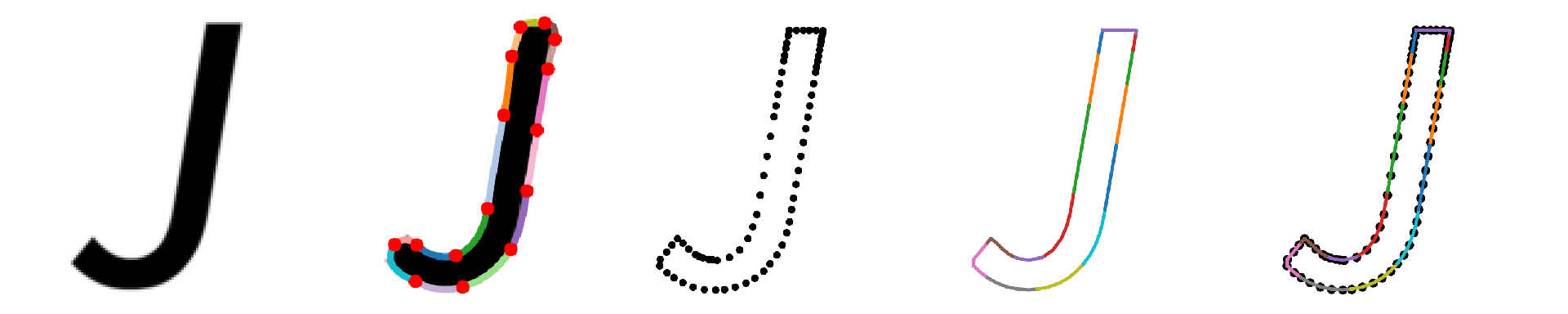} 
    \includegraphics[width=0.45\linewidth]{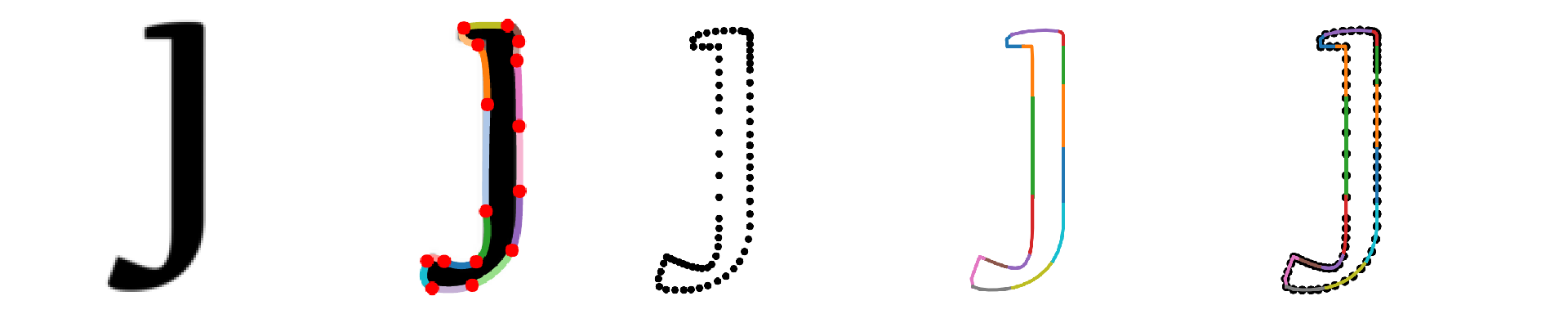} 
    \includegraphics[width=0.45\linewidth]{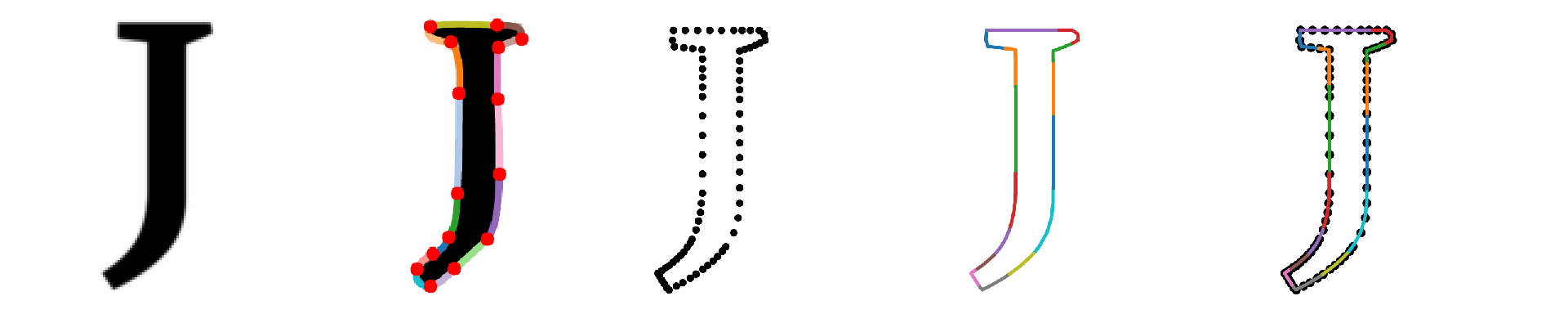} 
    \includegraphics[width=0.45\linewidth]{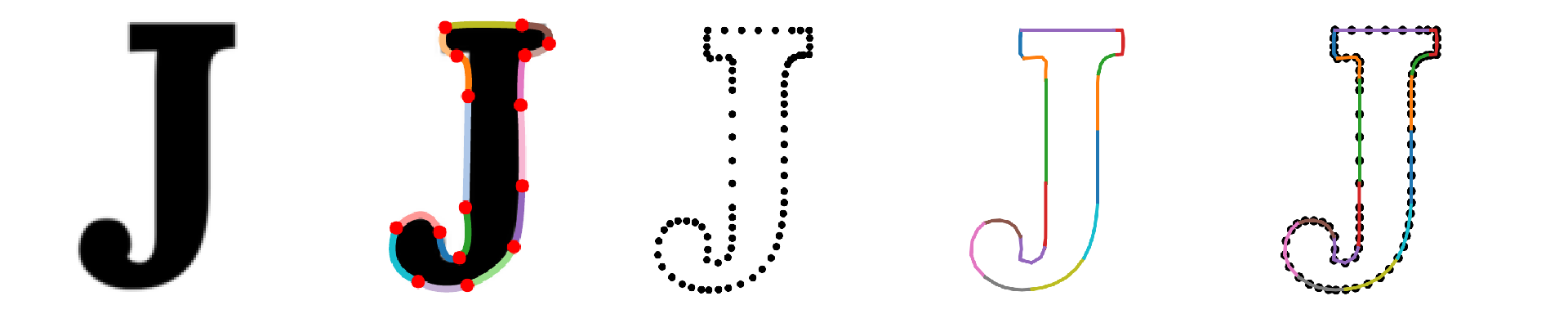} 
    \includegraphics[width=0.45\linewidth]{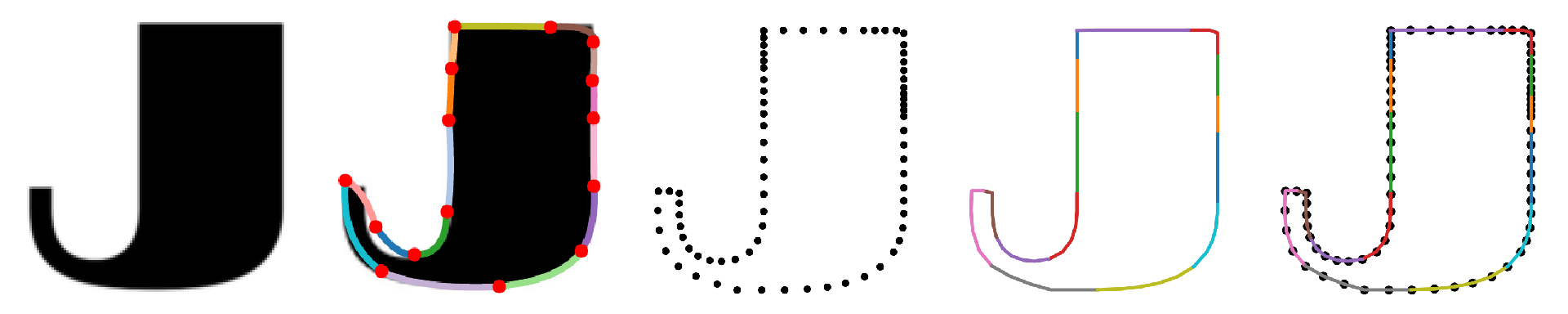} 
    \includegraphics[width=0.45\linewidth]{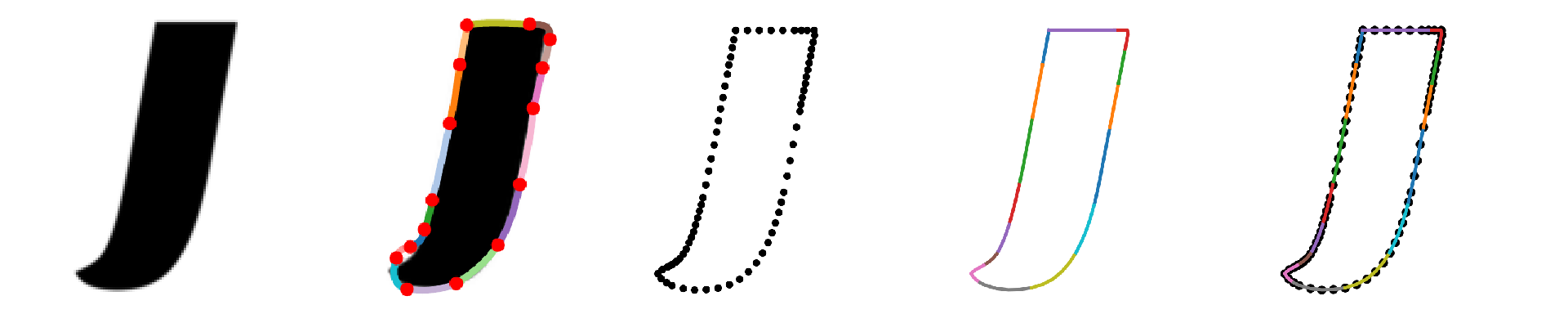} 
    \includegraphics[width=0.45\linewidth]{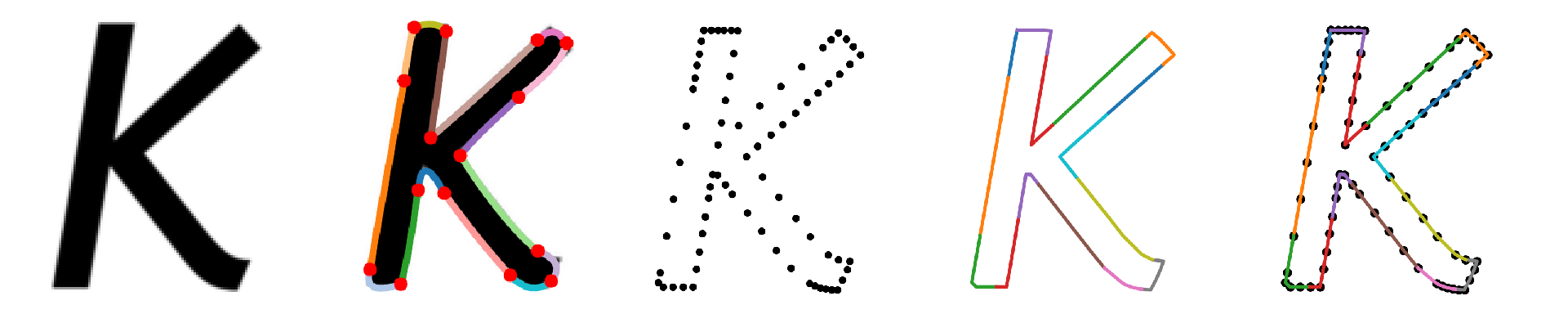} 
    \includegraphics[width=0.45\linewidth]{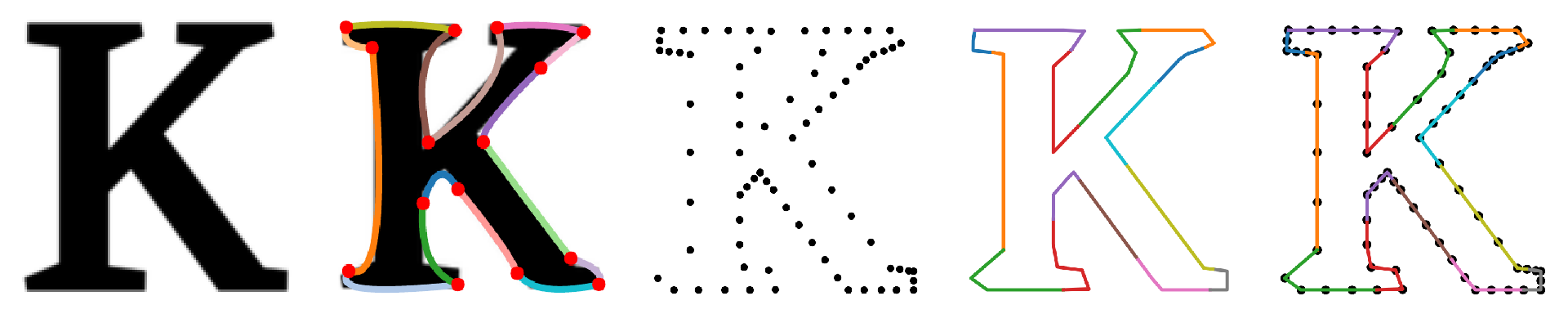} 
    \includegraphics[width=0.45\linewidth]{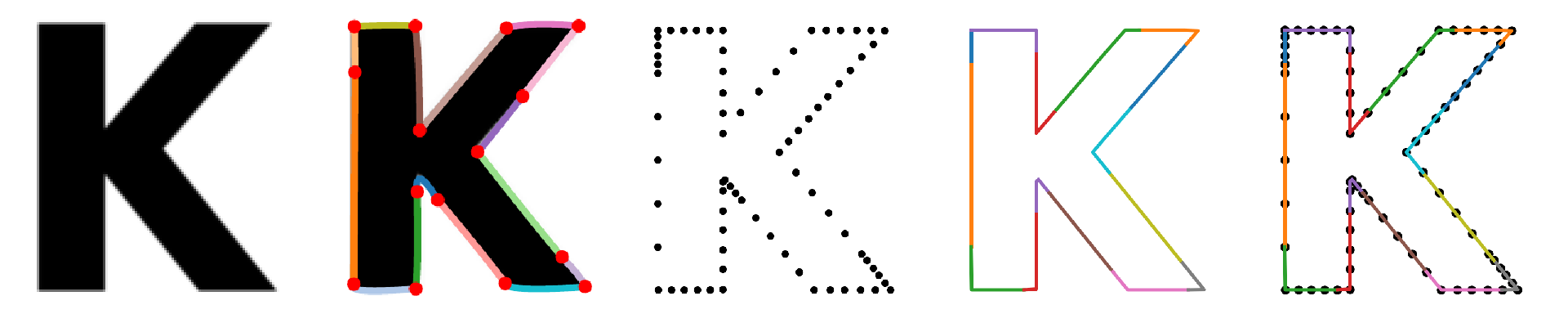} 
    \includegraphics[width=0.45\linewidth]{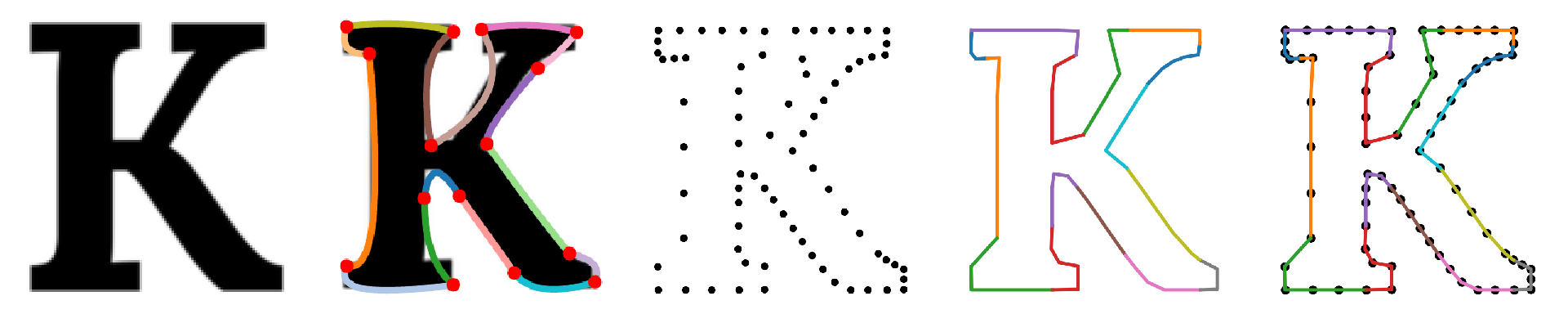} 
    \includegraphics[width=0.45\linewidth]{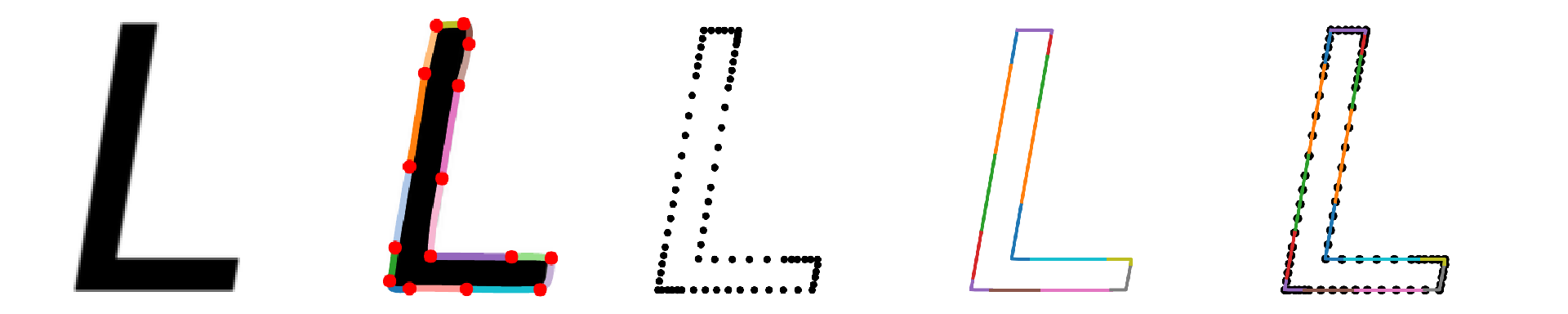} 
    \includegraphics[width=0.45\linewidth]{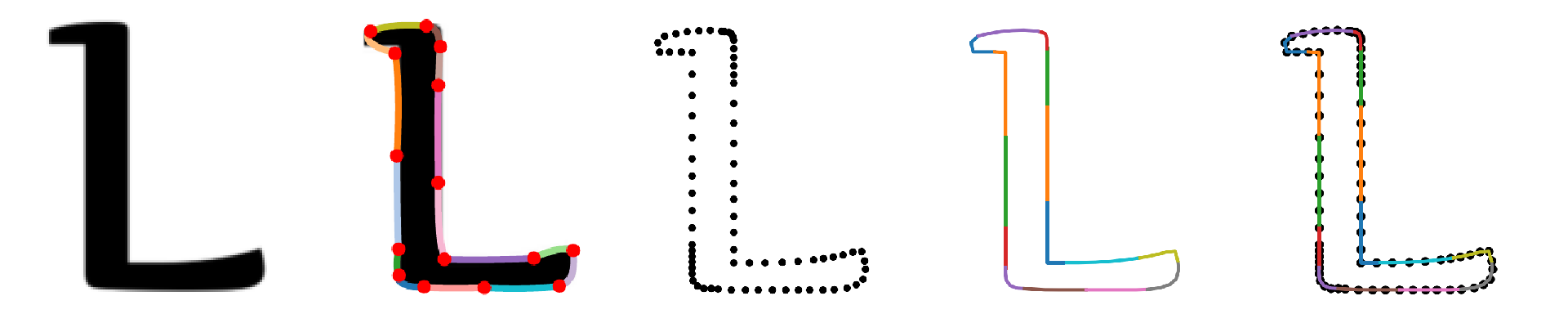} 
    \includegraphics[width=0.45\linewidth]{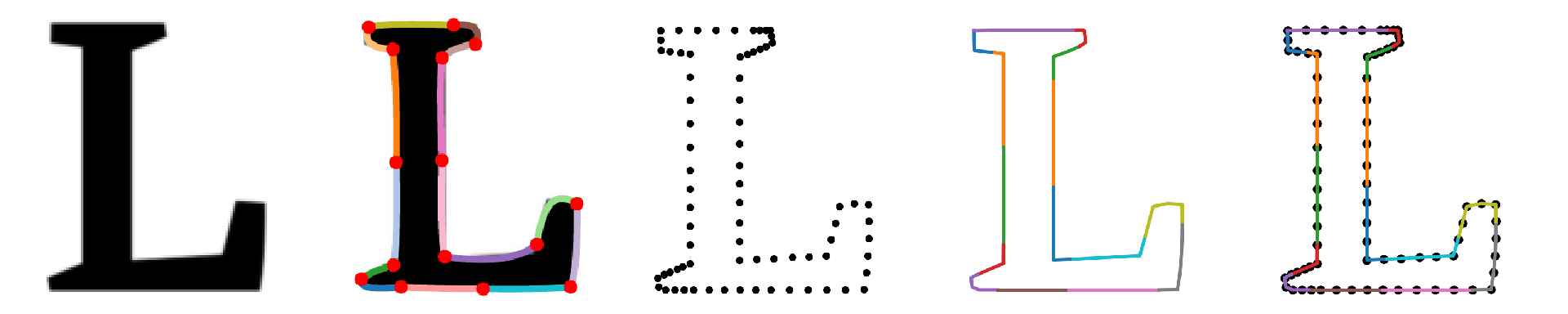} 
    \includegraphics[width=0.45\linewidth]{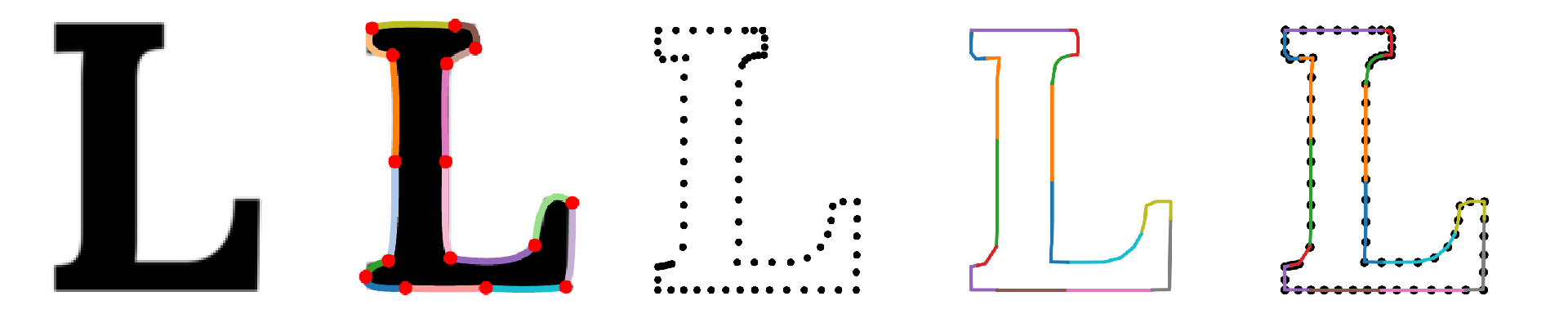} 
    \includegraphics[width=0.45\linewidth]{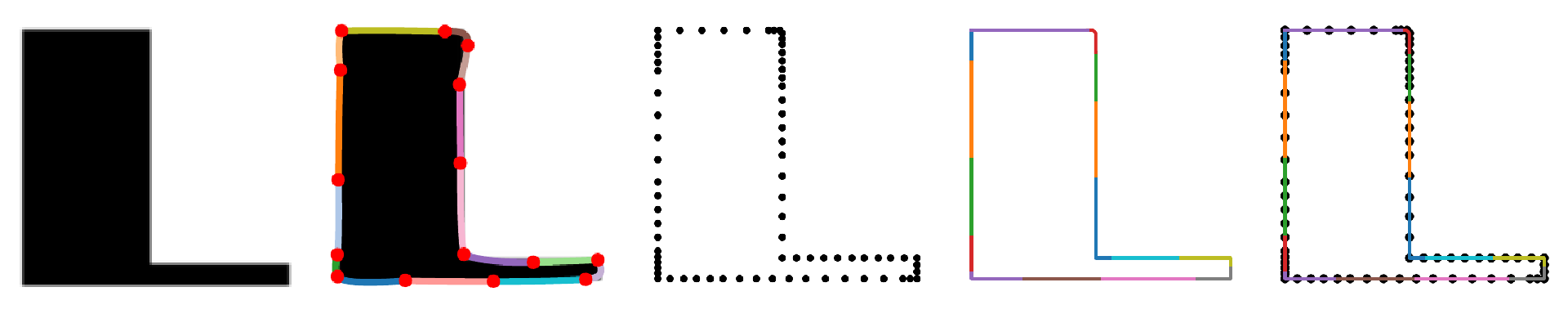} 
    \includegraphics[width=0.45\linewidth]{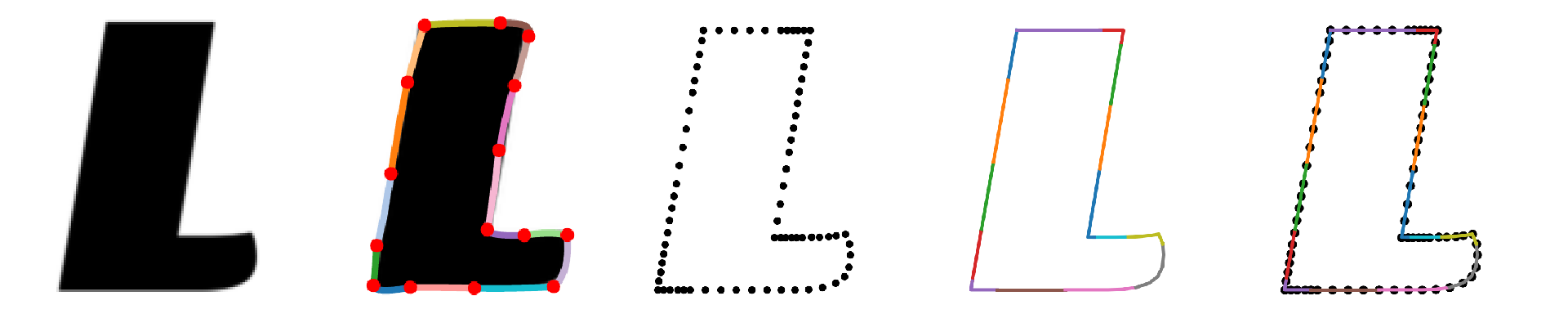} 
    \includegraphics[width=0.45\linewidth]{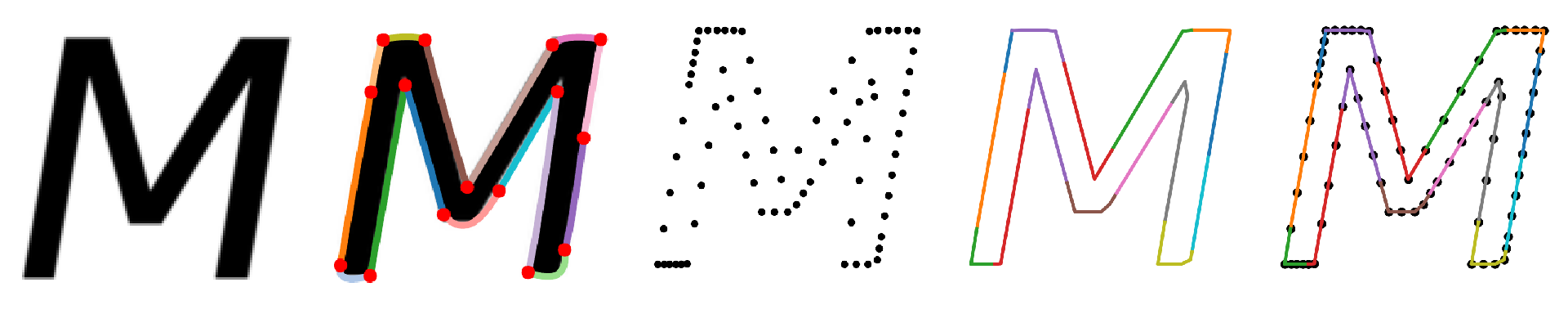} 
    \includegraphics[width=0.45\linewidth]{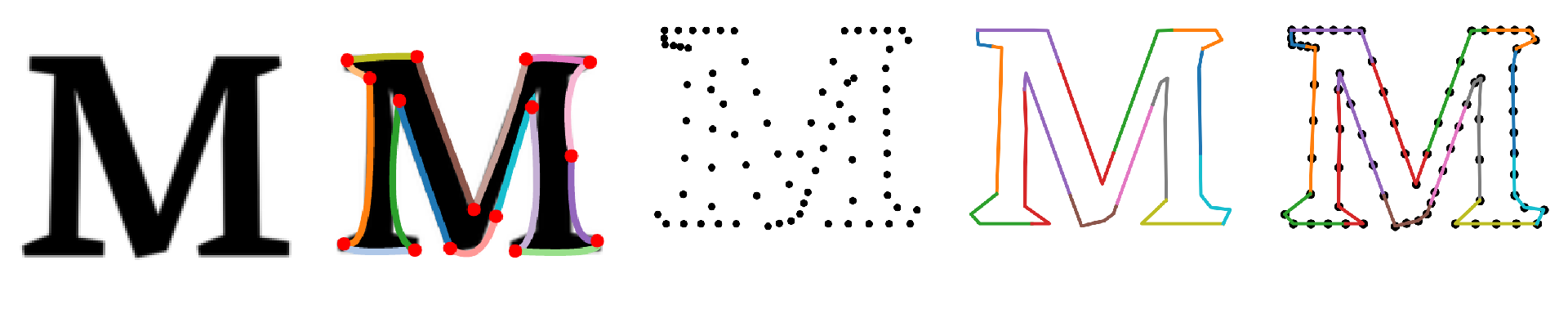} 
    \includegraphics[width=0.45\linewidth]{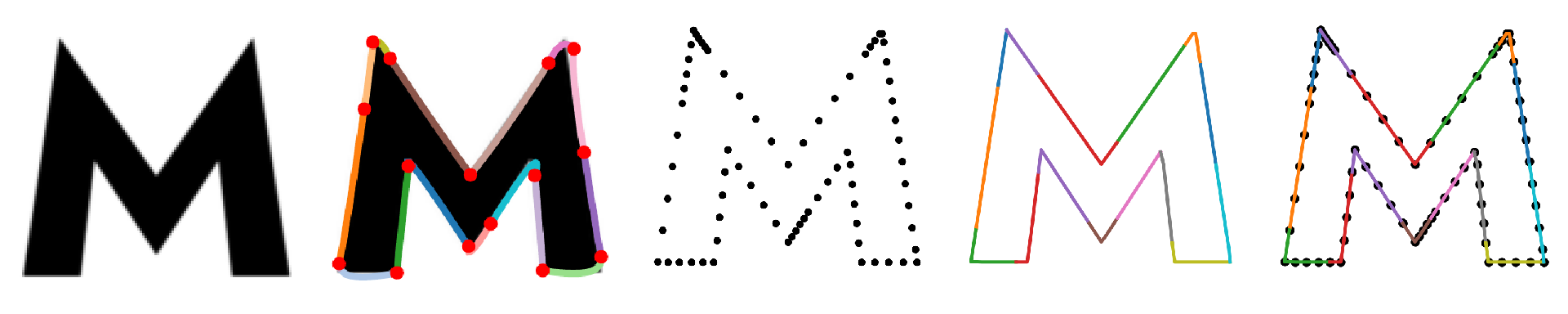} 
    \includegraphics[width=0.45\linewidth]{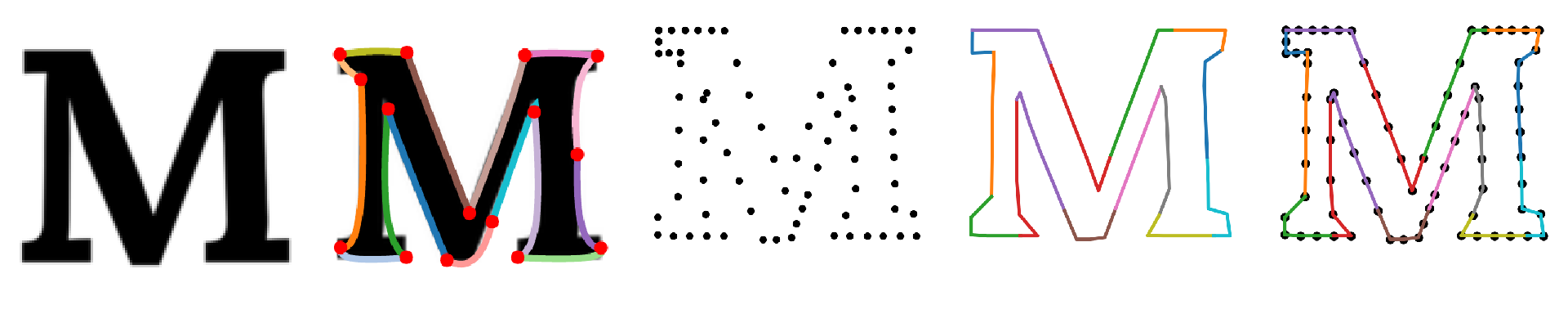} 
    \includegraphics[width=0.45\linewidth]{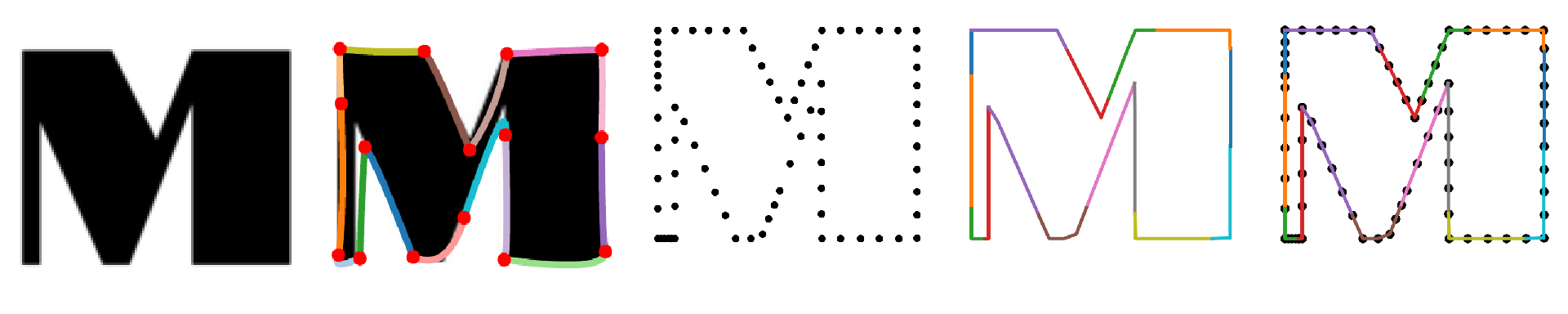} 
    \includegraphics[width=0.45\linewidth]{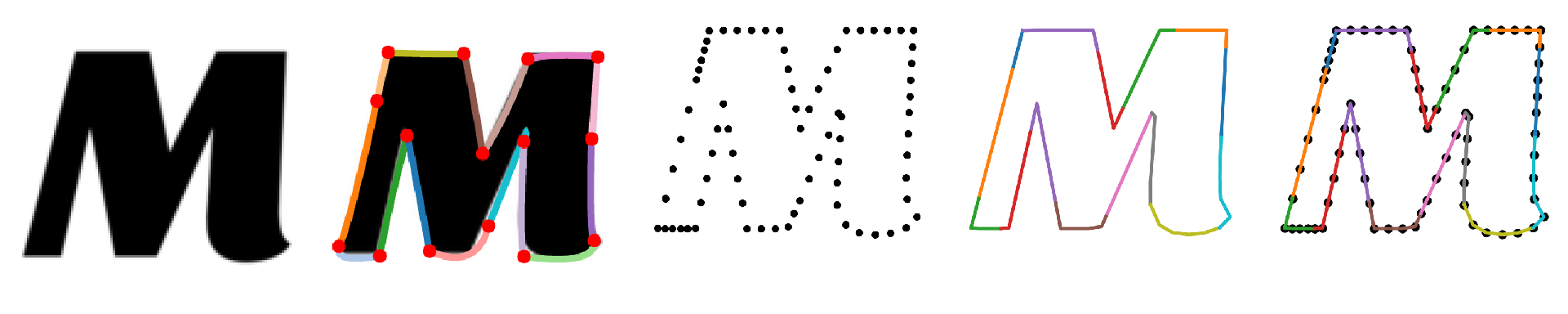} 
    \includegraphics[width=0.45\linewidth]{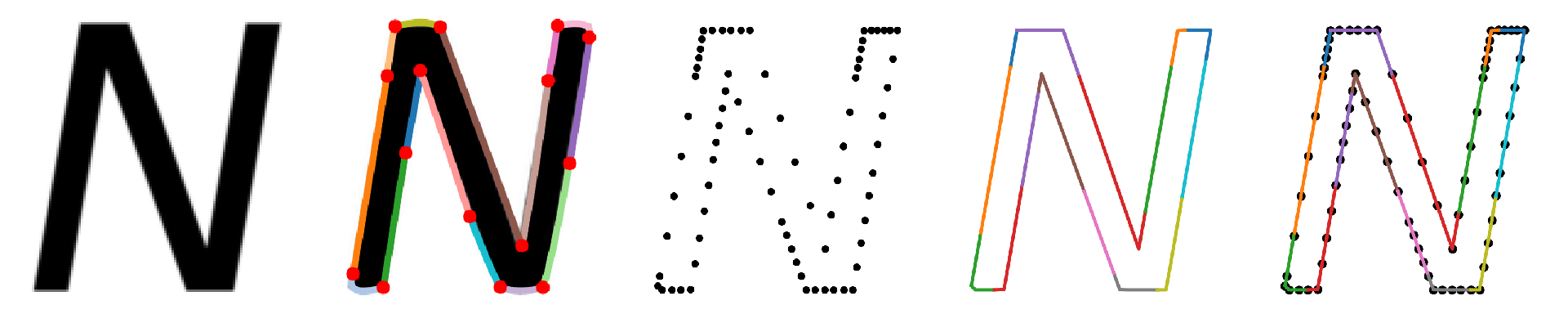} 
    \includegraphics[width=0.45\linewidth]{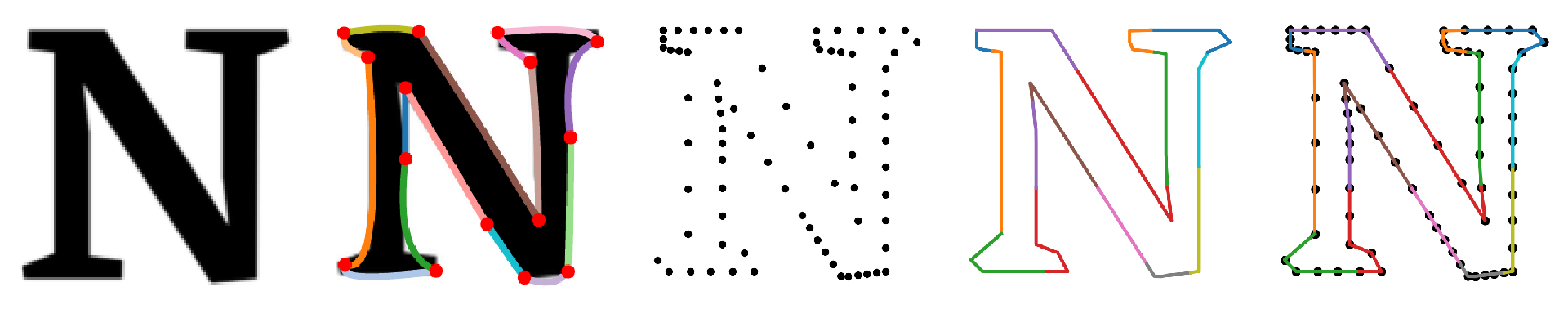} 
    \includegraphics[width=0.45\linewidth]{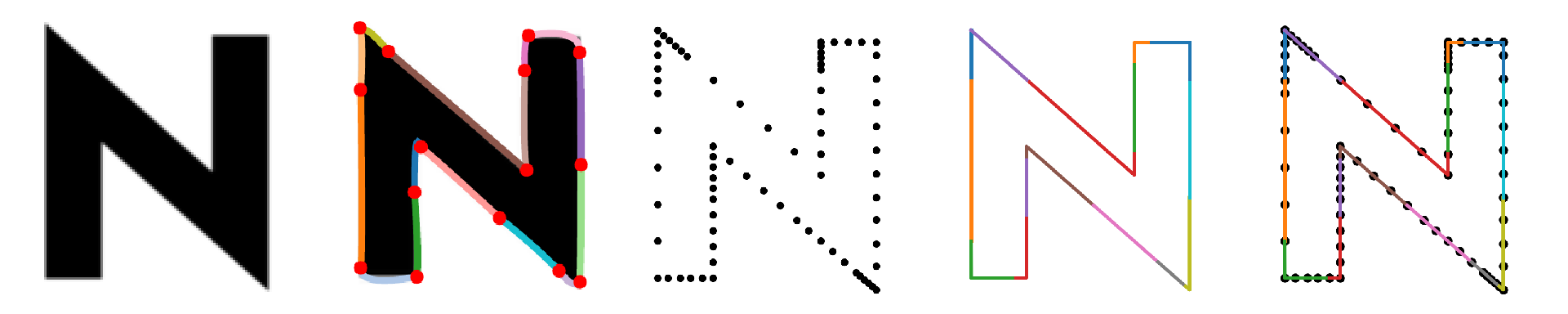} 
    \includegraphics[width=0.45\linewidth]{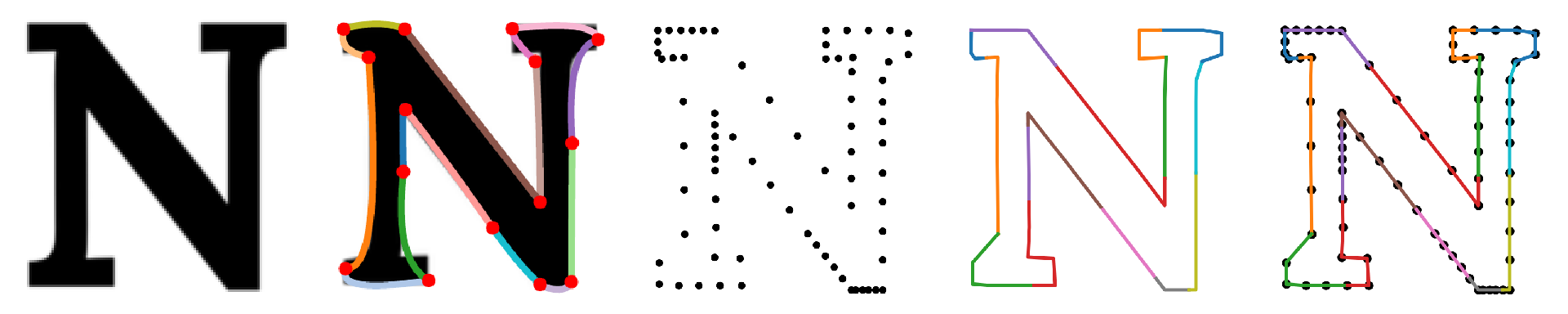} 
    \includegraphics[width=0.45\linewidth]{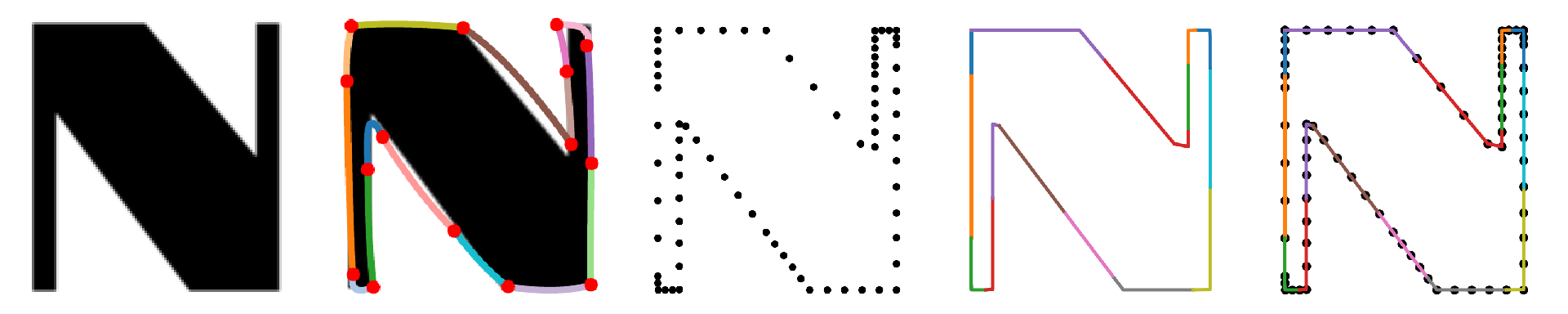} 
    \includegraphics[width=0.45\linewidth]{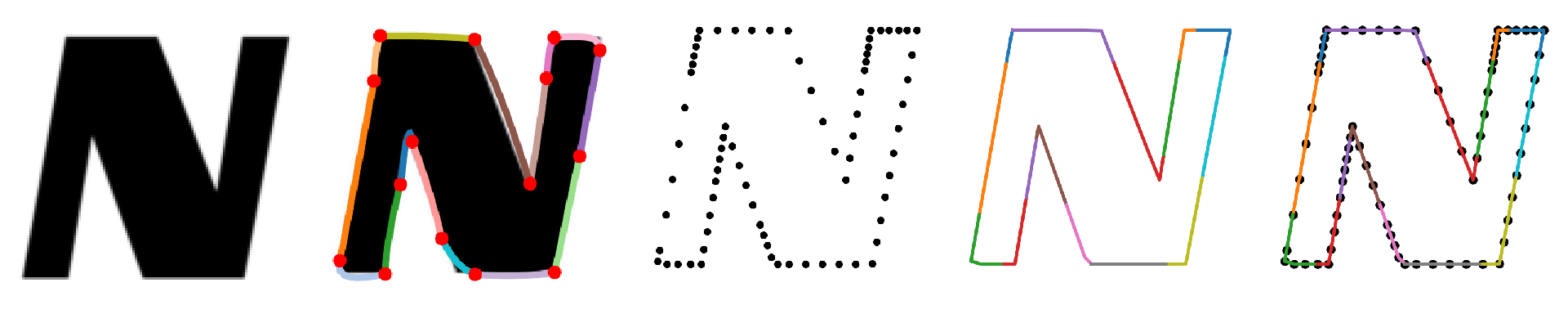} 
    \includegraphics[width=0.45\linewidth]{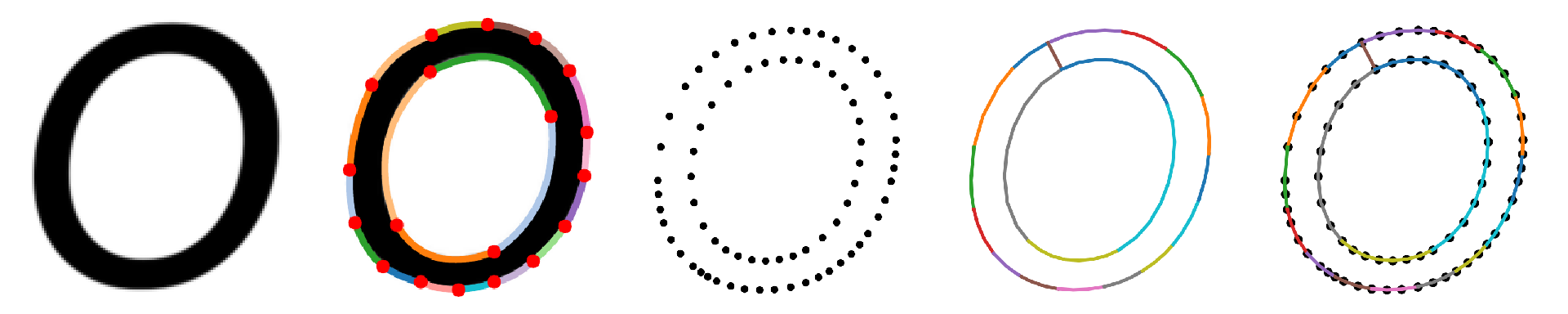} 
    \includegraphics[width=0.45\linewidth]{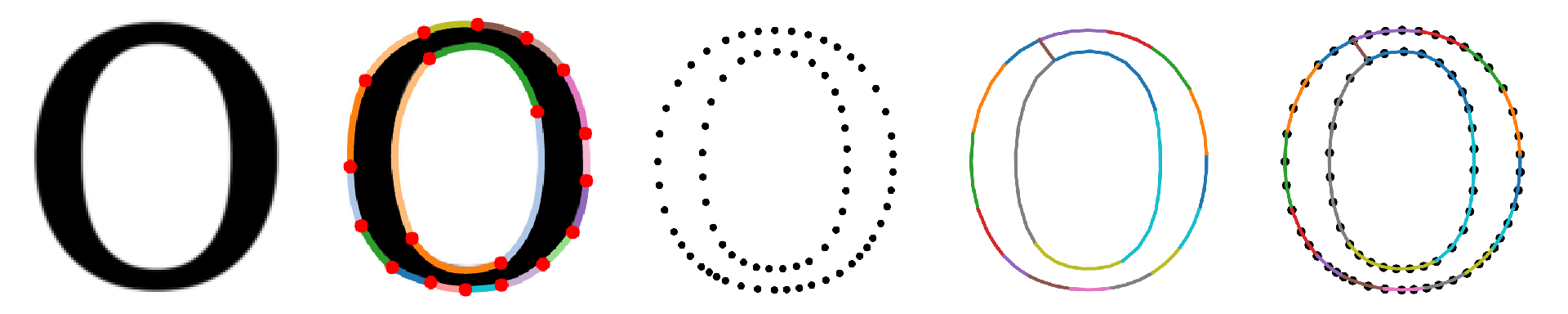} 
    \includegraphics[width=0.45\linewidth]{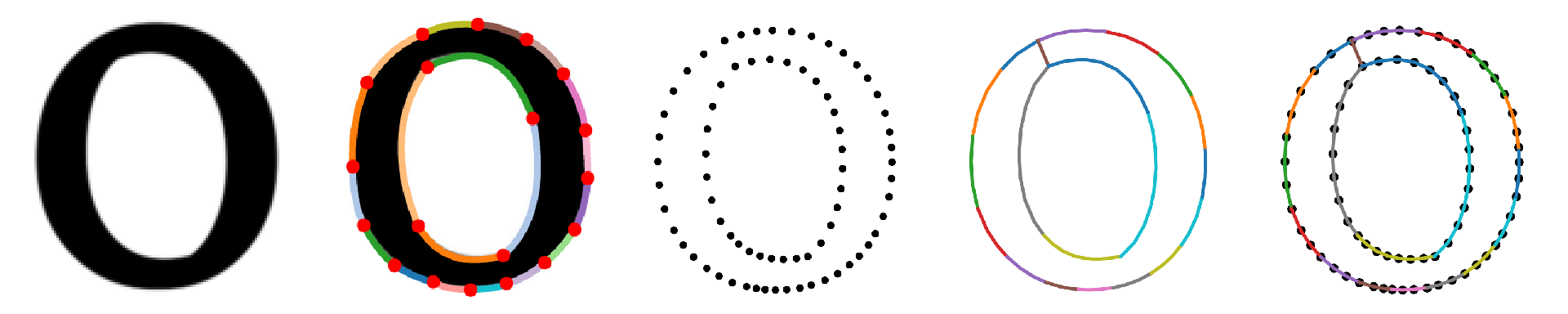} 
    \includegraphics[width=0.45\linewidth]{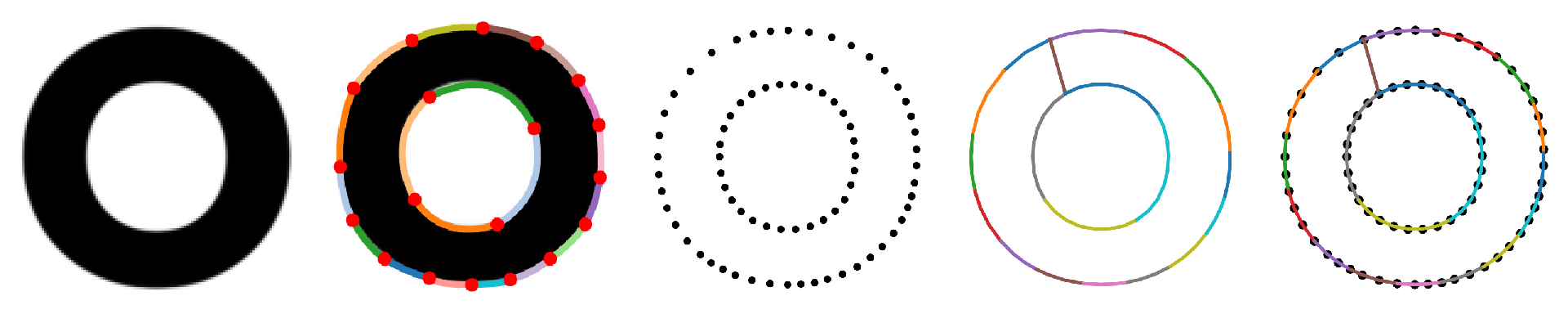} 
    \includegraphics[width=0.45\linewidth]{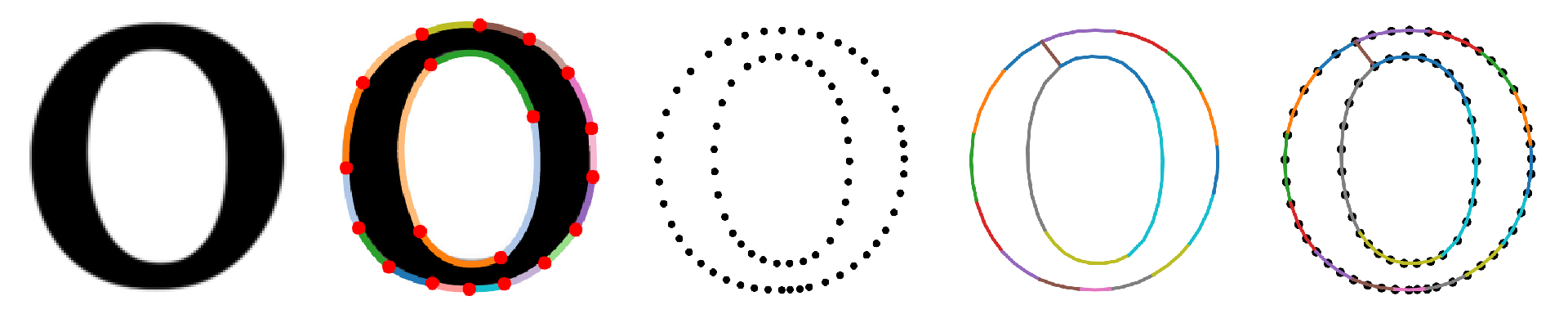} 
    \includegraphics[width=0.45\linewidth]{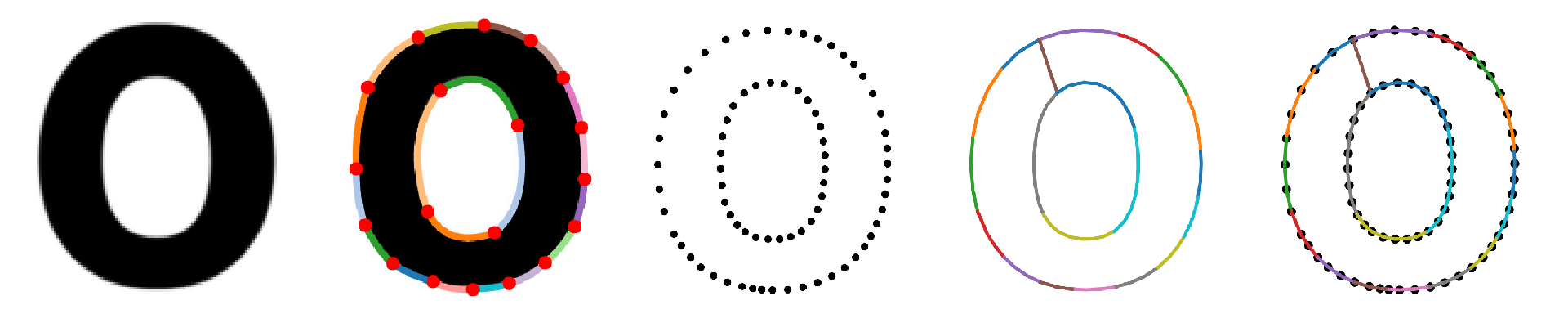}
    \includegraphics[width=0.45\linewidth]{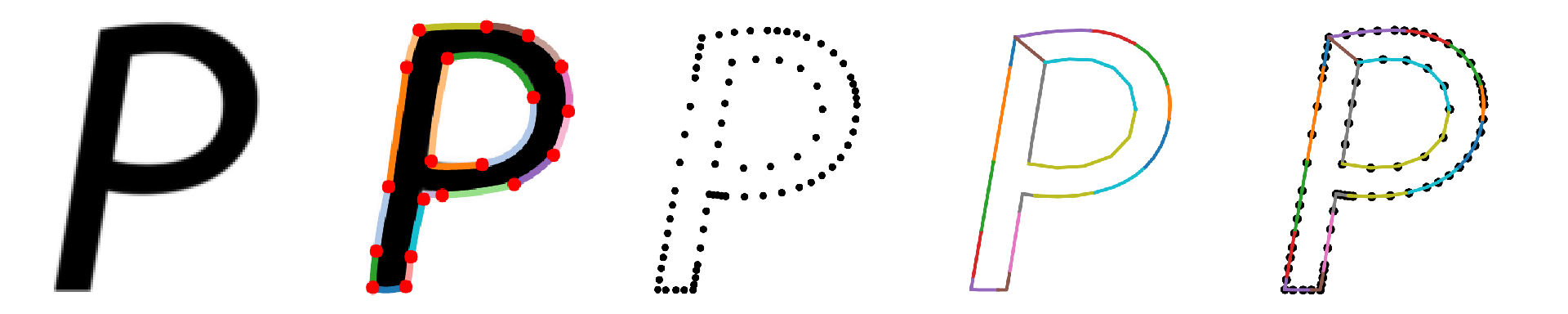} 
    \includegraphics[width=0.45\linewidth]{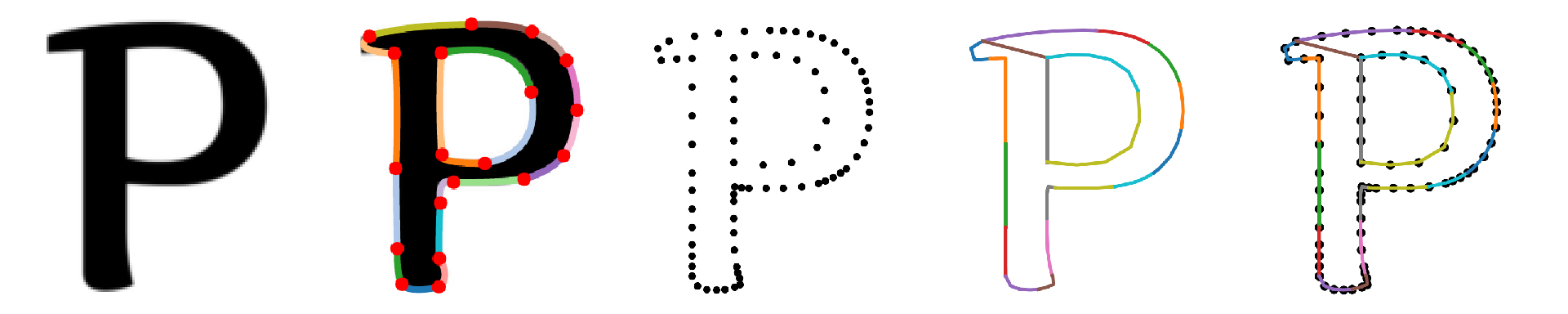} 
    \includegraphics[width=0.45\linewidth]{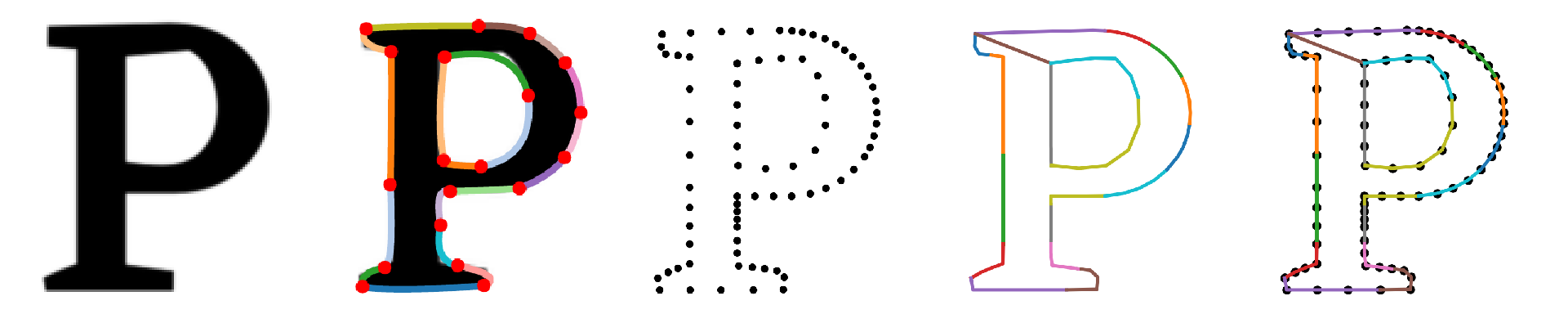} 
    \includegraphics[width=0.45\linewidth]{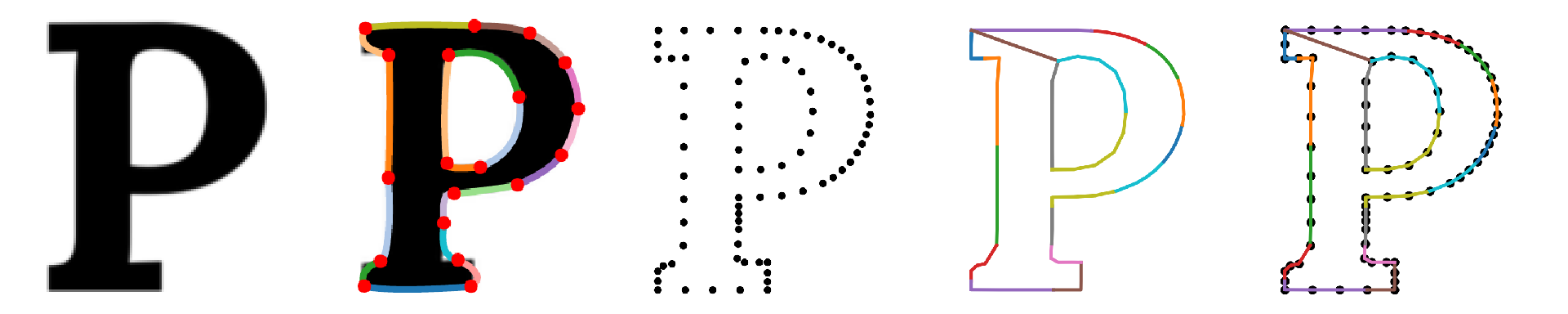} \vspace{-0.5em}
\end{center}
\caption{Different data modalities of font glyphs. For each example from left to right: image, keypoints, point set, graph template and directed graph representation. (To be continued.)} \vspace{-0.5em}
\label{graph_construction}
\end{figure}

\begin{figure}[h!]\ContinuedFloat
\begin{center}
    \includegraphics[width=0.45\linewidth]{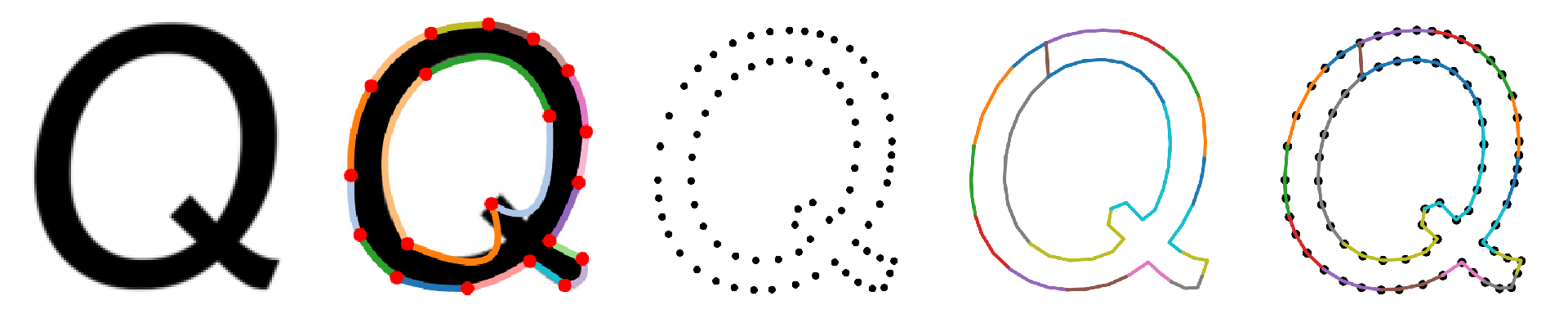} 
    \includegraphics[width=0.45\linewidth]{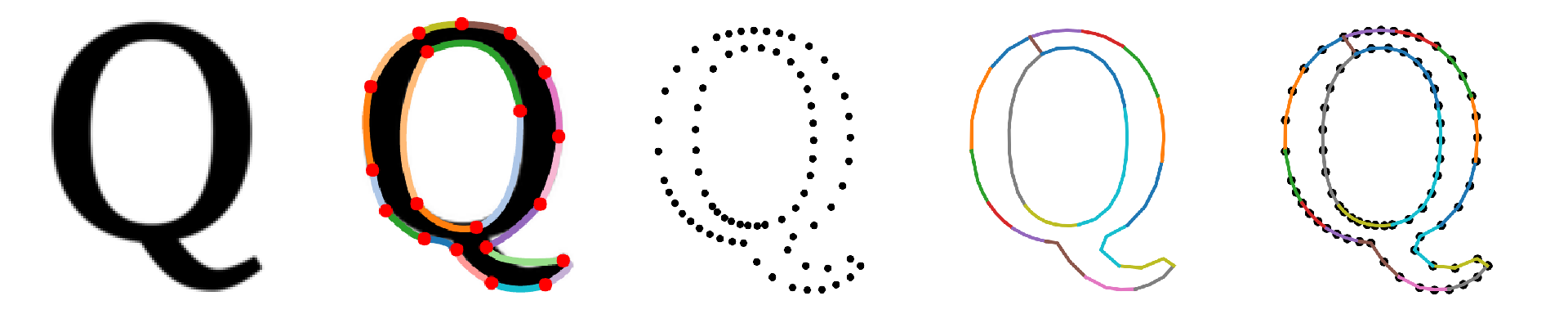} 
    \includegraphics[width=0.45\linewidth]{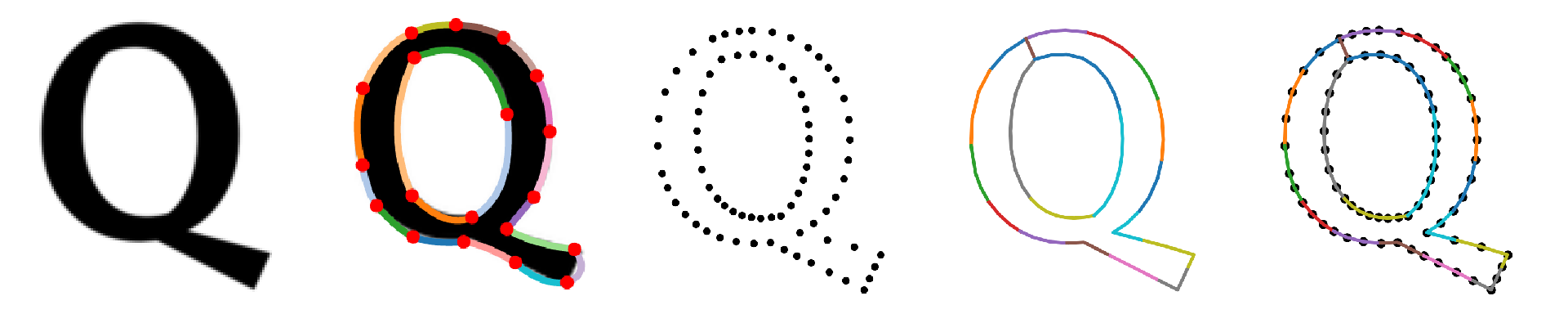} 
    \includegraphics[width=0.45\linewidth]{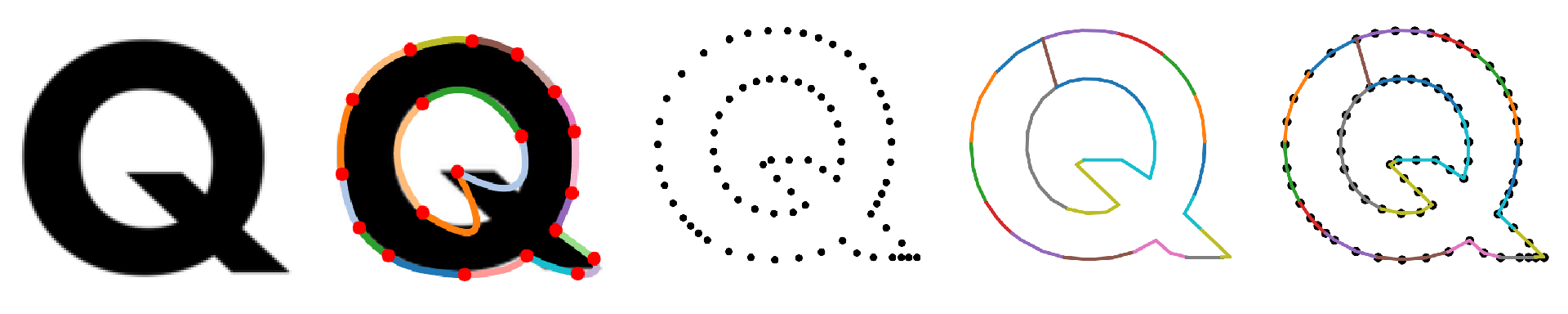} 
    \includegraphics[width=0.45\linewidth]{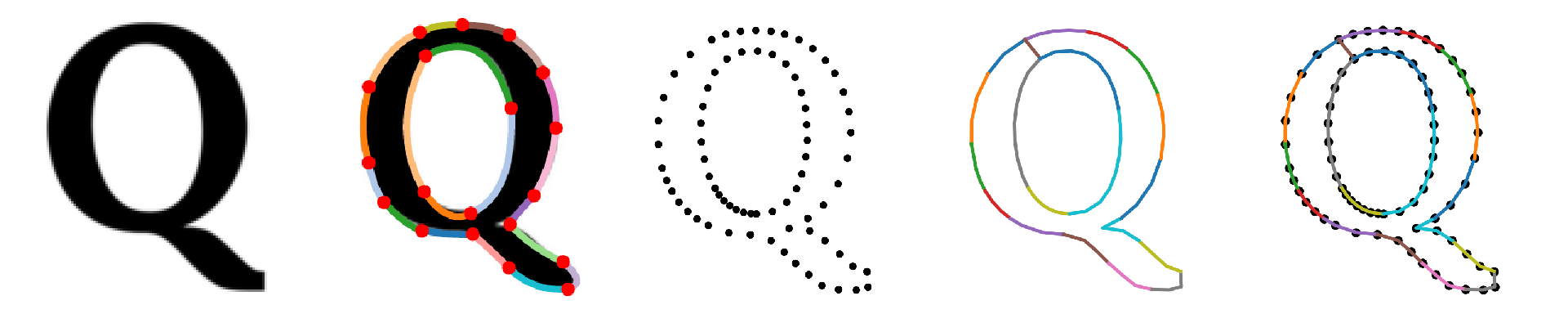} 
    \includegraphics[width=0.45\linewidth]{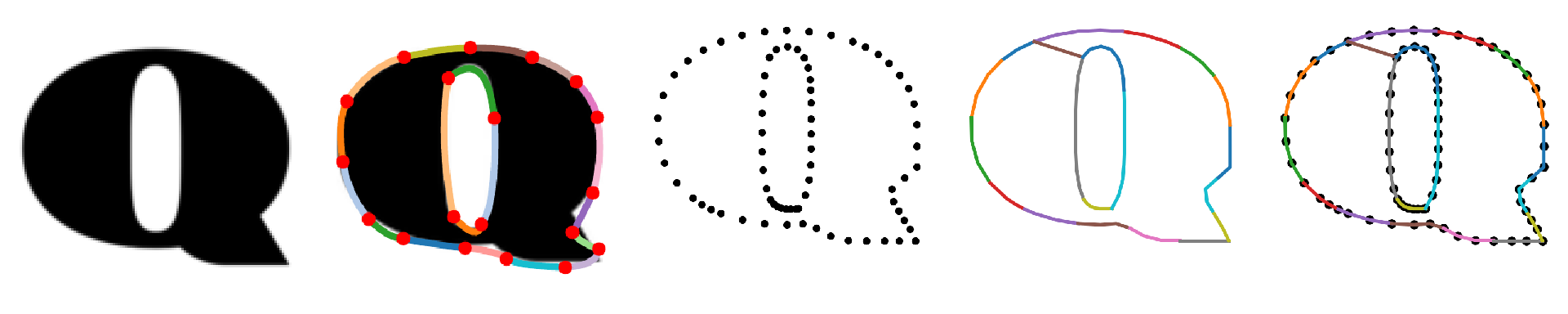} 
    \includegraphics[width=0.45\linewidth]{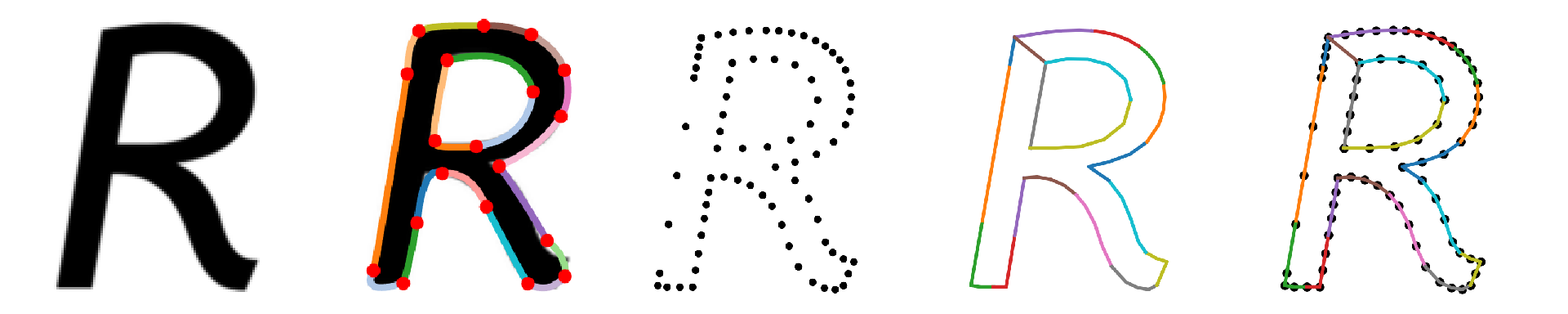} 
    \includegraphics[width=0.45\linewidth]{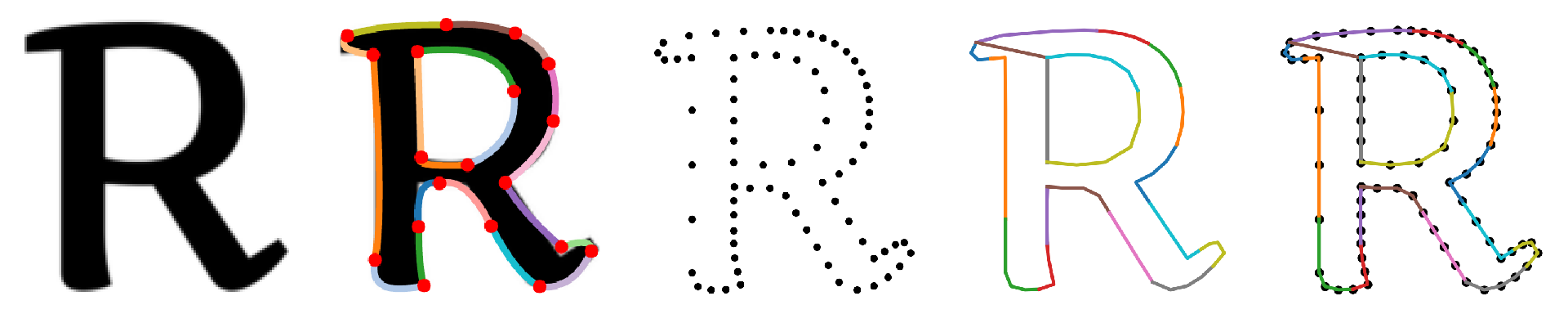} 
    \includegraphics[width=0.45\linewidth]{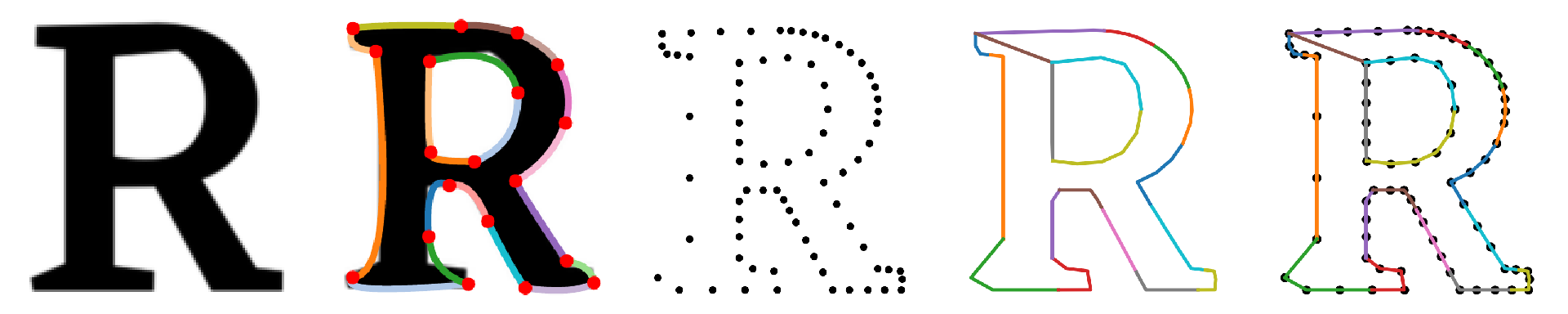} 
    \includegraphics[width=0.45\linewidth]{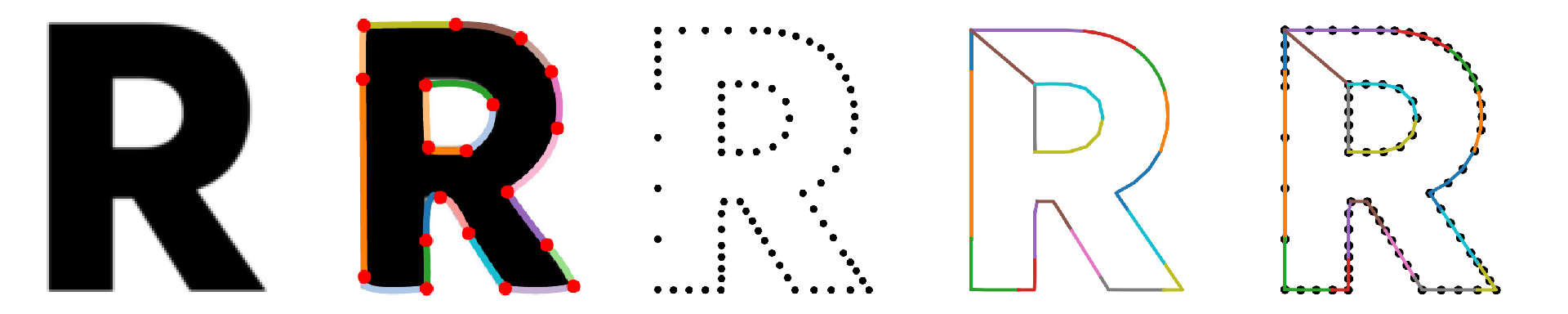} 
    \includegraphics[width=0.45\linewidth]{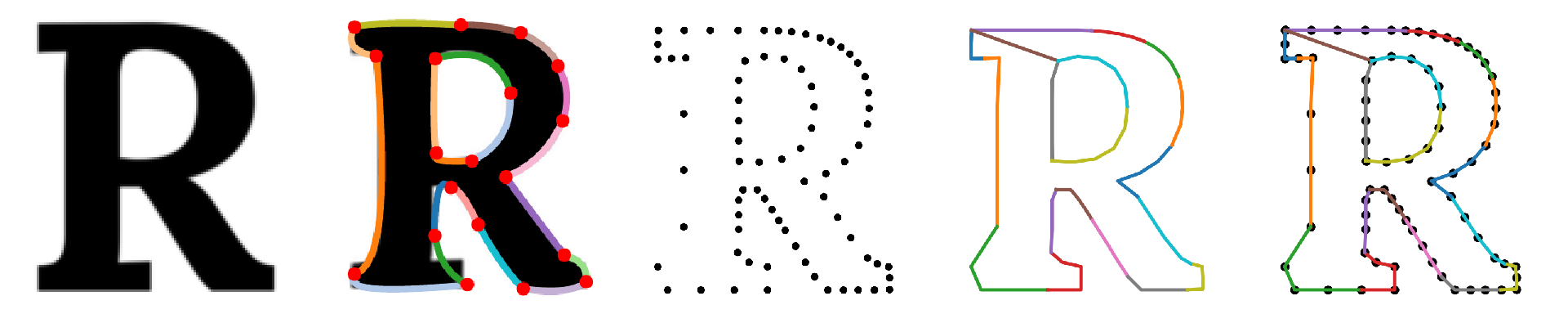} 
    \includegraphics[width=0.45\linewidth]{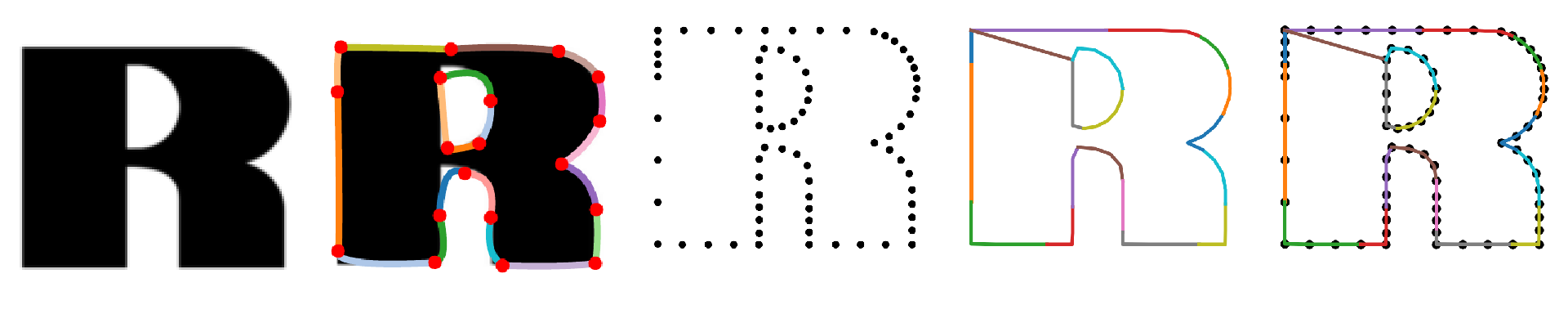} 
    \includegraphics[width=0.45\linewidth]{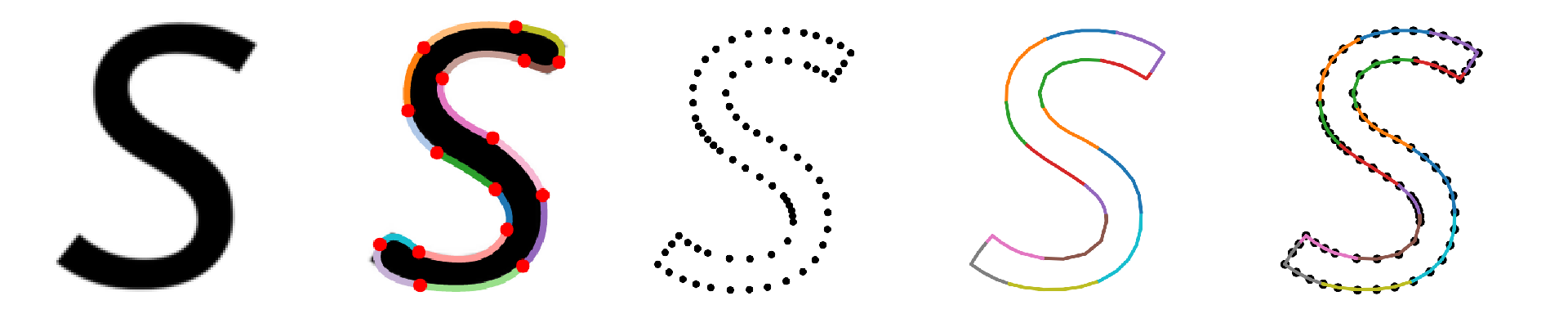} 
    \includegraphics[width=0.45\linewidth]{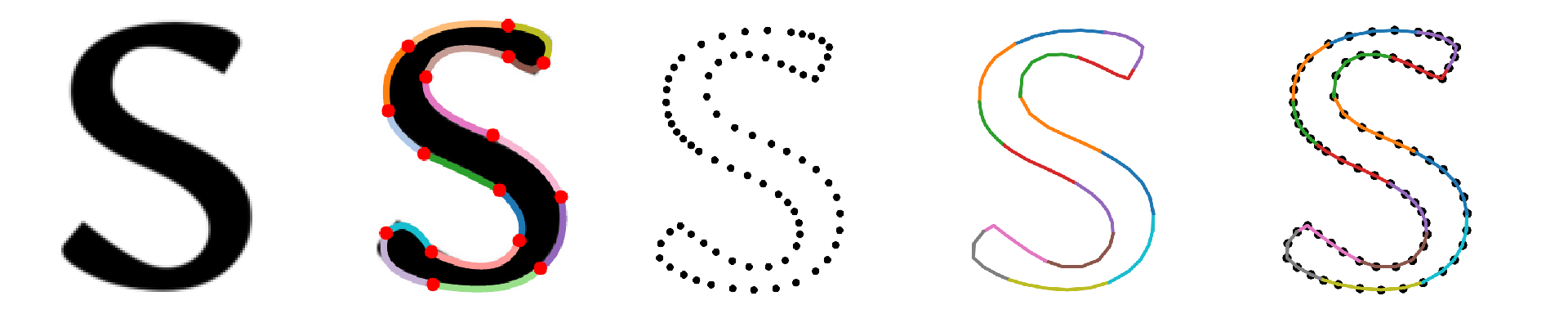} 
    \includegraphics[width=0.45\linewidth]{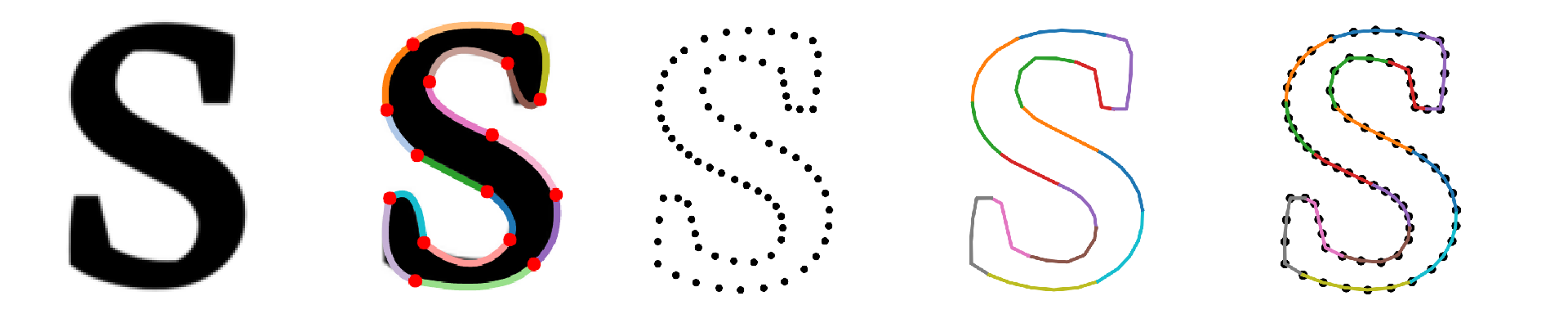} 
    \includegraphics[width=0.45\linewidth]{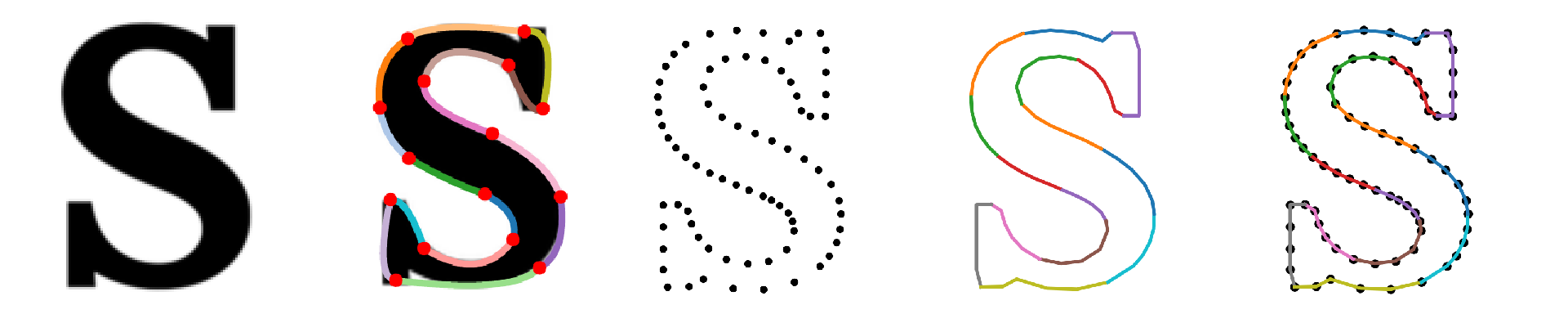} 
    \includegraphics[width=0.45\linewidth]{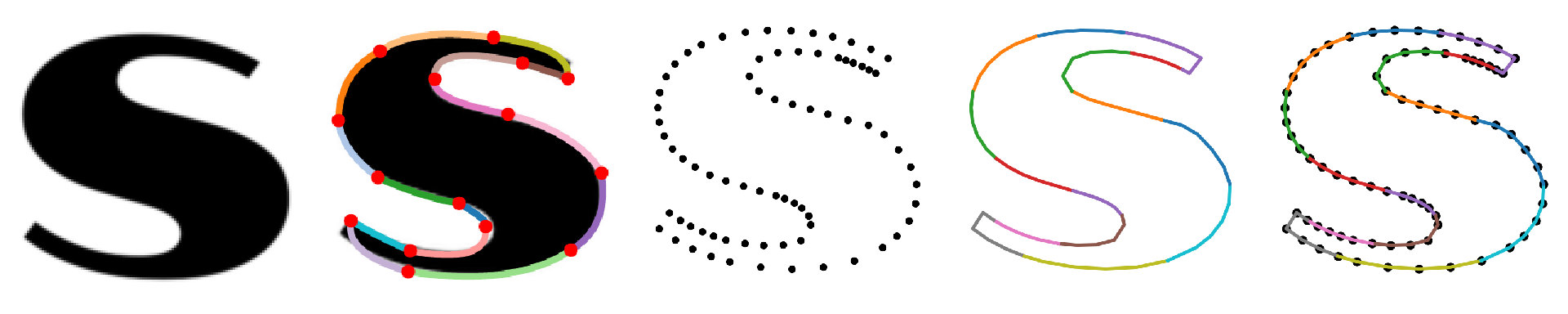} 
    \includegraphics[width=0.45\linewidth]{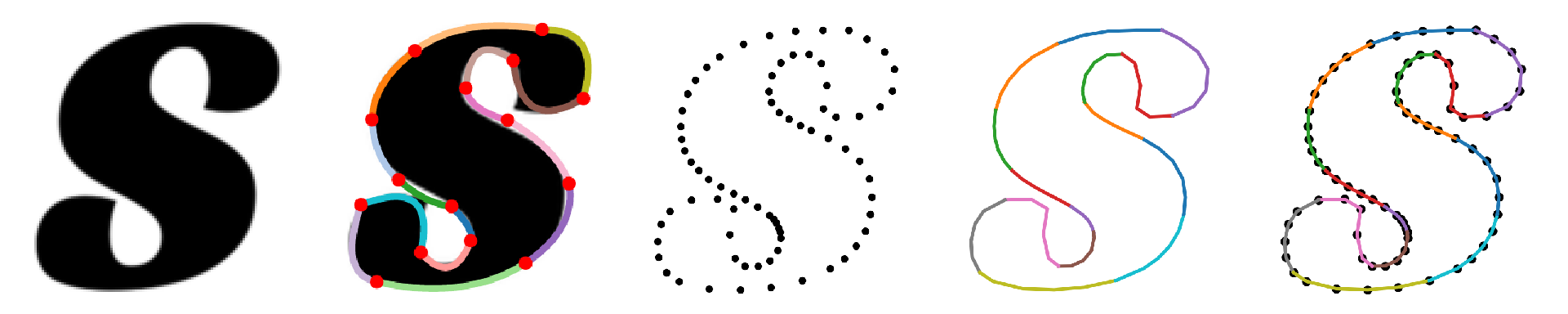} 
    \includegraphics[width=0.45\linewidth]{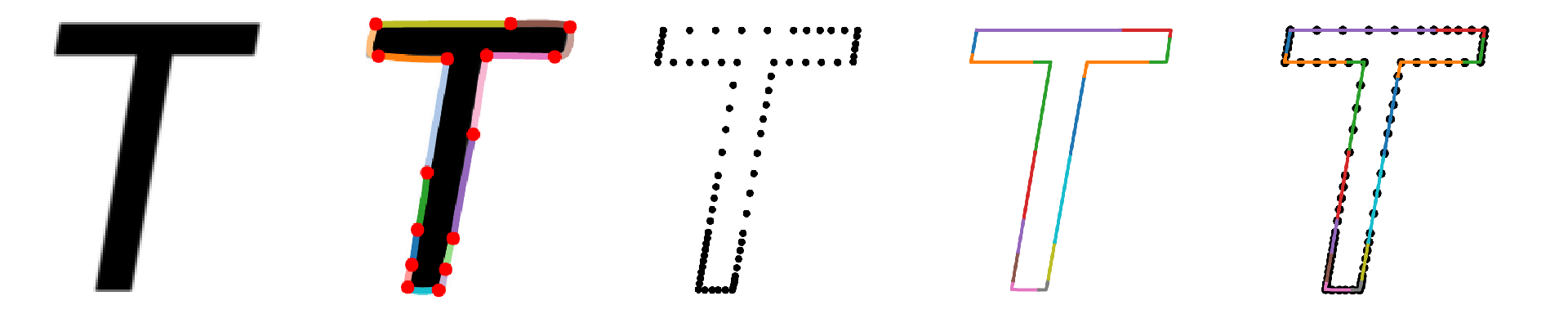} 
    \includegraphics[width=0.45\linewidth]{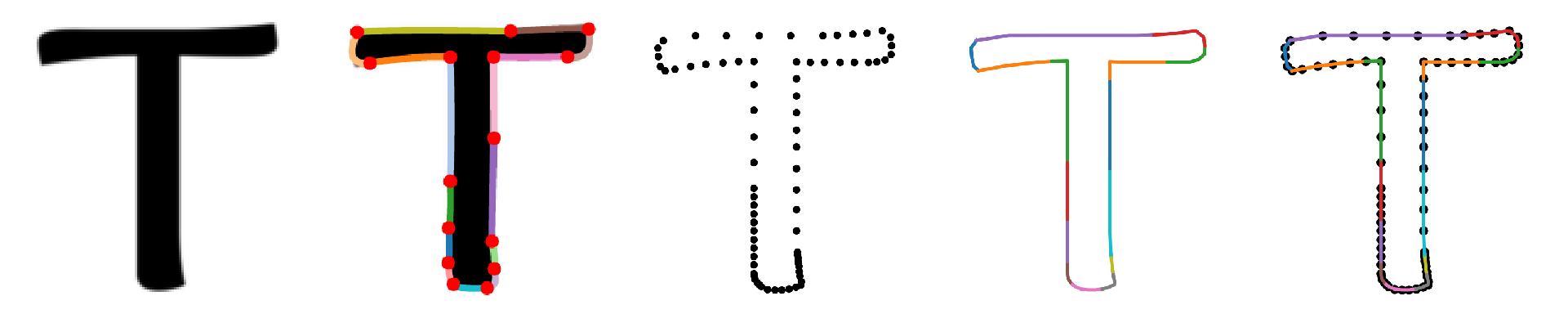} 
    \includegraphics[width=0.45\linewidth]{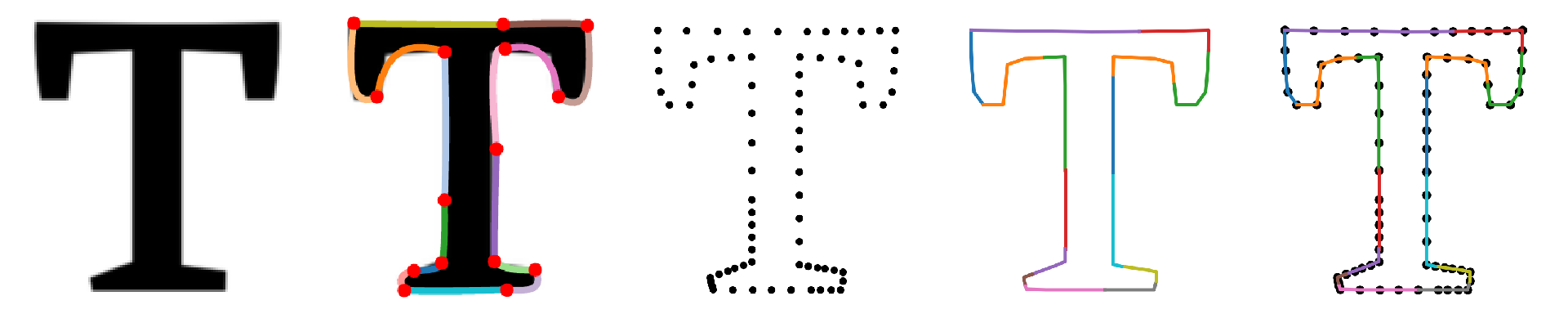} 
    \includegraphics[width=0.45\linewidth]{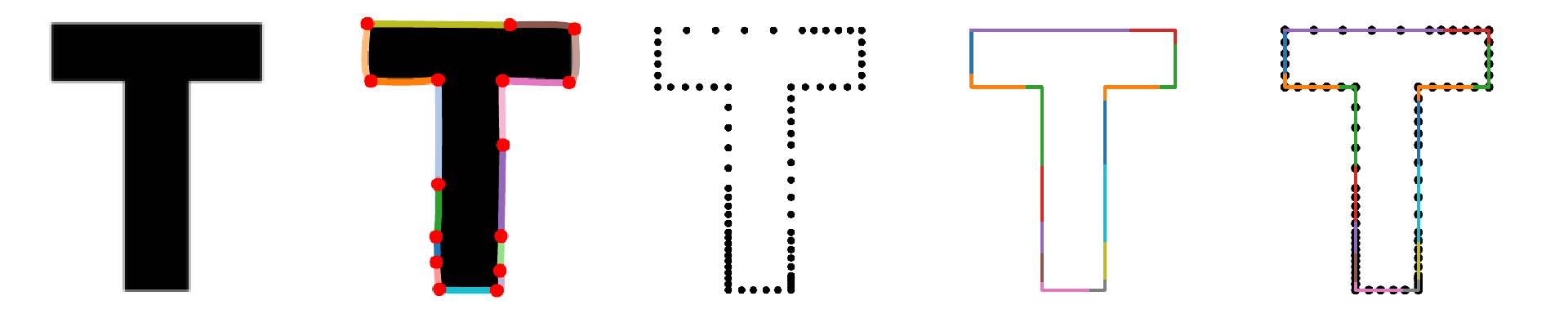} 
    \includegraphics[width=0.45\linewidth]{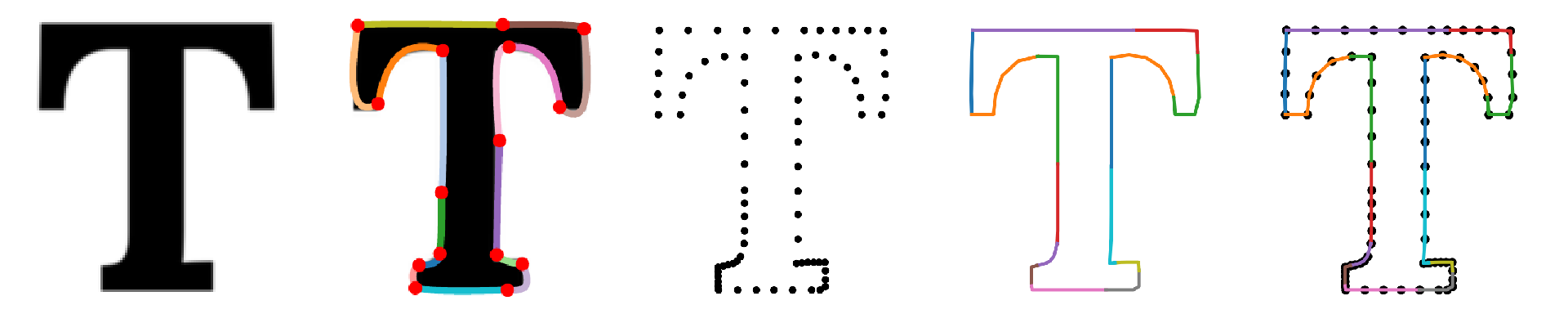} 
    \includegraphics[width=0.45\linewidth]{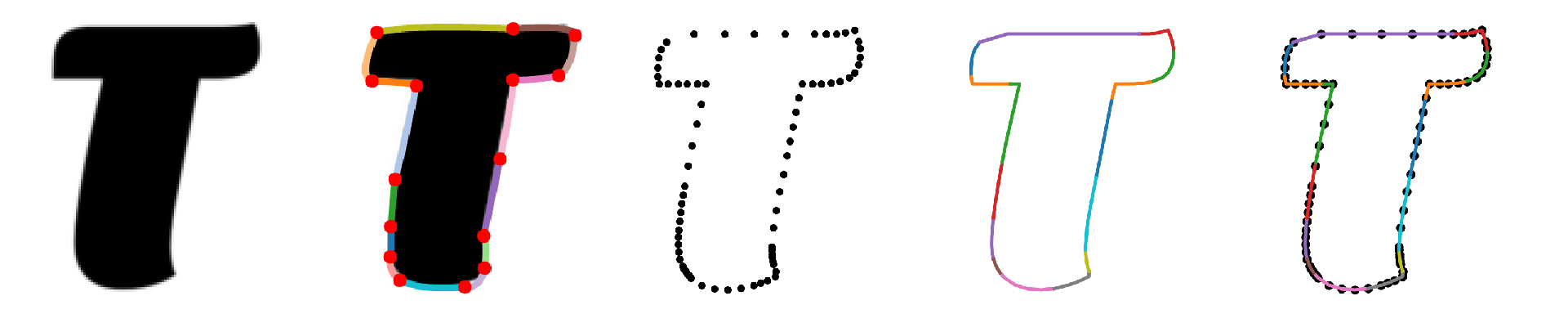} 
    \includegraphics[width=0.45\linewidth]{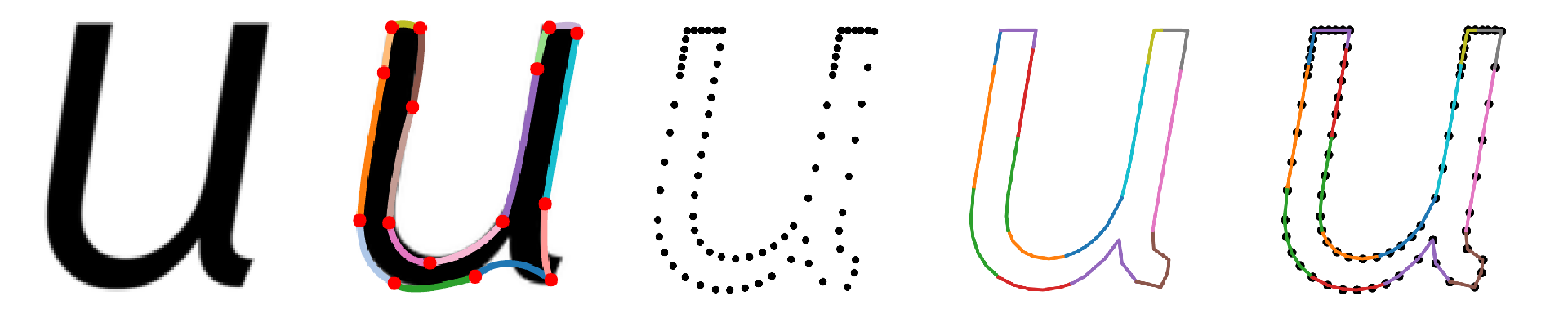} 
    \includegraphics[width=0.45\linewidth]{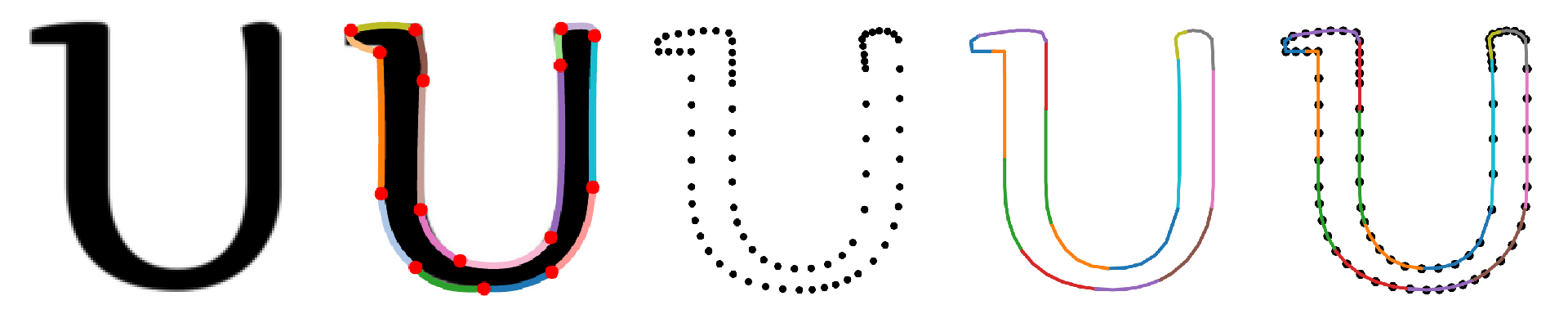} 
    \includegraphics[width=0.45\linewidth]{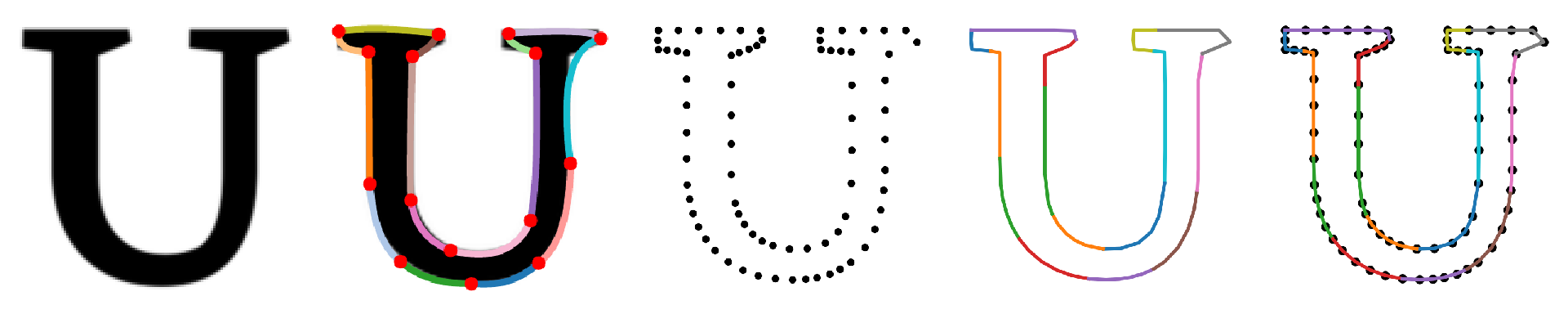} 
    \includegraphics[width=0.45\linewidth]{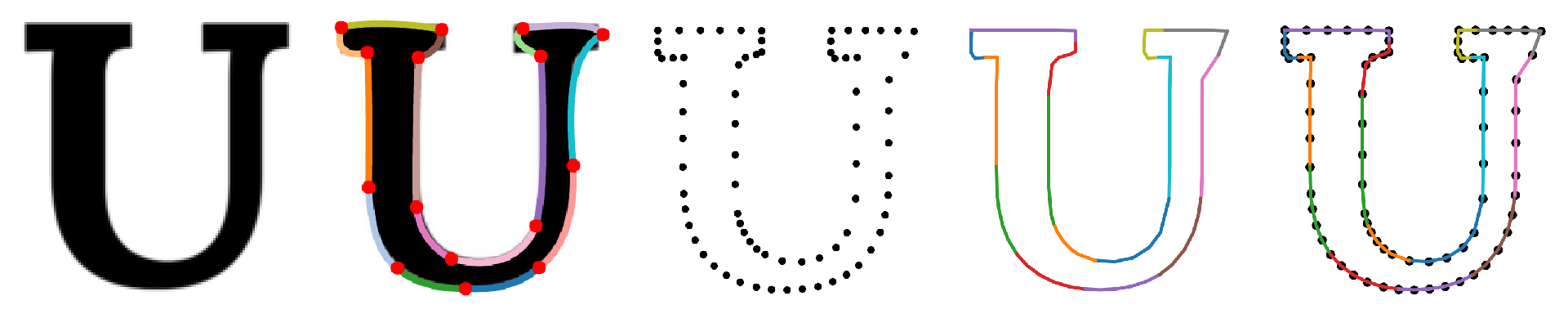} 
    \includegraphics[width=0.45\linewidth]{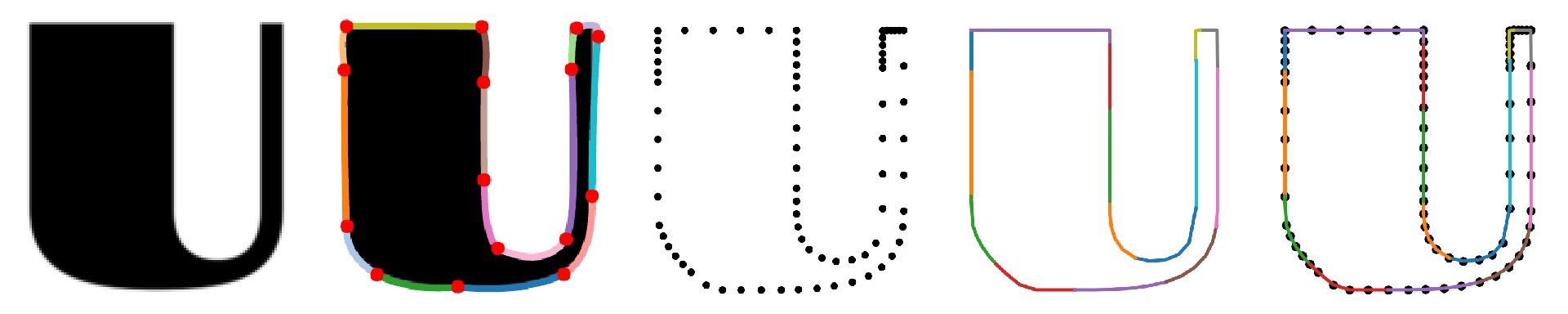} 
    \includegraphics[width=0.45\linewidth]{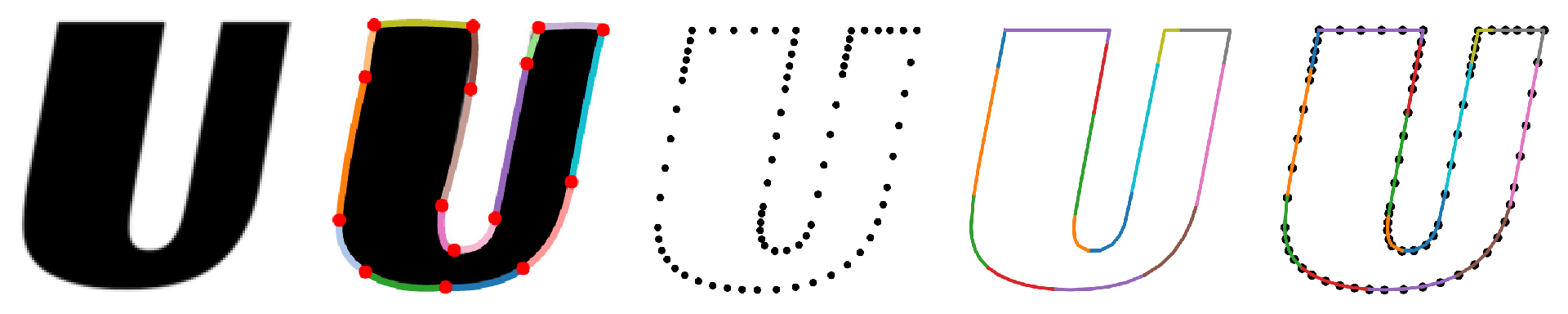} 
    \includegraphics[width=0.45\linewidth]{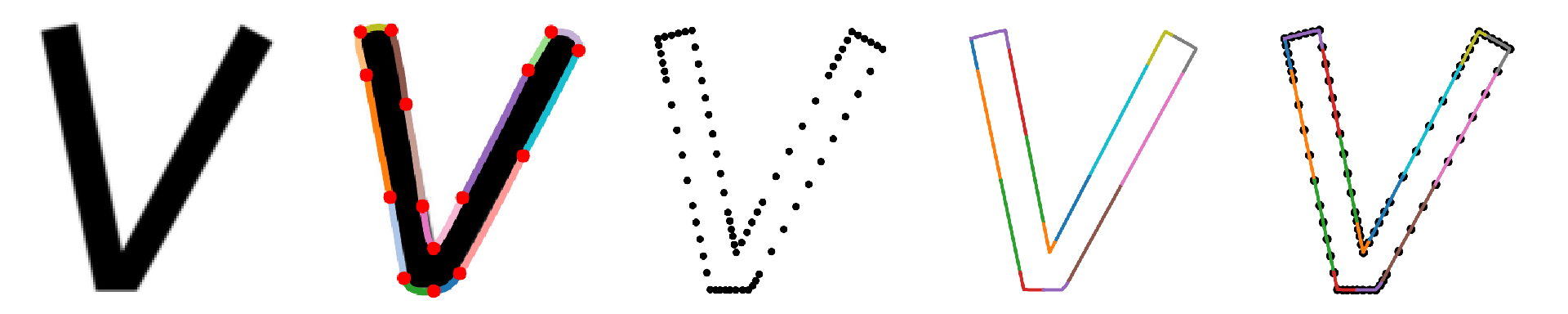} 
    \includegraphics[width=0.45\linewidth]{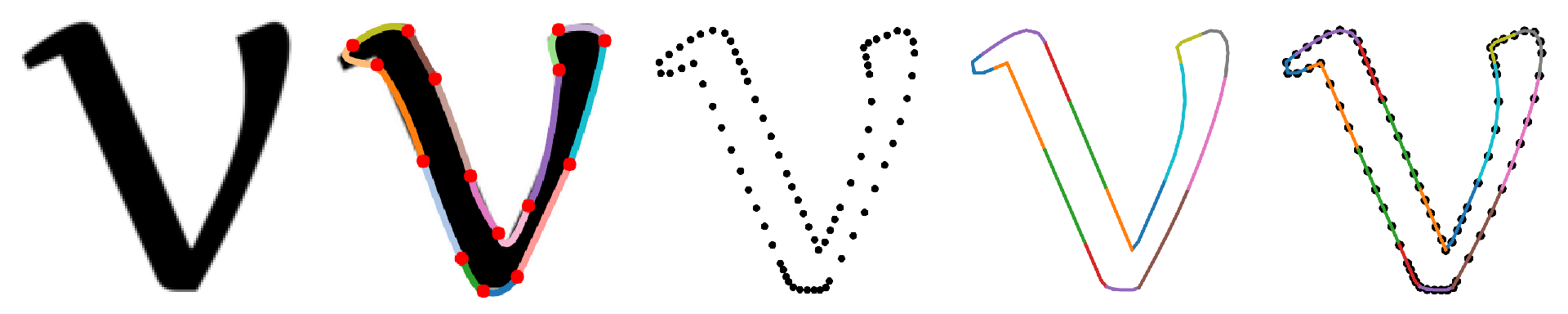} 
    \includegraphics[width=0.45\linewidth]{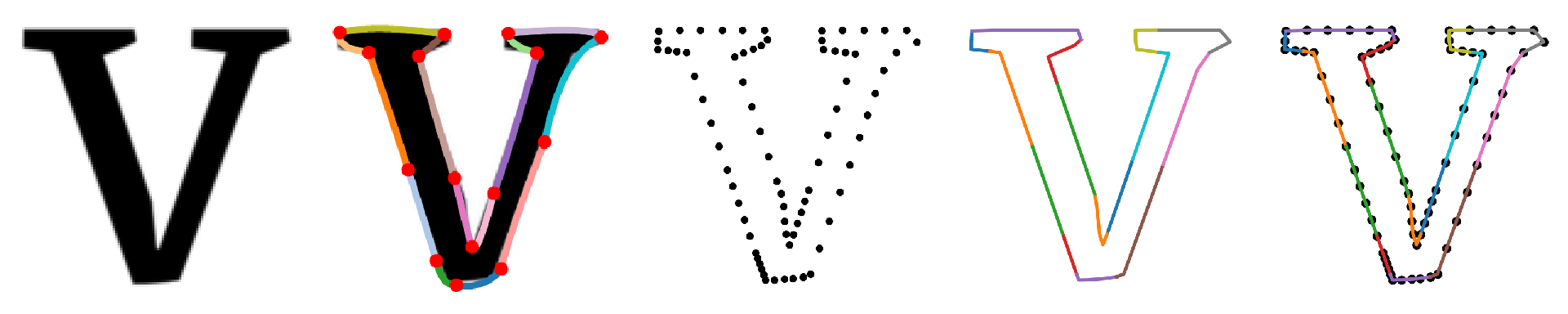} 
    \includegraphics[width=0.45\linewidth]{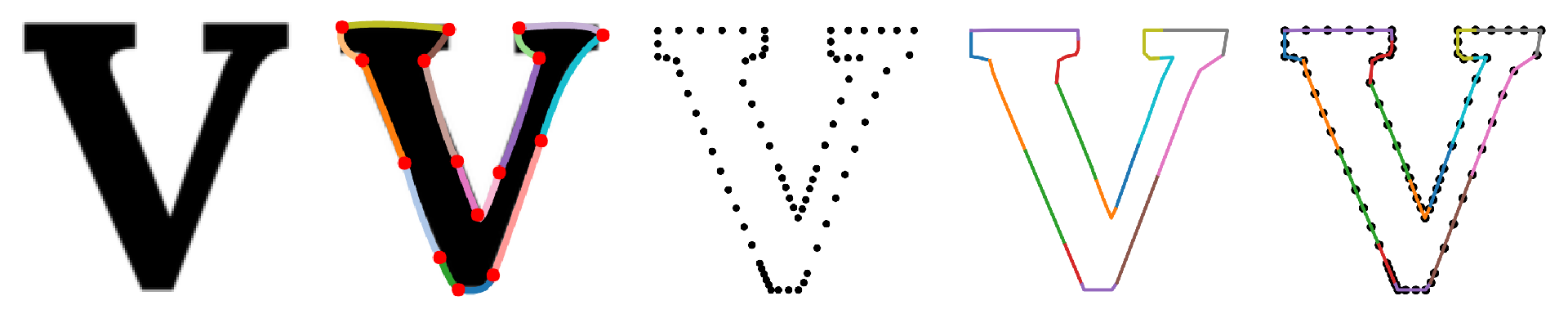} 
    \includegraphics[width=0.45\linewidth]{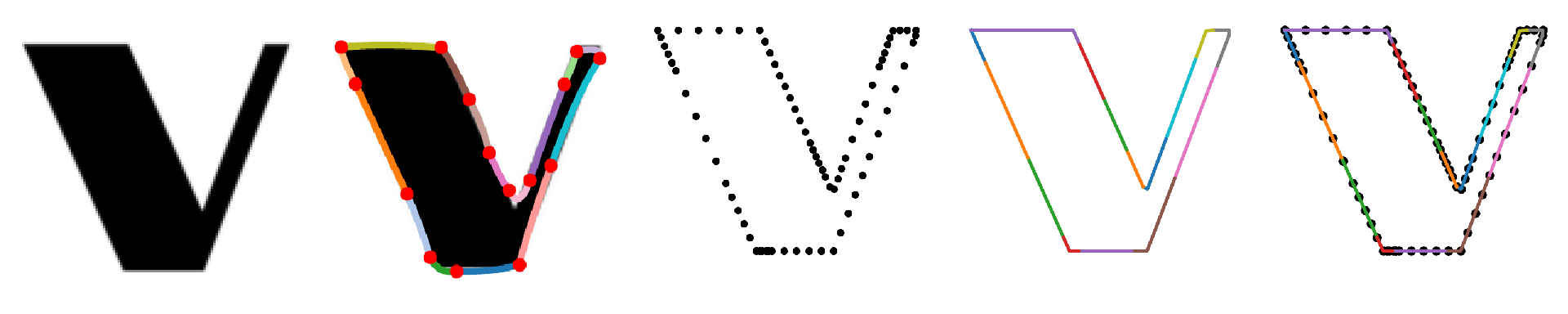} 
    \includegraphics[width=0.45\linewidth]{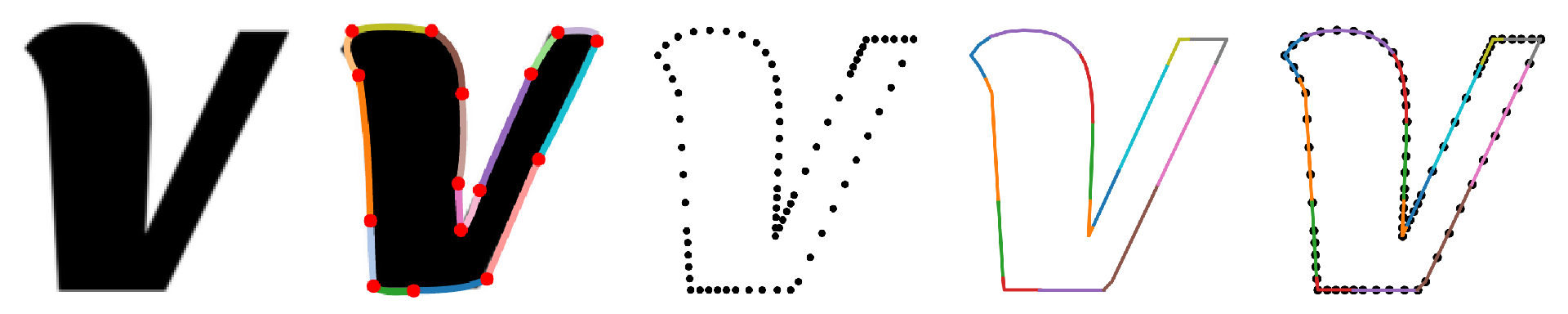} 
    \includegraphics[width=0.45\linewidth]{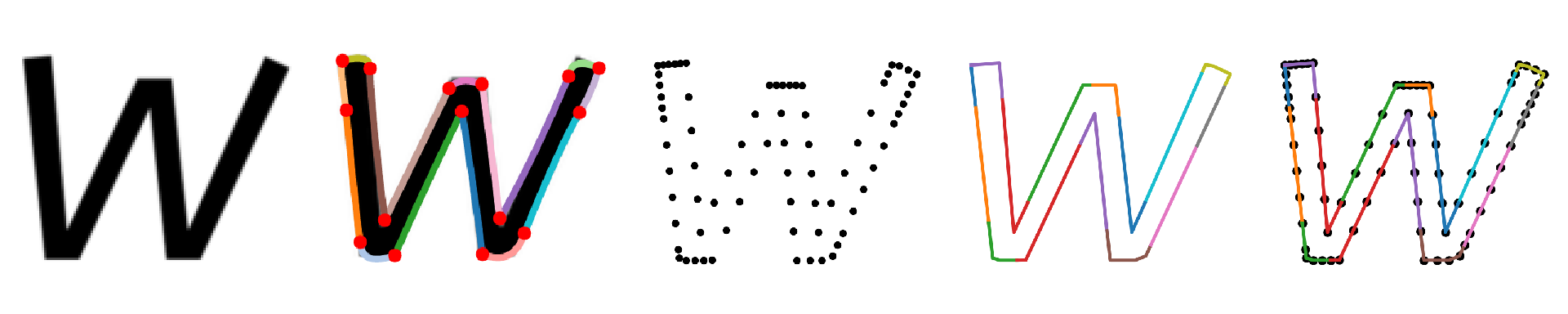} 
    \includegraphics[width=0.45\linewidth]{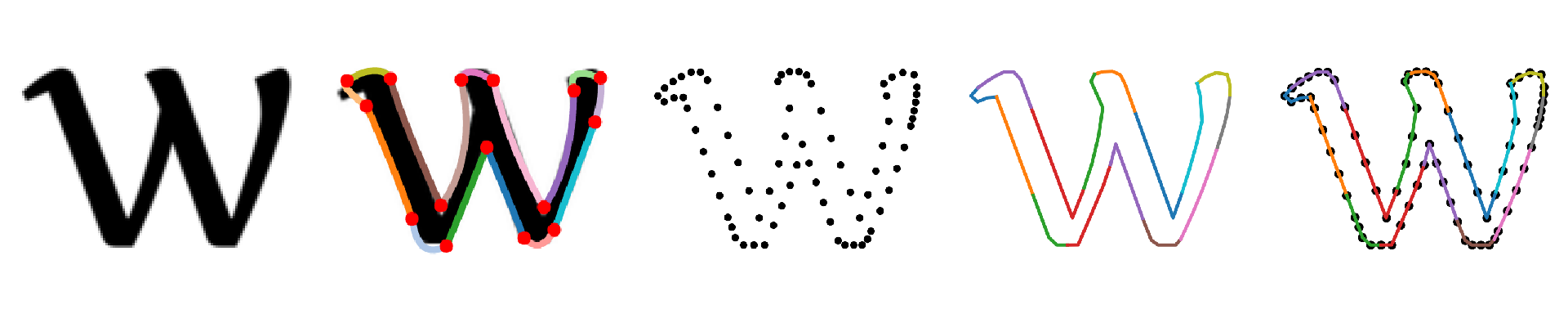} 
    \includegraphics[width=0.45\linewidth]{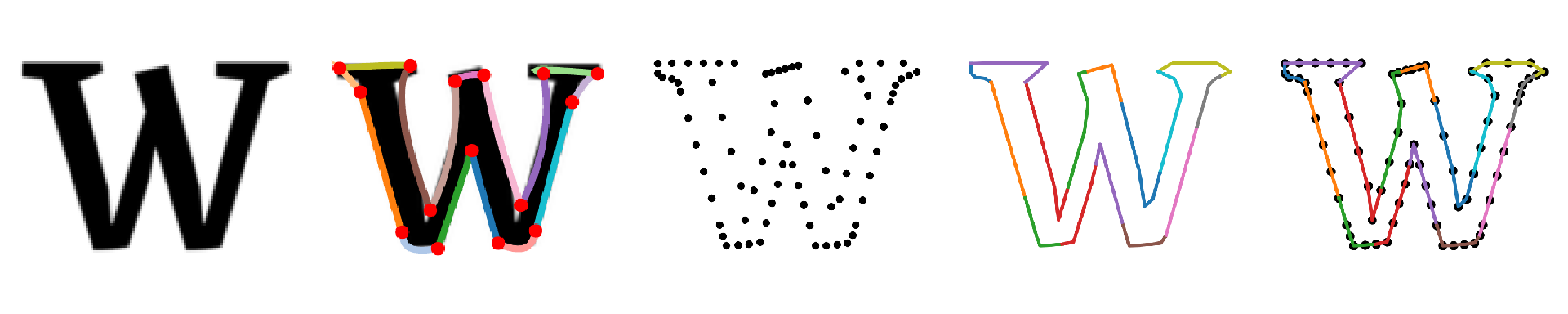} 
    \includegraphics[width=0.45\linewidth]{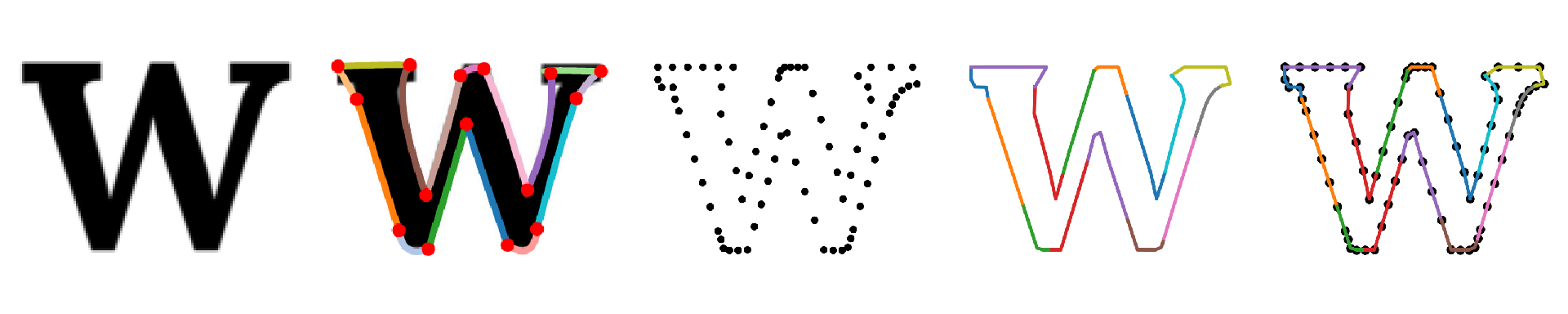} 
    \includegraphics[width=0.45\linewidth]{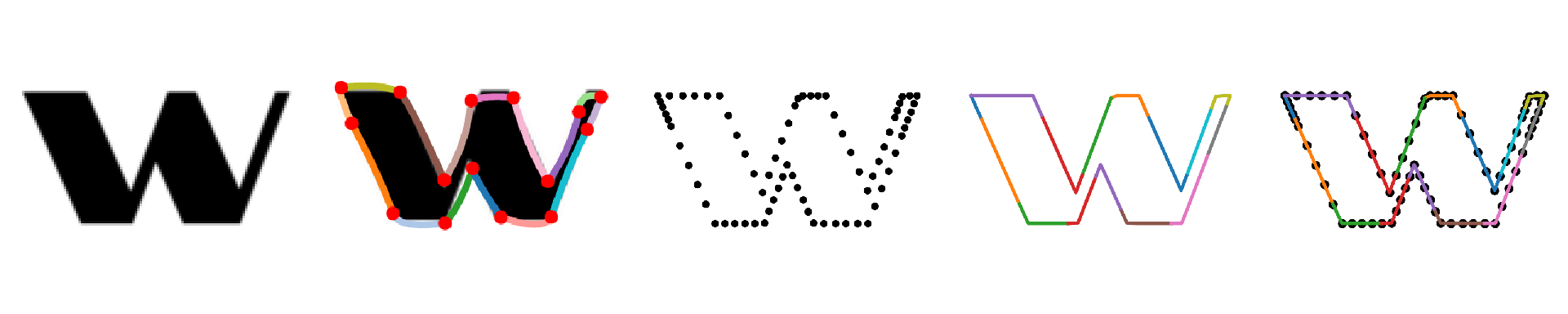} 
    \includegraphics[width=0.45\linewidth]{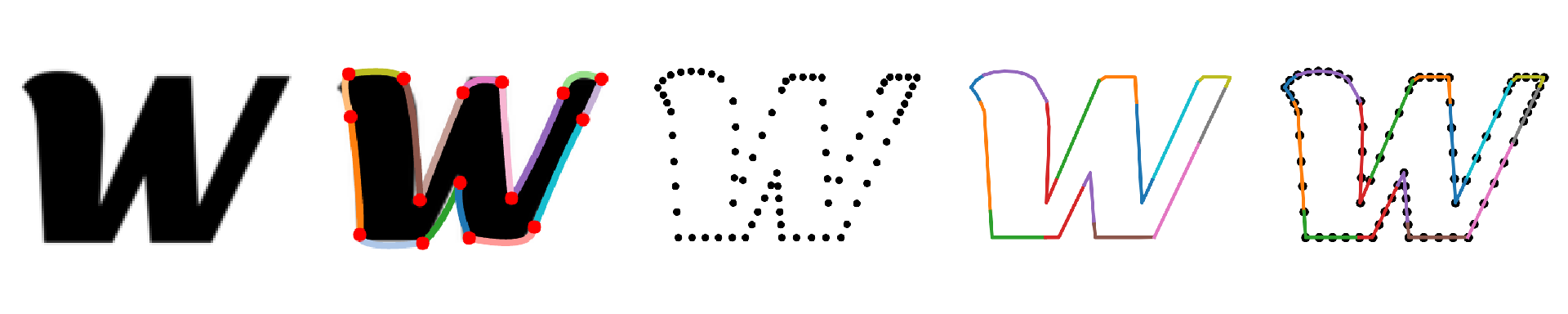} 
    \includegraphics[width=0.45\linewidth]{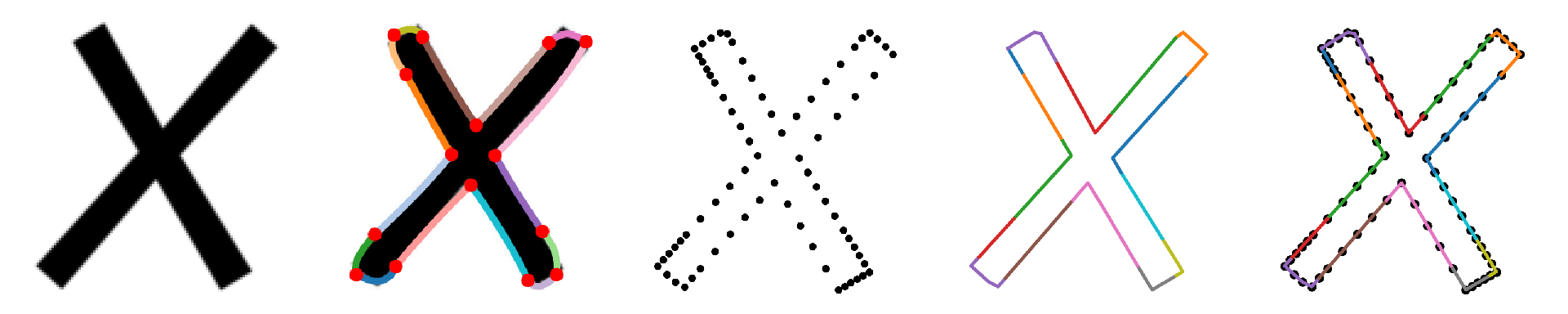} 
    \includegraphics[width=0.45\linewidth]{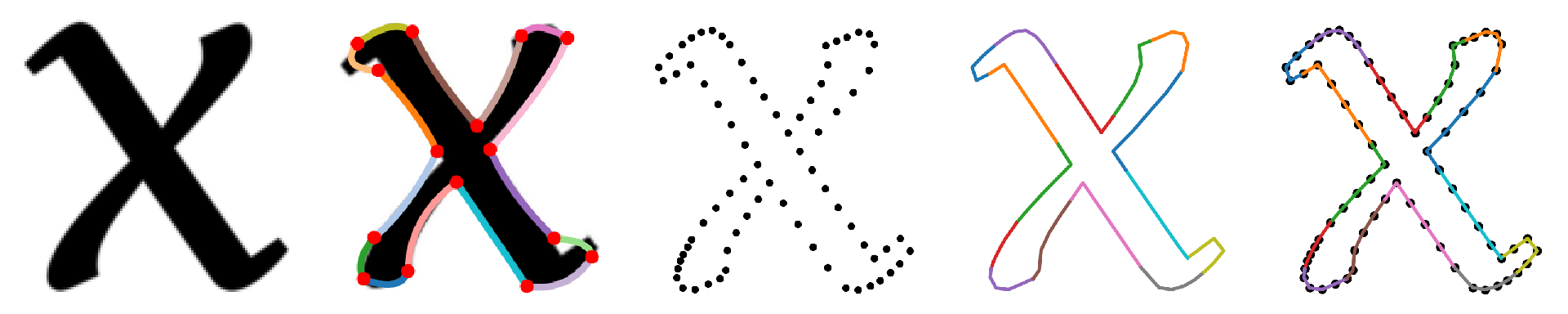} 
    \includegraphics[width=0.45\linewidth]{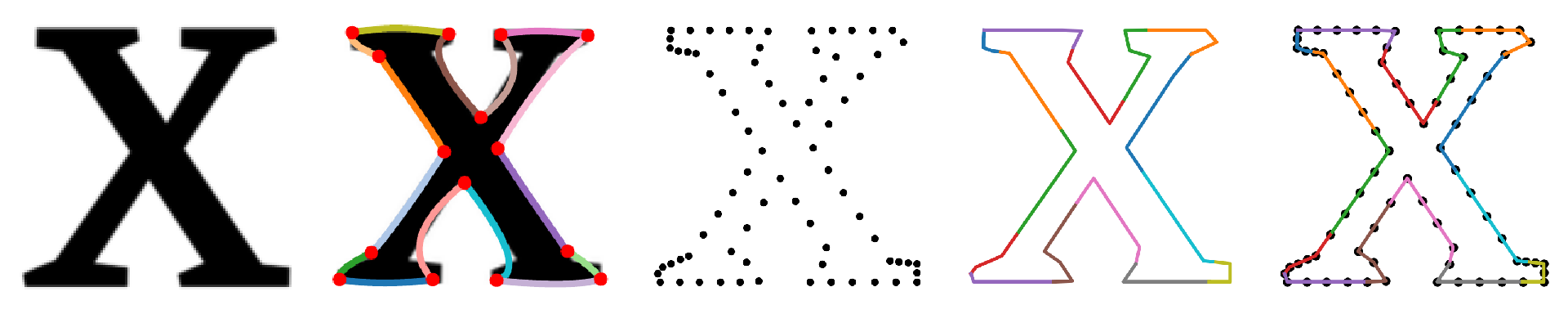} 
    \includegraphics[width=0.45\linewidth]{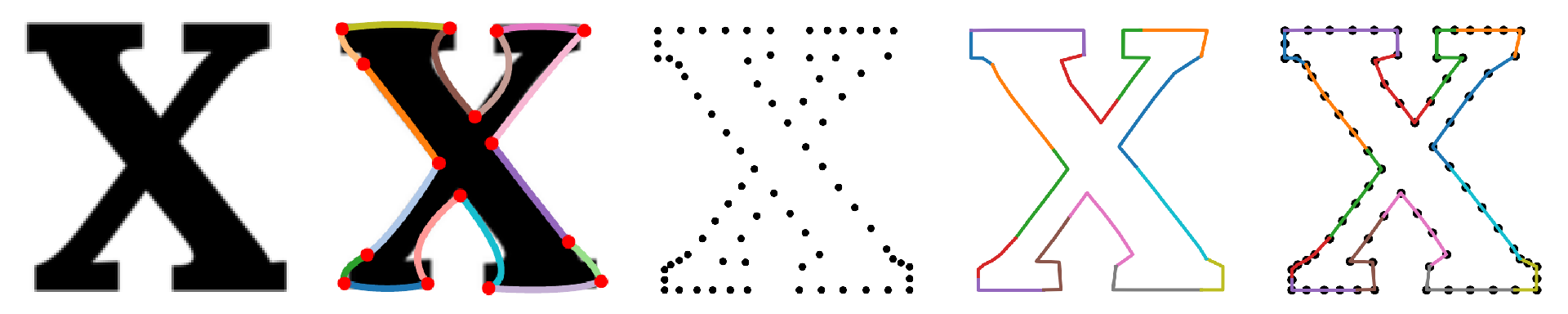} 
    \includegraphics[width=0.45\linewidth]{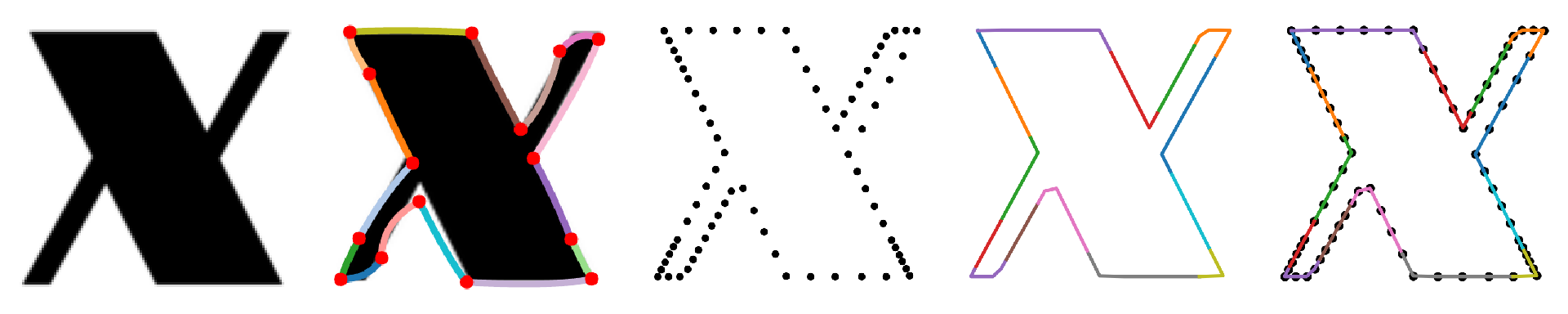} 
    \includegraphics[width=0.45\linewidth]{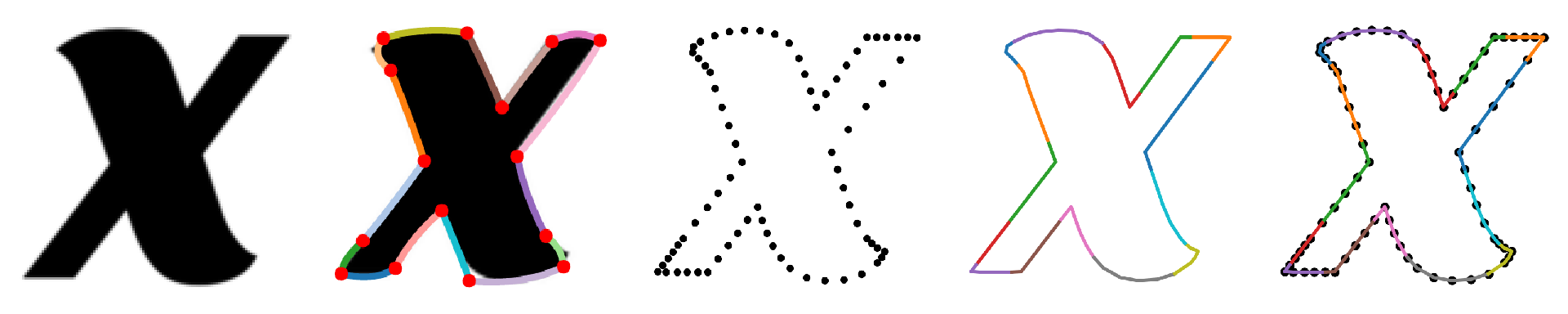} 
    \includegraphics[width=0.45\linewidth]{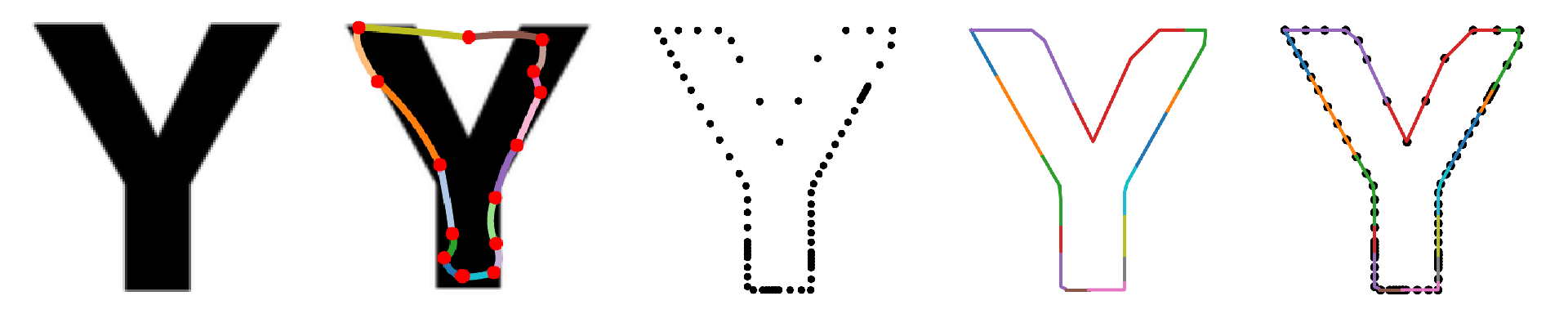} 
    \includegraphics[width=0.45\linewidth]{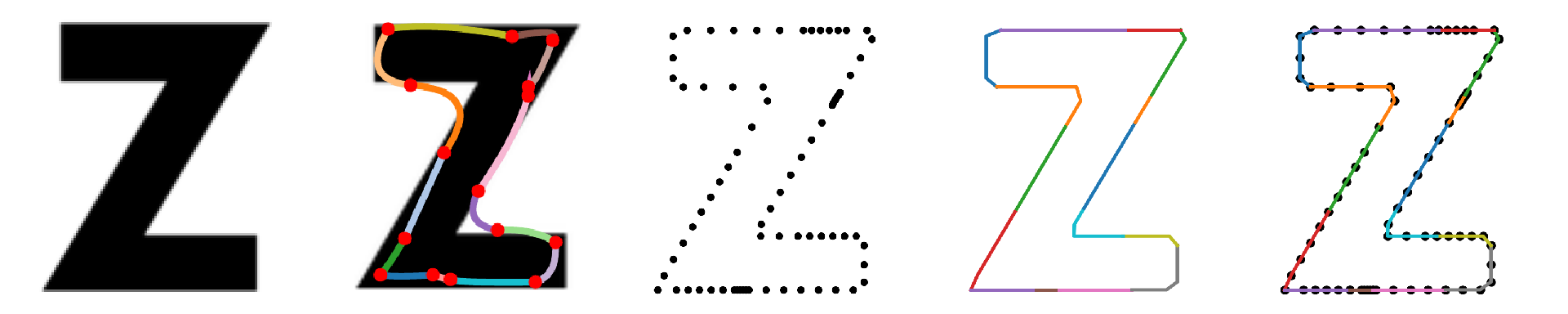} 
    \includegraphics[width=0.45\linewidth]{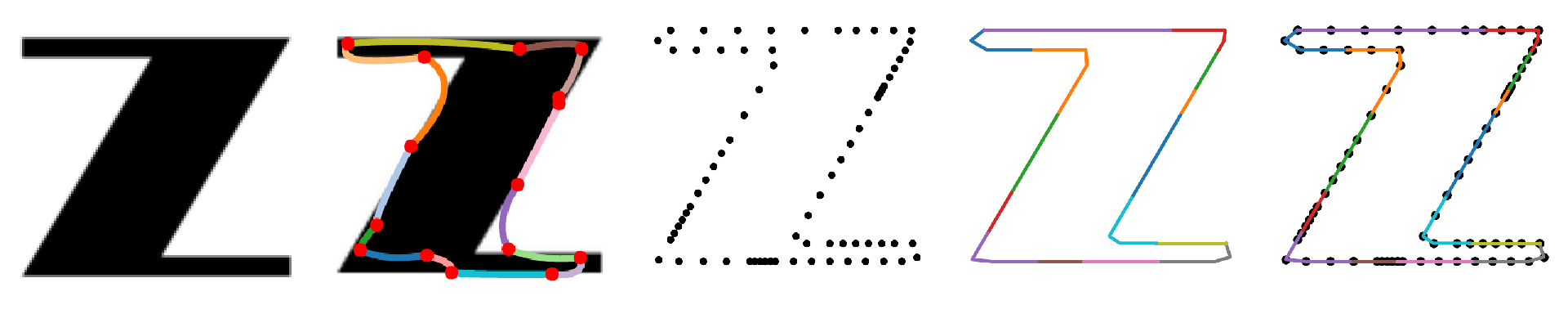} 
    \includegraphics[width=0.45\linewidth]{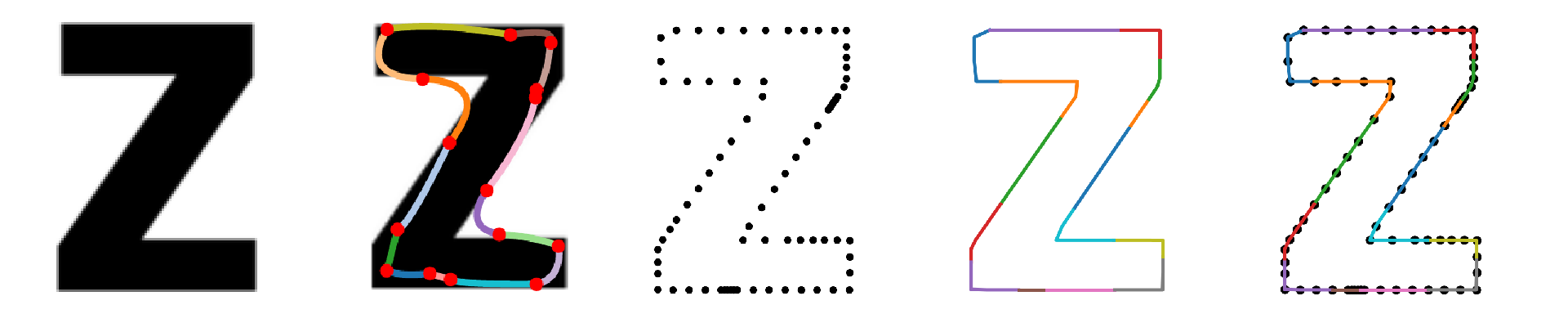}
\end{center}
\caption{Different data modalities of font glyphs. For each example from left to right: image, keypoints, point set, graph template and directed graph representation. } \vspace{4em}
\label{graph_construction}
\end{figure}

\newpage

\ifCLASSOPTIONcaptionsoff
  \newpage
\fi



\bibliographystyle{IEEEtran}
\bibliography{bare_jrnl}
%



%





\begin{IEEEbiography}[{\includegraphics[trim={20cm 0 20cm 0},width=1in,height=1.25in,clip,keepaspectratio]{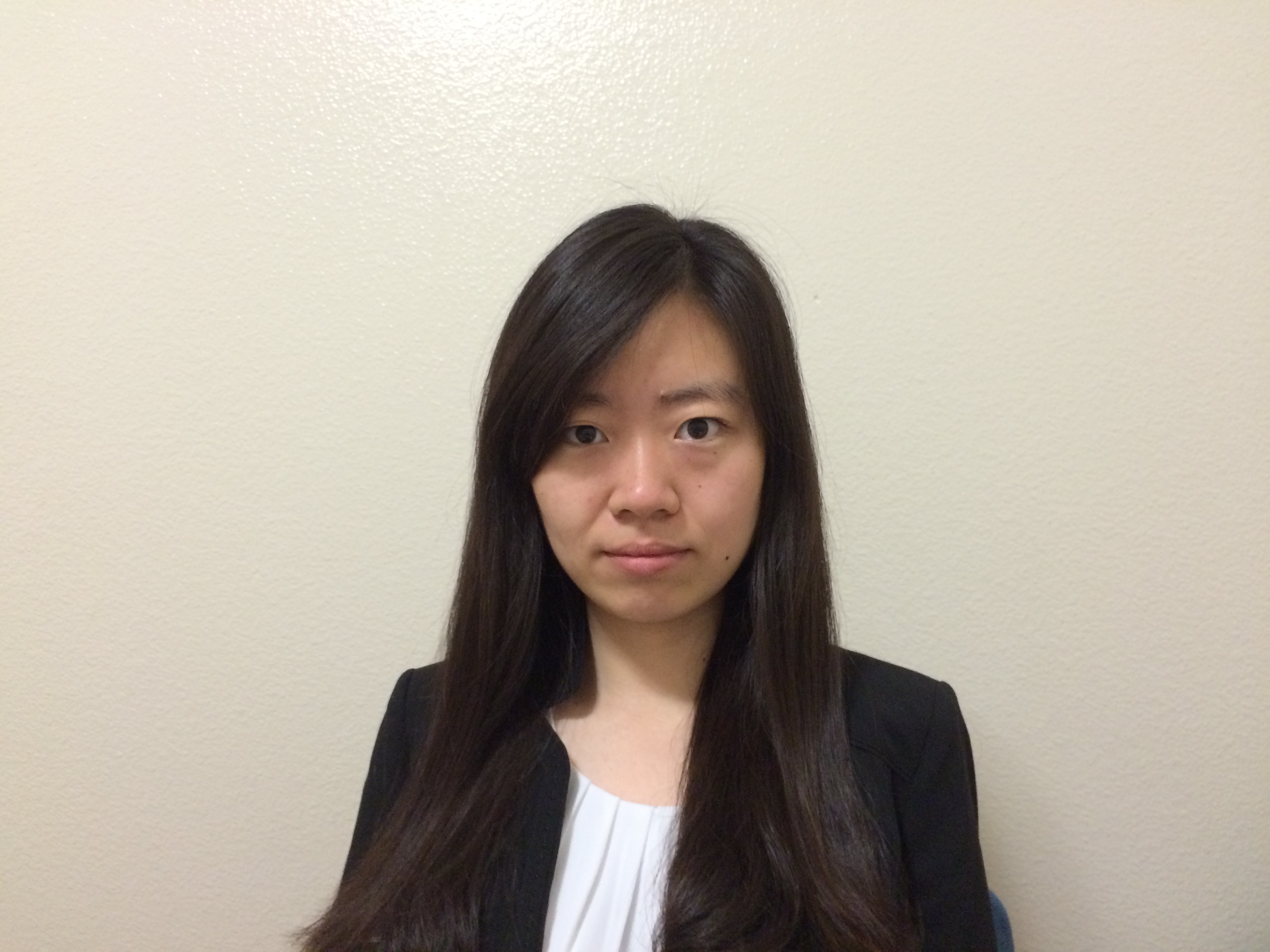}}]{Ye Yuan}
received her Ph.D. degree from the Department of Computer Science and Engineering, Texas A\&M University. Before that, she obtained her Bachelor degree from University of Science and Technology of China. He has also interned at Warmart AI Lab, Bytedance, and Adobe Research. Her research interests are broadly in computer vision and machine learning.
\end{IEEEbiography}

\begin{IEEEbiography}
[{\includegraphics[width=1in,height=1.25in,clip,keepaspectratio]{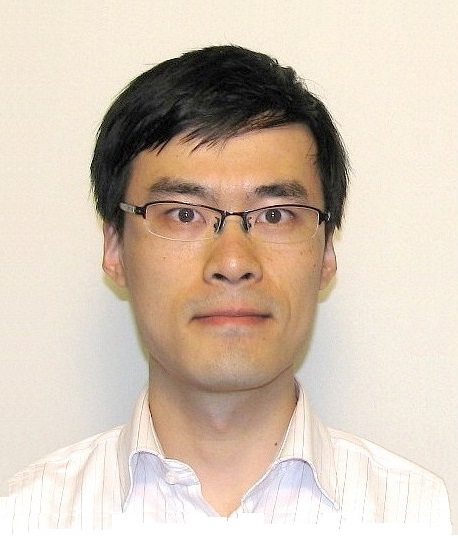}}]
{Zhaowen Wang} received the B.E. and M.S. degrees from Shanghai Jiao Tong  University, China, in 2006 and 2009 respectively, and the Ph.D. degree in ECE from UIUC in 2014. He is currently a Senior Research Scientist with the Creative Intelligence Lab, Adobe Inc. His research focuses on understanding and enhancing images, videos and graphics via machine learning algorithms, with a particular interest in sparse coding and deep learning.
\end{IEEEbiography}
\vspace{-3em}

\begin{IEEEbiography}
[{\includegraphics[width=1in,height=1.25in,clip,keepaspectratio]{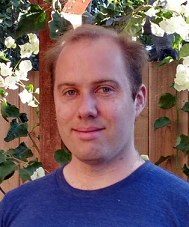}}]
{Matthew Fisher} is a senior research scientist at Adobe Research. He received his PhD from Stanford University supported by a Hertz Fellowship where he worked with Pat Hanrahan in the Graphics Lab. He completed his undergraduate degree at the California Institute of Technology working with Mathieu Desbrun. During a postdoc at Stanford, he has also worked on the data analytics team at Khan Academy.
His research focuses on combining computer graphics, vision, and machine learning to make it faster and more fun to complete creative tasks.
\end{IEEEbiography}
\vspace{-3em}

\begin{IEEEbiography}[{\includegraphics[trim={3cm 0 7cm 0},width=1in,height=1.25in,clip,keepaspectratio]{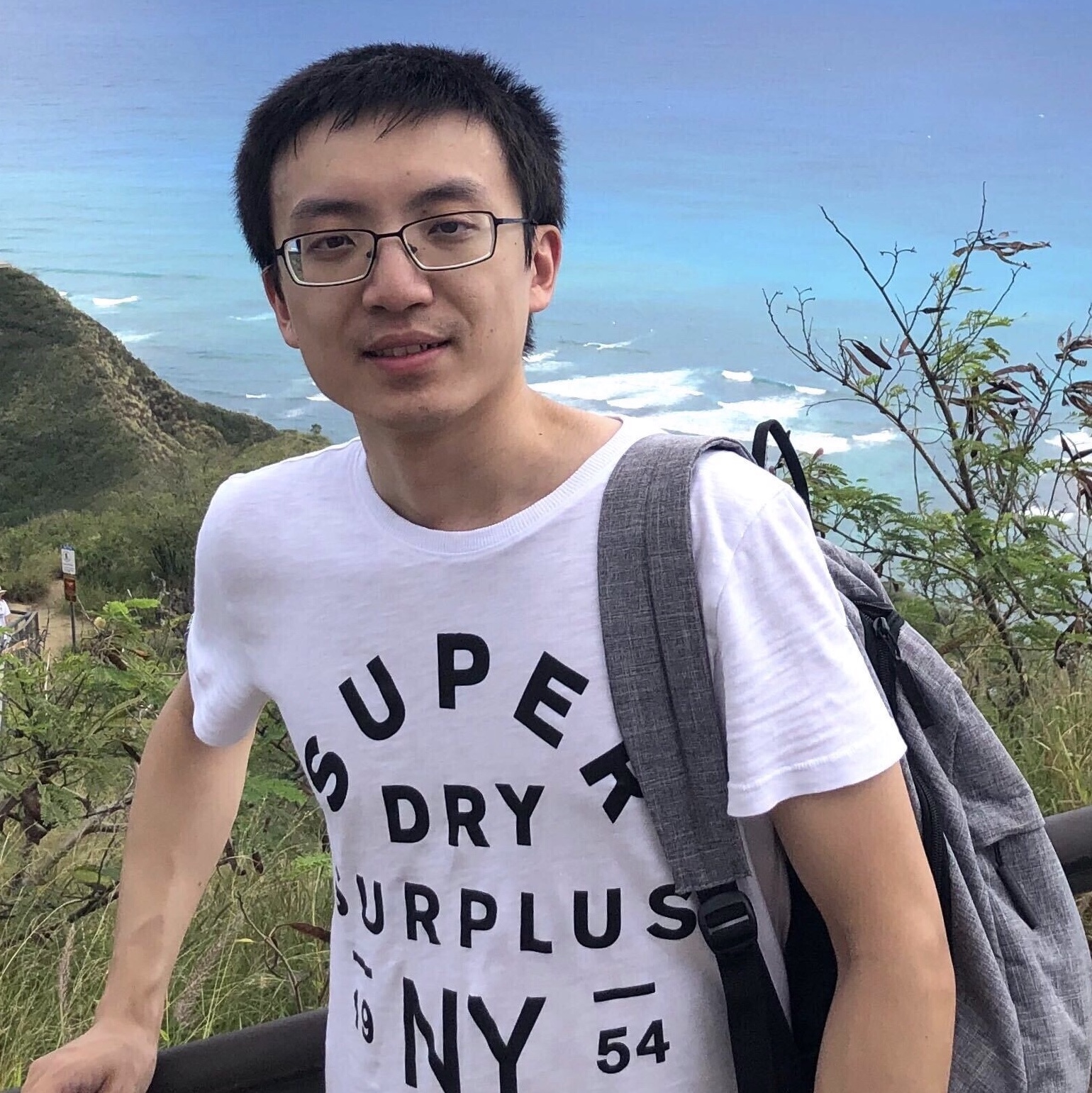}}]{Wuyang Chen}
is a Ph.D. student in Electrical and Computer Engineering at University of Texas at Austin. He received his M.S. degree in Computer Science from Rice University in 2016, and his B.S. degree from University of Science and Technology of China in 2014. Wuyang’s research focuses on addressing domain adaptation/generalization, self-supervised learning, and AutoML. He has published 6 papers in CVPR, ICLR and ICML as the first author, and co-authored more.
\end{IEEEbiography}

\begin{IEEEbiography}[{\includegraphics[trim={0 2.5cm 0 0},width=1in,height=1.25in,clip,keepaspectratio]{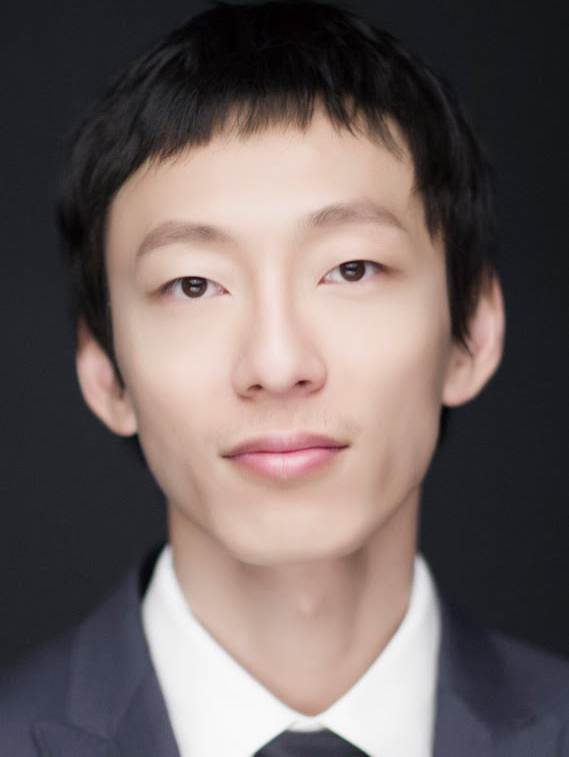}}]{Zhifei Zhang}
is currently a research engineer in Adobe Research. He obtained the Ph.D. in 2018 supervised by Prof. Hairong Qi in computer engineering at the University of Tennessee. He received the B.S. and M.S. in 2010 and 2013, respectively, from the Northeastern University and Zhejiang University, China. His interest lies in deep learning based image synthesis, computer vision, text detection and recognition, etc.
\end{IEEEbiography}

\begin{IEEEbiography}[{\includegraphics[trim={0 10cm 0 3cm},width=1in,height=1.25in,clip,keepaspectratio]{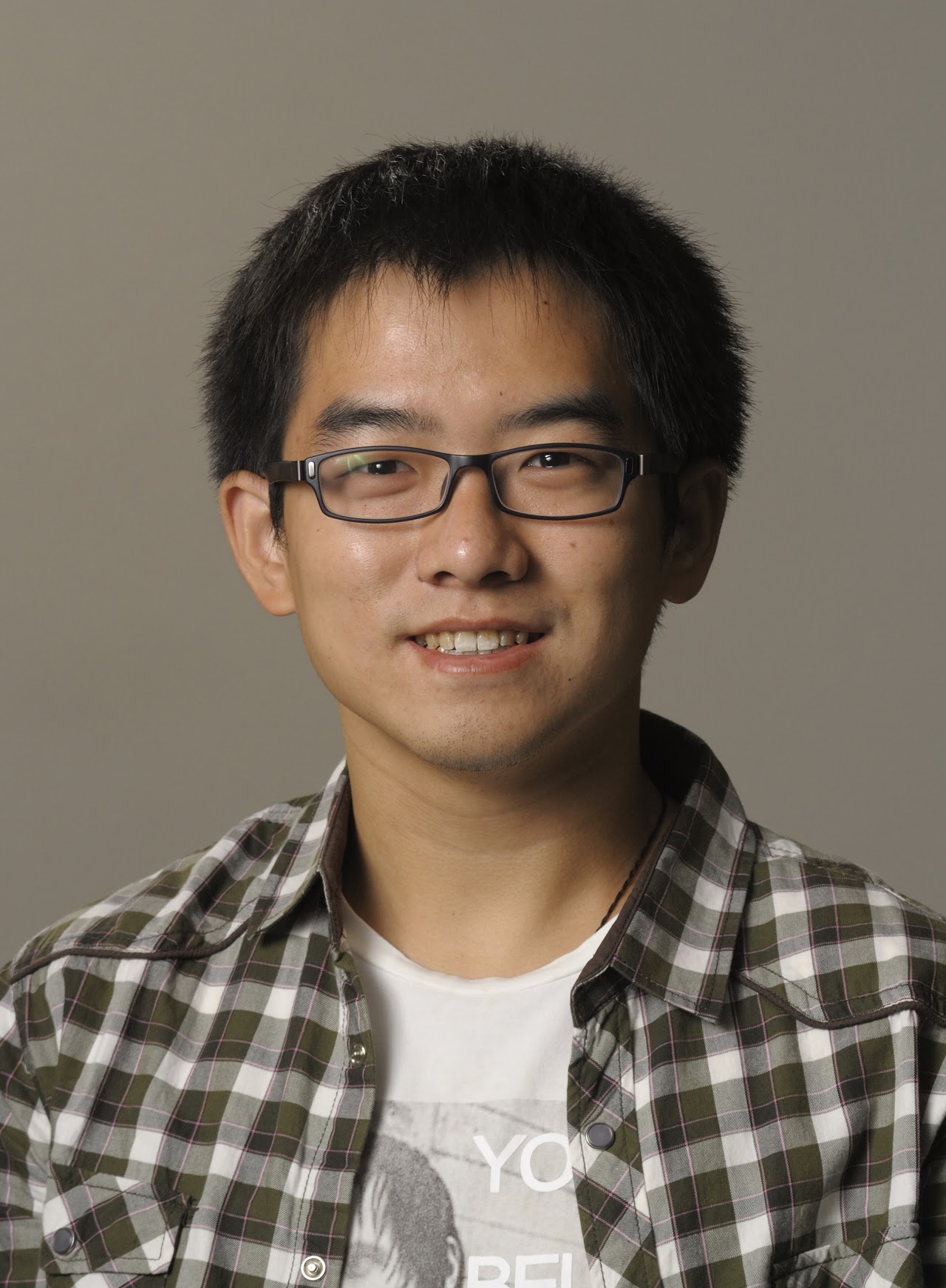}}]{Zhangyang Wang}
is currently an Assistant Professor of Electrical and Computer Engineering at University of Texas at Austin. He received his Ph.D. degree in ECE from UIUC in 2016, advised by Professor Thomas S. Huang; and his B.E. degree in EEIS from USTC in 2012. Prof. Wang is broadly interested in the fields of machine learning, computer vision, optimization, and their interdisciplinary applications. His latest interests focus on automated machine learning (AutoML), learning-based optimization, machine learning robustness, and efficient deep learning. His research is gratefully supported by NSF, DARPA, ARL/ARO, as well as a few more industry and university grants. He has co-authored over 130 papers, 2 books and 1 chapter; and has been granted 3 patents. He has received many research awards and scholarships, including most recently an ARO Young Investigator award, an IBM faculty research award, an Amazon research award (AWS AI), an Adobe Data Science Research Award, a Young Faculty Fellow of TAMU, and four research competition prizes from CVPR/ICCV/ECCV.
\end{IEEEbiography}

\begin{IEEEbiography}
[{\includegraphics[width=1in,height=1.25in,clip,keepaspectratio]{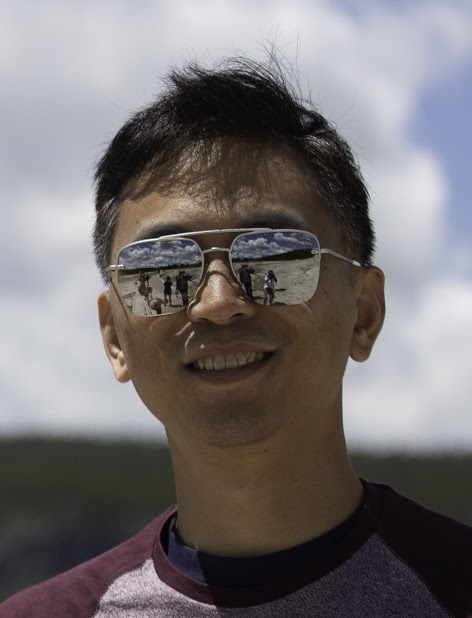}}]
{Hailin Jin} is a Senior Principal Scientist at Adobe Research. He received his M.S. and Ph.D. in EE from WUSTL in 2000 and 2003. Between fall 2003 and fall 2004, he was a postdoc researcher at the CS Department, UCLA. His current research interests include deep learning, computer vision, and natural language processing. His work can be found in many Adobe products, including Photoshop, After Effects, Premiere Pro, and Photoshop Lightroom.
\end{IEEEbiography}
\vspace{-3em}




\end{document}